\documentclass{article}
\usepackage{iclr2027_conference,times}

\usepackage{amsmath,amsfonts,bm}

\def\eqref#1{equation~\ref{#1}}

\def\1{\bm{1}}

\DeclareMathAlphabet{\mathsfit}{\encodingdefault}{\sfdefault}{m}{sl}
\SetMathAlphabet{\mathsfit}{bold}{\encodingdefault}{\sfdefault}{bx}{n}

\usepackage{hyperref}
\usepackage{url}
\usepackage{graphicx}
\usepackage{booktabs}
\usepackage{amsmath}
\usepackage{amssymb}
\usepackage{nicefrac}
\usepackage{microtype}
\usepackage{enumitem,kantlipsum}
\usepackage{xcolor}
\usepackage{graphicx}
\usepackage{pdflscape}
\usepackage{afterpage}
\usepackage{array}
\usepackage{placeins}

\newcommand{\IGP}{\mathrm{IGP}}
\newcommand{\CWA}{\mathrm{CWA}}
\newcommand{\SemD}{\mathrm{SemD}}
\newcommand{\CFR}{\mathrm{CFR}}
\newcommand{\addtext}[1]{#1}
\newenvironment{addblock}{}{}
\newif\ifrevmarks
\revmarksfalse
\ifrevmarks
  \colorlet{revblue}{blue}
  \colorlet{revred}{red}
\else
  \colorlet{revblue}{black}
  \colorlet{revred}{black}
\fi
\newcommand{\rev}[1]{{\color{revblue}#1}}
\newcommand{\revr}[1]{{\color{revred}#1}}
\newcommand{\revcolor}{\color{revblue}}
\newenvironment{revblock}{\par\color{revblue}}{\par}
\makeatletter
\newcommand{\igp@sym}[2]{\@ifnextchar_{\igp@sym@sub{#1}{#2}}{\rev{#1^{\mathrm{#2}}}}}
\def\igp@sym@sub#1#2_#3{\rev{#1^{\mathrm{#2}}_{#3}}}
\DeclareRobustCommand{\spop}{\igp@sym{s}{pop}}
\DeclareRobustCommand{\spair}{\igp@sym{s}{pair}}
\DeclareRobustCommand{\dpop}{\igp@sym{d}{pop}}
\DeclareRobustCommand{\dpair}{\igp@sym{d}{pair}}
\makeatother
\title{Inferred Generative-Process Diversity Predicts Correlated Failure Across Language Models}

\newif\ifpreprint
\preprinttrue
\ifpreprint
  \author{Ross Tieman$^{1,2}$ \& Evan Markou$^{1,2}$ \\
  $^{1}$Australian National University \quad $^{2}$Moir\'e Research \\
  \texttt{\{ross.tieman, evan.markou\}@anu.edu.au}}
  \iclrfinalcopy
\else
  \author{Anonymous Authors \\
  Anonymous Affiliation \\
  \texttt{anonymous@example.com}}
\fi

\begin{document}
\maketitle
\ifpreprint\lhead{Preprint}\fi

\begin{abstract}
  As language models are increasingly deployed in multi-model systems, robust system function depends on constituent models failing independently. Diversity is widely observed to buffer complex systems against correlated failure, but only when the type of diversity is relevant to the failure mode of concern. Existing \rev{diversity measures compare model outputs, by token frequency or by meaning. We instead ask how differently two models produce their outputs, which Algorithmic Information Theory formalises as the normalised information distance between them, based on the length of the shortest program (generative process) converting a response from one model into a response from the other. The processes themselves cannot be observed, so we evaluate this distance through their outputs. We approximate it with Normalised Compression Distance between response byte strings, residualised against a permutation control that removes what byte frequencies and string length alone would explain. We therefore term it inferred generative-process (IGP) diversity.} Estimated from responses of 38 language models to open-ended chat prompts, the measure predicts which model pairs choose the same wrong answer on ten disjoint benchmark families, beyond semantic similarity and capability, with a \rev{modest} negative partial association on every benchmark (mean $-0.216$, 95\% CI $[-0.309,-0.122]$). Semantic distance predicts no reduction in correlated failure once process diversity is held fixed. Five-model ensembles selected to maximise the Vendi score of the measure at fixed semantic distance converge on the same wrong answer less often than capability-matched random ensembles in every semantic band, whereas the reverse selection does not. \rev{Under majority vote, ensembles selected on the measure give about one in ten fewer wrong answers, and slightly more correct answers, than capability-matched ensembles.} Because it requires only observed outputs, inferred generative-process diversity provides a practical \rev{model-free, parameter-free} approach for auditing and constructing model \rev{ensembles where correlated failure limits reliability}.
\end{abstract}
\section{Introduction}
\label{sec:intro}

Language models are increasingly deployed as components of larger systems. Ensembles aggregate their predictions, oversight schemes assign models to critique one another, and multi-agent architectures distribute tasks across many model instances. Combining models in such ways can improve collective performance \rev{by providing complementary capabilities, and reliability by catching errors so that function is preserved} when one component fails \citep{kuncheva_measures_2003,wood_unified_2023}. Neither benefit follows from model count alone. If models share behaviours that \rev{reproduce} the same errors, an additional vote or reviewer adds nominal redundancy without adding another line of defence. \rev{In the 2026 Hugging Face hack, hundreds of agents coordinated on harmful activity that no operator had sanctioned, and refusals among the agents involved were rare} \citep{greenblatt_hugging_2026,huggingface_anatomy_2026,openai_hugging_2026}. A population of near-identical agents can fail as one, and counting agents, providers, or model families says little about whether it will. What matters is whether the components differ in ways that are relevant to the function and failure modes of the system.

Research on ecological resilience and collective adaptation makes this distinction explicit. Functional diversity can broaden the capabilities available to a collective, whereas response diversity---variation in how components contributing to the same function respond to perturbation---is what enables redundancy to buffer collective function against failure \citep{holling_resilience_1973,yachi_biodiversity_1999,elmqvist_response_2003}. \addtext{Heterogeneous information and strategies likewise preserve alternative responses as conditions change, whereas homogenisation drives a collective to converge on the same locally effective behaviour \citep{galesic_beyond_2023,tieman_landscape_2025}.} This motivates the question, \rev{\emph{which dimension of model variation is relevant to the failure mode that redundancy aims to mitigate?}}

Contemporary language models differ in provider, architecture, size, training procedure, and interface, yet these attributes do not ensure behavioural independence. Models from different providers select the same wrong answers more often than \rev{their individual error rates would predict} \citep{goel_great_2025,kim_correlated_2025}\rev{, and open-ended responses across model populations show substantial semantic similarity} \citep{jiang_artificial_2025}. Semantic similarity captures whether outputs express similar meanings\rev{. However, semantically similar answers can be produced by different processes, and semantically divergent answers by the same process. Existing diversity measures compare outputs, whether by token frequency \citep{yax_phylolm_2024} or by meaning \citep{jiang_artificial_2025}, and so cannot separate these two cases. DFPE and LLM-TOPLA select ensembles by response fingerprints and by error diversity on the target task \citep{cohen_dfpe_2025,tekin_llm-topla_2024}. Our measure needs no labelled data or tuning on the target task, as it is computed from raw responses to unrelated prompts. Appendix~\ref{app:phylolm} compares it with PhyloLM, and Appendix~\ref{app:extended-related-work} discusses related work.}

We introduce \rev{\emph{inferred generative-process (IGP) diversity}} to distinguish these \rev{cases}. A language model is a stochastic process that maps a prompt and context to a distribution over output sequences. \rev{The measure asks how differently two models produce their outputs, as opposed to how different a property of those outputs, such as meaning, is. Intuitively, models that produce answers in the same way are more likely to fail in similar ways, on the same inputs. Drawing from Algorithmic Information Theory (AIT), the difference between these generating processes can be described by the Normalised Information Distance (NID), the length of the shortest program converting a response from one model into a response from the other. The processes themselves cannot be observed, so we evaluate this distance through their outputs, over many responses that the two models give to the same prompts. As NID is not computable we approximate it with Normalised Compression Distance (NCD) between response byte strings, residualised against a permutation control that preserves byte frequencies and string length while destroying order \citep{cilibrasi_clustering_2005,vitanyi_normalized_2009}.}

We evaluate the measure using a task-transfer design. \rev{As a first test, we measure correlated failure on closed-form benchmarks, as how often two models give the same wrong answer. Agentic and multi-agent settings, with tool use and reasoning traces, are left to future work.} Diversity is estimated from repeated responses by 38 language models to 100 open-ended prompts from the Infinity-Chat taxonomy \citep{jiang_artificial_2025}. We then test whether this independently derived population geometry predicts cross-task variation in correlated wrong answers among pairs in the same model population on ten disjoint closed-form benchmark families, controlling for semantic similarity and capability. \addtext{\rev{Finally, we} ask whether the pairwise signal survives at the level of five-model ensembles when semantic distance is held fixed by construction\rev{, and whether ensembles selected on the measure give more correct and fewer wrong answers under voting rules}.} We make three contributions:
\begin{enumerate}[leftmargin=*]
  \item \rev{Introduce inferred generative-process diversity, a model-free, parameter-free measure computed from outputs alone. It is the compression distance between two models' responses after removing what the same responses with shuffled bytes predict. Shuffling keeps length and byte frequencies and destroys order, so what remains reflects how the responses are ordered and structured.}
  \item \addtext{Show the permutation control residualisation separates an order-specific component from marginal output composition, that these components have opposing relationships with correlated failure, and that the order-specific component is related to, but non-redundant with, semantic similarity.}
  \item Demonstrate that inferred generative-process diversity predicts cross-task variation in correlated failure among evaluated model pairs beyond semantic similarity and capability. Under the cross-controlled specification, the association is negative on all ten benchmark families with a cross-benchmark mean of $-0.216$ and 95\% interval of $[-0.309,-0.122]$\rev{. It stays negative under two stronger embedding models and for three further constructions of the measure (Table~\ref{tab:main-measures})}. \addtext{At the ensemble level, five-model sets selected for generative-process diversity at a fixed level of semantic distance show lower chance-corrected convergent failure than capability-matched sets \revr{in every band of semantic distance}, whereas the reverse selection does not.} \rev{Ensembles selected on the measure also give more correct and fewer wrong answers under majority vote (Figure~\ref{fig:rev-vote-rules}).}
\end{enumerate}

\section{\rev{Measuring IGP Diversity and Correlated Failure}}
\label{sec:method}

\subsection{Problem formulation and compression distance}

Let $m_i$, $i\in\{1,\ldots,M\}$, be a black-box deployed model configuration \rev{(a model with its prompting and sampling settings; $M=38$)} and $x_{iqr}$ its $r$-th response to prompt $q$. We seek a pairwise statistic $d(i,j)$ that compares deployed configurations through strings generated under matched prompting conditions. Because finite observations do not identify a unique mechanism, ``inferred generative-process diversity'' denotes an output-derived relation among deployed configurations, not recovered neural computation. It can reflect persistent regularities introduced by model weights, post-training, prompting, decoding, or their interaction.

For \rev{two responses $x$ and $y$,} compressed length $C(\cdot)$ \rev{in bytes} and concatenation $xy$, NCD is
\begin{equation}
\operatorname{NCD}_{C}(x,y)=
\frac{C(xy)-\min\{C(x),C(y)\}}
{\max\{C(x),C(y)\}}.
\label{eq:ncd}
\end{equation}
\addtext{NCD is the $p=\infty$ member of a family $V_p^C$ that combines the two directional compression increments\rev{, $C(xy)-C(x)$ and $C(xy)-C(y)$}, retaining only the larger one. We use $V_1^C$ and $V_2^C$, which also retain the smaller increment, to test whether results depend on this aggregation choice; Appendix~\ref{app:compression} gives the construction and its theoretical properties.}
Lower values indicate more shared compressible structure\rev{, which means that the two responses compress better together than apart}. Reported distances use PPMd variant I\rev{, a compressor that predicts each byte from the bytes before it,} through \texttt{pyppmd} at library-default memory settings \citep{miura_miurahrpyppmd_2026}. Responses are literal UTF-8 byte strings. No token clustering, language-model symbolisation, or other learned representation enters the compression path. For each prompt and unordered model pair, we average position-paired response distances over $\min(R_i,R_j,50)$ pairs\rev{, where $R_i$ is the number of responses of model $i$ to the prompt}, then average prompts with equal weight. The permutation residualisation below is applied separately to each member of the family.

\subsection{Response corpus and permutation residual}
\label{sec:permutation-residual}

We estimate the compression and semantic predictors from a fixed corpus of repeated responses to Infinity-Chat prompts, generated independently of the benchmark outcomes. \rev{The two sets of texts are separate. The models are the same by design, since the question is whether one set predicts the other, and the capability controls come from benchmark correctness.} Appendix~\ref{app:response-corpus} gives the corpus composition, sampling settings, analysed text field, and filtering procedure.

Raw NCD responds to \rev{response length and byte composition} as well as order. For every response, we generate $P=20$ deterministic random byte permutations. Each surrogate preserves length and the exact byte multiset while destroying the original order. Let $d^{\mathrm{NCD}}_{ijq}$ denote the mean raw NCD for model pair $(i,j)$ on prompt $q$, and let $d^{\mathrm{perm}}_{ijq}$ denote the corresponding mean after applying the same distance and aggregation procedure to the permuted responses\rev{, which we call the permutation control}. For each prompt, we regress the raw distances on their permutation controls across the $\binom{M}{2}$ unordered model pairs as follows.
\begin{equation}
\begin{aligned}
(\widehat{\alpha}_q,\widehat{\beta}_q)
  &=\operatorname*{arg\,min}_{a,b}\sum_{1\leq i<j\leq M}
    \left(d^{\mathrm{NCD}}_{ijq}-a-bd^{\mathrm{perm}}_{ijq}\right)^2,\\
\widehat{\varepsilon}_{ijq}
  &=d^{\mathrm{NCD}}_{ijq}
    -\left(\widehat{\alpha}_q+\widehat{\beta}_q
    d^{\mathrm{perm}}_{ijq}\right),
\qquad
\spop_{ij}=\frac{1}{Q}\sum_{q=1}^{Q}\widehat{\varepsilon}_{ijq}.
\end{aligned}
\label{eq:resid}
\end{equation}
The projection separates the raw distance into a component explained by the \rev{permutation} control and an order-specific remainder, averaged with equal weight across \rev{the $Q=100$} prompts. \rev{The remainder is the part of NCD that response length and byte composition do not predict. A positive value means that two models' responses are farther apart than these predict.} The residual does not remove every low-order statistic or isolate the mechanism underlying the association. \rev{We call $\spop$ the \emph{population IGP score}, our measure of compression diversity. It is signed and is not a distance (a negative value means that two models are closer than their response length and byte composition predict), whereas the IGP distances defined below are metrics.}

\rev{Because Equation~\ref{eq:resid} is fitted across all pairs, the population IGP score mixes properties of single models, such as response length and style, with properties of particular pairs. Any pair quantity can be written $x_{ij}=\mu+g_i+g_j+e_{ij}$, with model effects $g_i$ and a pair-specific part $e_{ij}$. The \emph{pair-specific IGP score} $\spair$ removes the model effects from the raw and the permuted NCD within each prompt before the residualisation (Appendix~\ref{app:pair-specific}). The \emph{IGP distance} turns either score into a distance, $\sqrt{2-2K_{ij}}$, between two models' feature vectors in the ensemble kernel $K$ that Section~\ref{sec:ensemble-method} uses for the Vendi score (Appendix~\ref{app:igp-distance}). It is the quantity that ensemble selection uses, so the pairwise analysis can test it directly against semantic distance. This gives a population and a pair-specific IGP distance, $\dpop$ and $\dpair$. The population IGP score, which the main figures use, is the most conservative of the four, because differences between single models dilute it.}

\subsection{Semantic baseline}
\label{sec:semantic}

Each response is embedded with \texttt{all-MiniLM-L6-v2} and $\ell_2$-normalised \citep{reimers_sentencebert_2019}. Semantic similarity is the mean cross-response cosine similarity for a model pair within a prompt, averaged over prompts; semantic distance is $\SemD=1-\text{similarity}$. This follows the embedding-cosine construction used to study the Artificial Hivemind on the same prompt taxonomy \citep{jiang_artificial_2025}. To complete the comparison with the Artificial Hivemind analysis, Appendix~\ref{app:within-model} also characterises variation among repeated responses from the same model under the common prompting and sampling policy. \rev{To test how far the results depend on the embedding, we repeat the analyses with two stronger embedding models, text-embedding-3-small (TE3-small) and Qwen3-Embedding (Qwen3) (Appendix~\ref{app:embeddings}).}

\subsection{Epoch-native correlated failure}
\label{sec:cwa}

The outcome panel comprises ten benchmark families: TruthfulQA, MMLU-Pro, WorldSense, BBEH-mini, GSM8K, AIME, MuSR, GPQA-Diamond, Humanity's Last Exam, and AGIEval \citep{lin_truthfulqa_2022,cobbe_gsm8k_2021,rein_gpqa_2024,wang_mmlupro_2024,sprague_musr_2024,zhong_agieval_2023,phan_hle_2025,kazemi_bbeh_2025,benchekroun_worldsense_2023}. Each model answers every available item in five independently sampled epochs under the same generation policy. Multiple-choice letters and numeric answers are normalised by benchmark-specific evaluators; unparseable outputs are tracked separately.

For question $q$, let $c_{iq}(a)$ count parsed epochs in which model $i$ gives answer $a$, let $n_{iq}=\sum_a c_{iq}(a)$, and let $y_q$ be the reference answer. Pairwise correlated wrong-answer agreement pools epoch cross-products as follows.
\begin{equation}
\CWA_{ij}=
\frac{\sum_q\sum_{a\neq y_q}c_{iq}(a)c_{jq}(a)}
{\sum_q\left[n_{iq}n_{jq}-c_{iq}(y_q)c_{jq}(y_q)\right]}.
\label{eq:cwa}
\end{equation}
The numerator counts pairings in which both models are wrong with the same answer. The denominator counts every pairing in which at least one model is wrong; it excludes only both-correct pairings. Crossing five epochs per model yields up to 25 pairings per item and retains stochastic answer structure that modal aggregation discards \rev{(illustrated in Figure~\ref{fig:cwa})}.

Raw agreement depends on the answer space and on how strongly an item's distractors attract the model population. The baseline used in the primary analysis is therefore estimated from the remaining $M-2$ models separately for each pair under test. Let $p_i$ and $p_j$ denote the pair's respective wrong-answer rates, and let $\bar h_{-ij}$ denote the mean question-specific probability that a pair of wrong epochs sampled from the remaining models select the same answer. The expected agreement under independence and its chance-corrected form are
\begin{equation}
\CWA^{\mathrm{exp}}_{ij}
=\frac{p_i p_j}{p_i+p_j-p_i p_j}\bar h_{-ij},
\qquad
\kappa_{ij}=\frac{\CWA_{ij}-\CWA^{\mathrm{exp}}_{ij}}
{1-\CWA^{\mathrm{exp}}_{ij}}.
\label{eq:kappa}
\end{equation}
The first factor in $\CWA^{\mathrm{exp}}_{ij}$ is the probability that both models are wrong conditional on at least one being wrong. \rev{A $\kappa$ of zero means that a pair shares wrong answers as often as chance predicts, and a positive value more often.} This per-question-to-pair procedure is used throughout. Appendix~\ref{app:outcomes} gives the full construction \rev{and two sensitivity analyses, a pair-empirical marginal baseline and a chance term weighted like the observed term}.

\rev{CWA conditions on at least one model being wrong because, for a pair, the relevant question is whether the second model exposes the first one's errors: a question on which one model is wrong and the other is right lowers CWA. CWA factorises exactly as $\CWA=J\times E$. Joint failure $J$ is the probability that both models are wrong given that at least one is, and error agreement $E$ is the probability that they give the same wrong answer given that both are wrong (Appendix~\ref{app:joint-agreement}).}

\subsection{Association and uncertainty}
\label{sec:association}

For each benchmark, the primary statistic is the partial Spearman correlation between $\spop$ and $\kappa$, controlling for semantic distance, pair capability level $(a_i+a_j)/2$, and capability gap $|a_i-a_j|$\rev{, where $a_i$ is model $i$'s accuracy on that benchmark. A negative value means that more diverse pairs share fewer wrong answers. The controls follow from the question. We ask whether diversity in how models produce their outputs predicts shared wrong answers beyond diversity in what the outputs mean, so semantic distance is held fixed. Capability relates both to the measure and to shared wrong answers, so the level of the pair and the gap between its members are held fixed too; without them, the estimate would mix these confounds with the source of diversity we want to isolate (Appendix~\ref{app:estimand}). Ranks accommodate capability relationships that are monotone but not linear.}

\rev{Each model appears in 37 of the 703 pairs, so the pairs are not independent \citep{aronow_cluster_2015}. We therefore resample models rather than pairs and report percentile intervals from 2,000 model-node bootstrap replicates (Appendix~\ref{app:node-bootstrap}). Each benchmark is a separate test on new questions, so} cross-benchmark summaries weight the ten benchmark families equally and report a $t$ interval over the benchmark-level estimates.

\rev{We test the pair-specific measures against member-adjusted CWA, the part of $\kappa$ that each model's typical value does not explain, so that the measure and the outcome both exclude single-model effects. Semantic distance and the controls are adjusted the same way. Capability level, a sum of single-model terms, then drops out, while the gap remains. Capability can still act through the combination of the two models. Under independence, both are wrong with probability $(1-a_i)(1-a_j)=1-a_i-a_j+a_ia_j$, and the accuracy product $a_ia_j$ is the part the adjustment leaves, so we add it as a control (Appendix~\ref{app:pair-specific}).} \addtext{Ensemble-level \rev{summaries} are $t$-intervals over nine source benchmarks; within-benchmark intervals resample selected ensembles and are descriptive only (Appendix~\ref{app:ensemble-inference}).}

\subsection{Ensemble selection at fixed semantic distance}
\label{sec:ensemble-method}

\begin{addblock}
For an ensemble $S$ of $k=5$ models, convergent failure is the event that every member is wrong and all give the same wrong answer. With all five epochs per member retained, let $A_q$ count the epoch tuples on item $q$ in which every member is wrong and $C_q$ those that additionally agree (Figure~\ref{fig:cwa}). Chance is estimated per item from the wrong epochs of the other $M-k$ models, $N_q$ in total with $n_x$ on wrong answer $x$, as the hypergeometric probability that $k$ draws coincide. The conditional convergent-failure rate (CFR), its expectation, and the chance-corrected form are
\begin{equation}
\CFR_S=\frac{\sum_q C_q}{\sum_q A_q},
\quad
\CFR^{\mathrm{exp}}_S=\frac{\sum_q A_q\,b_q}{\sum_q A_q},
\quad
b_q=\frac{\sum_x\binom{n_x}{k}}{\binom{N_q}{k}},
\quad
\kappa_S=\frac{\CFR_S-\CFR^{\mathrm{exp}}_S}{1-\CFR^{\mathrm{exp}}_S}.
\label{eq:cfr}
\end{equation}
Observed and expected rates share the same items and tuple weights (Appendix~\ref{app:ensemble-outcome}). \rev{CFR conditions on the worst case for an ensemble, in which every member is wrong. For two models it reduces to the error agreement $E$ of Section~\ref{sec:cwa}. Appendix~\ref{app:ensemble-pair-rule} repeats the ensemble analysis under the pairwise rule, with at least one member wrong.} We enumerate all $\binom{38}{5}=501{,}942$ ensembles per benchmark and keep those with at least 30 estimable items, which admits nine source benchmarks and excludes GSM8K, on which too few ensembles fail five ways.

Because \rev{the two measures are strongly collinear (Section~\ref{sec:population-structure})}, ranking ensembles on one measure largely ranks them on the other, so we decouple the two by construction. Each ensemble is scored on both measures by the Vendi score \citep{friedman_vendi_2023}, \rev{the effective number of distinct members of the ensemble, between 1 and 5. We compute it from the ensemble's $5\times5$ block of the ensemble kernel $K_{ij}=u_i^{\top}u_j$, where the unit feature vector $u_i$ is built from model $i$'s scores against all 38 models} (Appendix~\ref{app:ensemble-kernel}). We divide eligible ensembles into five percentile bands of one measure's mean pairwise \rev{value} and, within each band, take the top 200 by the other measure's Vendi score. Each selection is compared with \rev{capability-matched ensembles, which are} random ensembles from the same band \rev{with a similar held value,} mean member accuracy and mean pairwise accuracy gap \rev{(Appendix~\ref{app:ensemble-matching})}. The reported contrast is the selected-minus-control difference in $\kappa_S$, \revr{with each control shifted by regression to the selected ensemble's capability (Appendix~\ref{app:ensemble-inference}),} so negative values mean the selected ensembles converge on the same wrong answer less often at \revr{equal} capability and fixed level of the other measure. \rev{The} difference-form outcome and alternative selectors are in Appendix~\ref{app:ensemble-results}.
\end{addblock}

\rev{CFR conditions on every member being wrong, so we also score the answers that ensembles give. Under a $t$-of-5 rule an ensemble answers when at least $t$ members give the same answer and abstains otherwise. Majority vote is 3 of 5. Under plurality the most common answer wins. Per 100 questions we count correct answers, wrong answers and questions not answered. On every benchmark the selected ensembles are the 200 with the highest Vendi score on a measure among all 501,942, so selection uses the open-ended responses alone. Within each benchmark they are matched on capability using half of the questions and scored on the other half, over 20 splits. We report the difference from capability-matched ensembles, after a regression adjusts the matched ensembles to the selected ensembles' capability (Appendix~\ref{app:ensemble-votes}). The vote outcomes form two different families of tests, each with its own Holm correction (Appendix~\ref{app:vote-inference}): majority vote, which tests whether selection improves a voting ensemble, and abstention at 4 and 5 of 5, which tests what a stricter rule changes.}

\section{Experiments and Results}
\label{sec:results}

\subsection{Semantic and compression-based population structure}
\label{sec:population-structure}

\begin{figure}[!t]
  \centering
  \includegraphics[width=\linewidth]{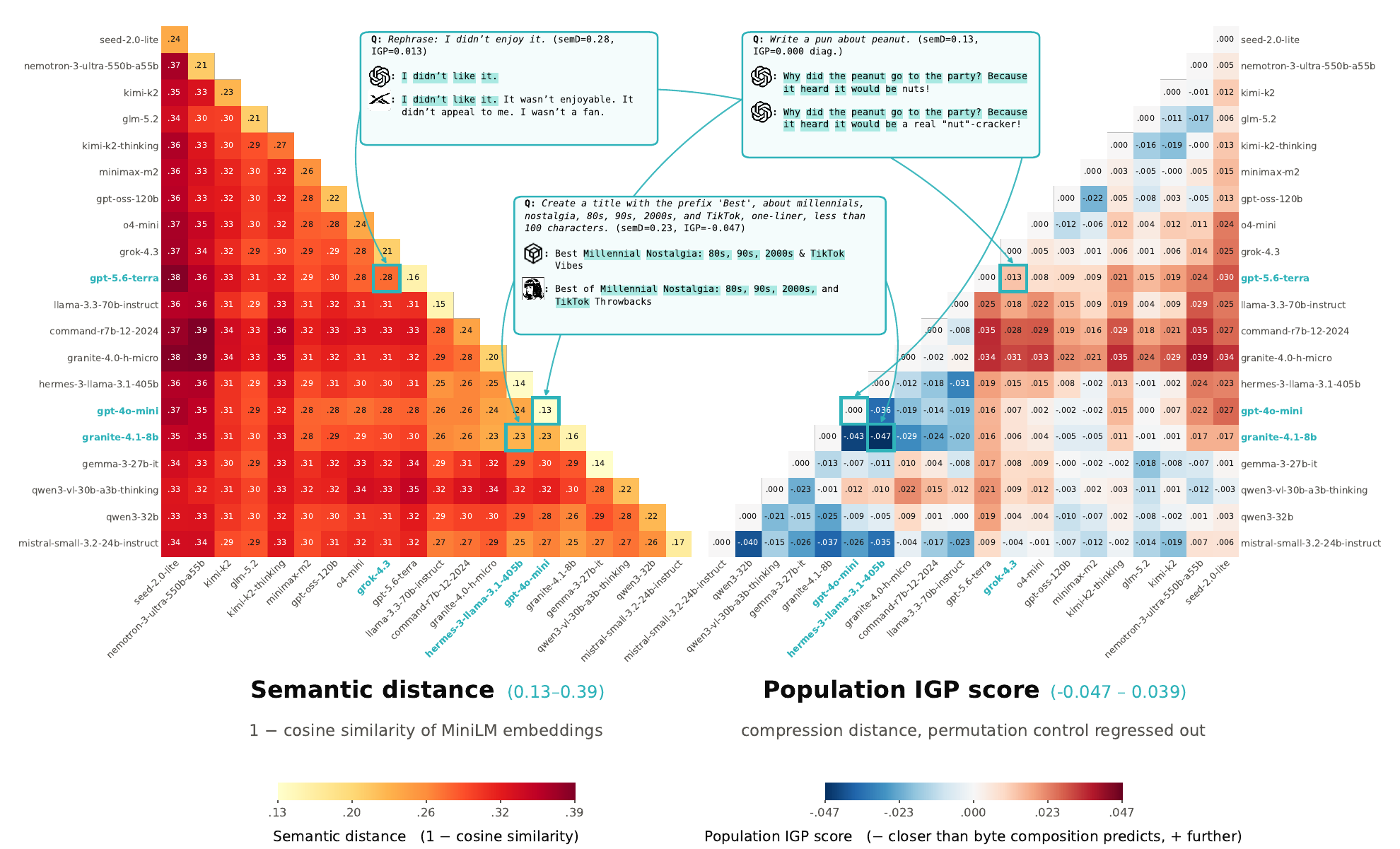}
  \caption{\addtext{\textbf{Semantic distance and the \rev{population IGP score} for the same responses.} Left: semantic distance. Right: the residual obtained by regressing raw NCD on the within-prompt byte-permutation control and averaging across prompts; positive values are pairs further apart in sequential organisation than byte composition predicts, negative values closer. 20 of 38 models are shown; reported associations use all 703 pairs. Annotated boxes contain response excerpts for a positive, near-zero, and negative residual.}}
  \label{fig:views}
\end{figure}

\begin{addblock}
Figure~\ref{fig:views} places semantic distance beside the \rev{population IGP score} for the same responses. Semantic distance records whether outputs mean the same thing and occupies a narrow band of low distance (similar meaning), $[0.159,0.388]$ across all 703 pairs. \rev{The score} records how outputs are sequentially organised, and the two disagree on individual pairs. Exemplar annotations show GPT-4o-mini--Granite-4.1 responses reuse one peanut-pun structure and sit at a near-zero \rev{score}. The Granite--Hermes and GPT-5.6--Grok pairs have raw NCD values of 0.675 and 0.678, indistinguishable within the observed range $[0.484,0.859]$ (Figure~\ref{fig:supp-views-stacked}), yet \rev{scores} of $-0.047$ and $+0.013$: two closely aligned one-line titles are closer than their byte composition predicts, and two responses that share meaning but differ in elaboration are further apart. Residualisation exposes this structure, but it does not simply extract what semantics miss. \rev{The score's} rank association with semantic distance is 0.789, higher than raw NCD's 0.662 (Figure~\ref{fig:supp-residualisation-colour}), so its non-redundant value must be tested against failure with semantic distance held fixed. Subsequent analyses use the \rev{population IGP score}, with positive values denoting pairs further apart than byte composition predicts.
\end{addblock}

\subsection{Conditional associations with correlated failure}
\label{sec:conditional}

\begin{figure}[!t]
  \centering
  \includegraphics[width=\linewidth]{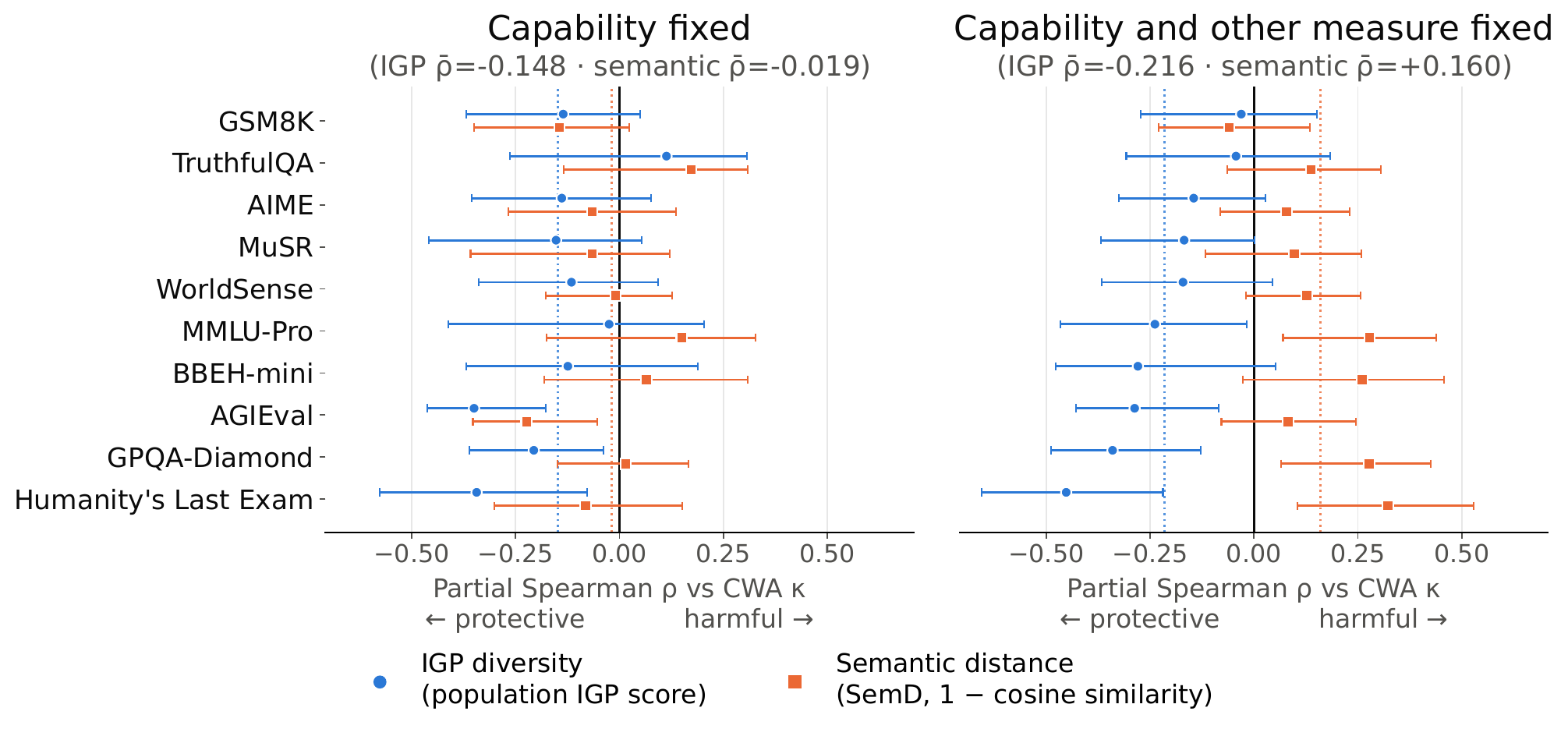}
  \caption{\textbf{Dissociation between \rev{IGP diversity} and semantic distance.} Left: each measure with capability level and gap fixed. Right: each measure with capability and its rival fixed. \rev{IGP diversity is the population IGP score.} Bars are model-level bootstrap intervals; headings and dotted lines give cross-benchmark means.}
  \label{fig:dissociation}
\end{figure}

Figure~\ref{fig:dissociation} compares the \rev{population IGP score} with semantic distance under matched control sets. With capability fixed, \rev{IGP diversity} is associated with less correlated failure ($-0.148$, $[-0.245,-0.050]$), while semantic distance is unresolved ($-0.019$, $[-0.108,+0.070]$). Holding the rival measure fixed strengthens the \rev{IGP} estimate to $-0.216$ $[-0.309,-0.122]$; the surviving component of semantic distance is $+0.160$ $[+0.074,+0.246]$. \rev{The association is modest in size and consistent in sign. It holds for both parts of CWA, $J$ and $E$, and for ensembles scored under the pairwise rule (Appendices~\ref{app:joint-agreement} and~\ref{app:ensemble-pair-rule}).}

\rev{Only about 38\% of semantic rank variance remains after controlling for the population IGP score (Figure~\ref{fig:supp-residualisation-colour}).} The positive cross-controlled semantic estimate therefore describes this remaining component, not the standalone association of semantic distance with correlated failure. Under cross-control, the \rev{IGP} estimate is negative on all ten benchmarks and its model-node bootstrap interval excludes zero on four, with the largest effects on Humanity's Last Exam ($-0.452$) and GPQA-Diamond ($-0.340$). The semantic estimate is positive on nine benchmarks. GSM8K is the exception at $-0.060$, with its interval spanning zero. The benchmark-paired difference between the two cross-controlled partial correlations has mean $-0.375$ and 95\% $t$-interval $[-0.545,-0.206]$; Appendix~\ref{app:paired-difference} gives its construction and benchmark-level estimates. \addtext{Binning pairs into percentile bands gives the same picture (Appendix~\ref{app:dose}).} \rev{With capability and the other measure held fixed, CWA $\kappa$ falls across bands of the population IGP score and not across bands of semantic distance, under all three embeddings.}

\rev{The association belongs to pairs, not to single models. Table~\ref{tab:main-measures} asks whether the association reflects what each model is on its own or how two models relate. The pair-specific estimates remove each model's own level from the measure and from the outcome. They are larger than the population estimates and do not shrink under the stronger embeddings, so neither differences between single models nor a weak semantic baseline account for the association. All twelve estimates are negative and every interval excludes zero. Semantic distance adds nothing beyond either IGP distance and keeps a small negative association beyond the pair-specific IGP score ($-0.063$ to $-0.105$; Appendix~\ref{app:embeddings}). With other compressors, the population IGP score stays negative under CTW and weakens under bzip2 and gzip, which model less of the sequential structure of a response (Appendix~\ref{app:compressor-failure}).}

\begin{table}[t]
  \caption{\rev{\textbf{The four IGP measures against CWA $\kappa$ (PPMd).} Mean partial Spearman correlation over ten benchmarks, controlling capability and semantic distance, with 95\% interval and number of negative benchmarks. Pair-specific rows use member-adjusted CWA with the accuracy-product control (Appendix~\ref{app:pair-specific}).}}
  \label{tab:main-measures}
  \centering
  \rev{\resizebox{\linewidth}{!}{%
  \begin{tabular}{llll}
    \toprule
    IGP measure & MiniLM & TE3-small & Qwen3 \\
    \midrule
    Population score ($\spop$) & $-0.216$ $[-0.309,-0.122]$ 10/10 & $-0.128$ $[-0.207,-0.048]$ 9/10 & $-0.106$ $[-0.199,-0.013]$ 8/10 \\
    Population distance ($\dpop$) & $-0.227$ $[-0.297,-0.158]$ 10/10 & $-0.195$ $[-0.252,-0.139]$ 10/10 & $-0.188$ $[-0.252,-0.123]$ 10/10 \\
    Pair-specific score ($\spair$) & $-0.257$ $[-0.367,-0.147]$ 9/10 & $-0.279$ $[-0.400,-0.157]$ 9/10 & $-0.248$ $[-0.353,-0.144]$ 10/10 \\
    Pair-specific distance ($\dpair$) & $-0.317$ $[-0.465,-0.169]$ 9/10 & $-0.338$ $[-0.493,-0.182]$ 9/10 & $-0.305$ $[-0.444,-0.165]$ 9/10 \\
    \bottomrule
  \end{tabular}}}
\end{table}

\begin{figure}[!t]
  \centering
  \includegraphics[width=0.72\linewidth]{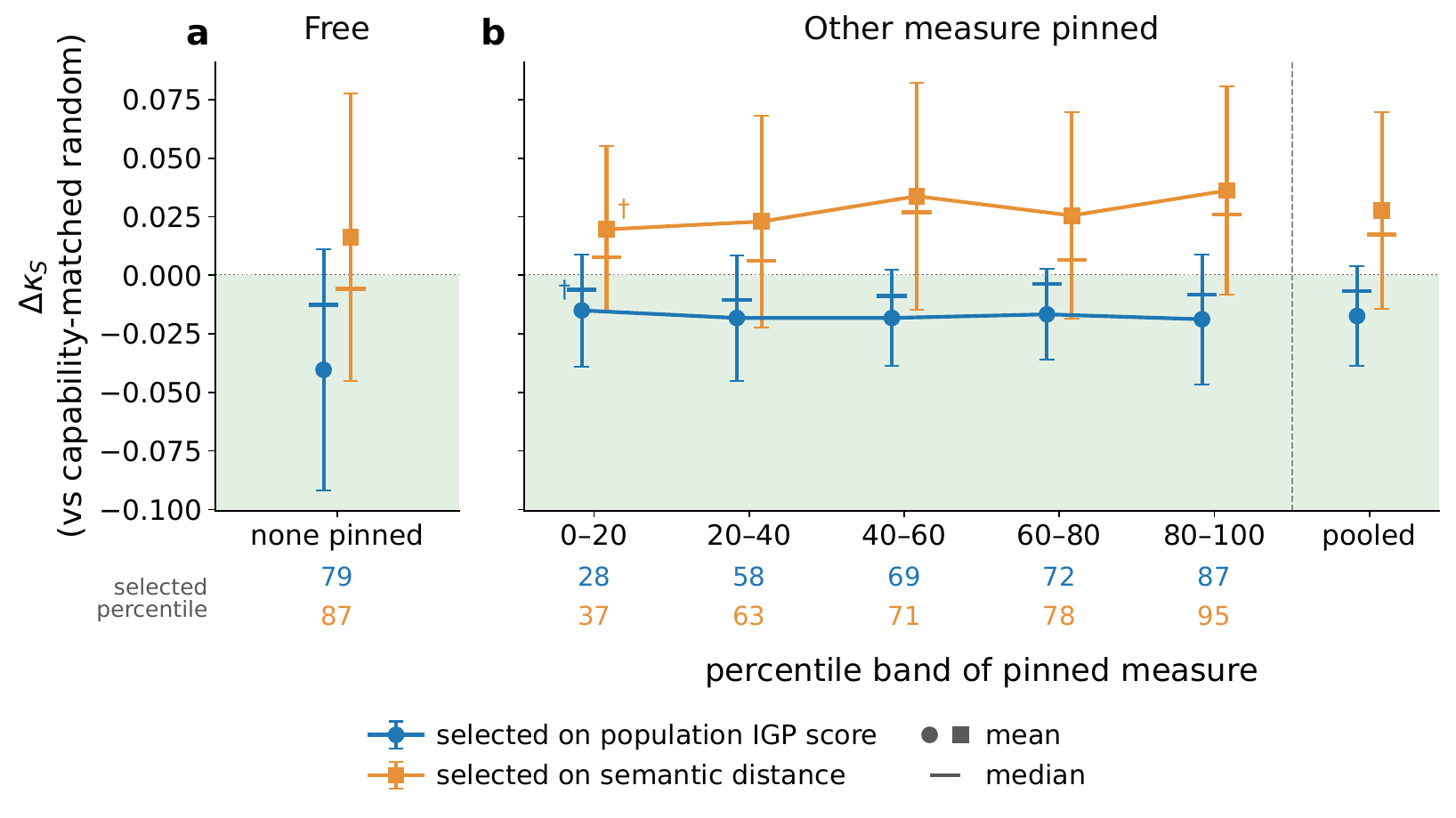}
  \caption{\addtext{\textbf{Vendi selection free and pinned to a percentile band of the other \rev{measure}.} Each point is the difference in $\kappa_S$ (Equation~\ref{eq:cfr}) between the top 200 five-model ensembles by Vendi score of one measure and capability-matched random ensembles\revr{, after the capability adjustment of Appendix~\ref{app:ensemble-inference}}, averaged over nine  benchmarks. Bars are 95\% $t$-intervals and horizontal marks are medians. Coloured numerals are the median percentile of selected ensembles on selected measure. Daggers mark bands with at least one selection with $<$ 50 controls.} \rev{The pooled marker averages the five band differences within each benchmark before averaging over benchmarks (Appendix~\ref{app:ensemble-pooled}).}}
  \label{fig:ensemble}
\end{figure}

\subsection{\rev{IGP-selected} ensembles fail together less often}
\label{sec:results-ensemble}
\rev{Figure~\ref{fig:ensemble} compares selected ensembles with capability-matched random ensembles. Selected freely, ensembles with the highest IGP Vendi score converge on the same wrong answer less often than matched ensembles \revr{($\Delta\kappa_S=-0.040$ $[-0.092,+0.011]$)}, although WorldSense and Humanity's Last Exam dominate this arm (Appendix~\ref{app:ensemble-results}). With semantic distance held within a band, IGP selection stays below matched ensembles in every band, and pooled across bands the estimate is \revr{$-0.017$ $[-0.039,+0.004]$} (Figure~\ref{fig:ensemble}b)\revr{, and $-0.008$ and $-0.009$ under the two stronger embeddings (Appendix~\ref{app:ensemble-pooled}). The selected ensembles also fail together less often. Under free selection all five members are wrong on 0.39 fewer questions per 100, $[-0.72,-0.05]$}. The reverse selection, on semantic distance with the IGP score held, lies above matched ensembles in every band. \rev{The IGP score computed from} $V_1$ and $V_2$ \rev{gives} the same pattern (Appendix~\ref{app:ensemble-figures}).}

\begin{figure}[!t]
  \centering
  \includegraphics[width=0.9\linewidth]{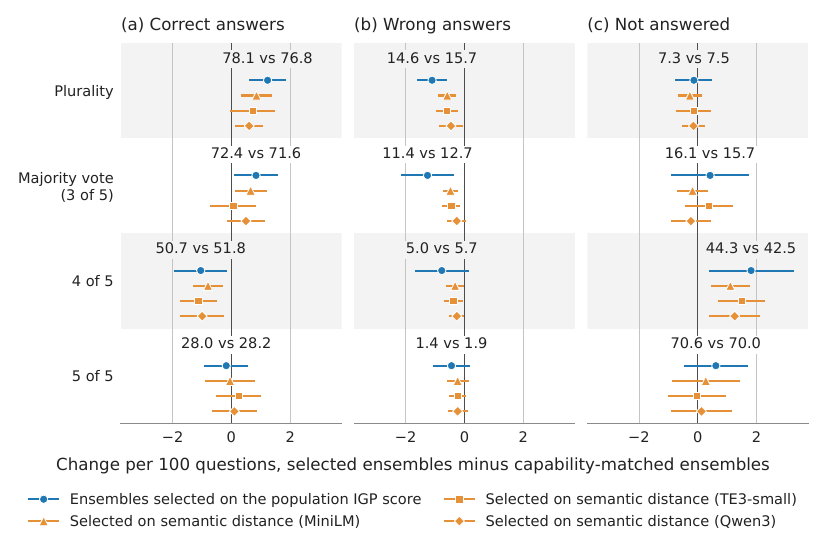}
  \caption{\revcolor{}\textbf{Selection on diversity under four voting rules.} Each point is the mean over ten benchmarks of the change per 100 questions in (a) correct answers, (b) wrong answers and (c) questions not answered: the 200 five-model ensembles with the highest Vendi score on a measure minus capability-matched ensembles, after the regression adjustment of Section~\ref{sec:ensemble-method}. Bars are 95\% $t$-intervals over benchmarks. Numbers above each blue point (selection on the population IGP score) are the rates of the selected and of the matched ensembles.}
  \label{fig:rev-vote-rules}
\end{figure}

\rev{Under majority vote, selection removes about one wrong answer in ten (Figure~\ref{fig:rev-vote-rules}). The 200 ensembles with the highest Vendi score on the population IGP score give 11.4 wrong answers per 100 questions against 12.7 for capability-matched ensembles ($-1.26$ $[-2.17,-0.35]$) and 72.4 correct answers against 71.6 ($+0.84$ $[+0.10,+1.57]$). Both differences survive the Holm correction (Appendix~\ref{app:vote-inference}). The members of the selected ensembles are as accurate as those of the matched ensembles (Appendix~\ref{app:vote-members}), which is consistent with Figure~\ref{fig:ensemble}: when the members' wrong answers agree less often, a wrong majority forms less often. The reduction holds at the same coverage, with semantic distance held within a band, and under plurality (Appendices~\ref{app:vote-coverage} and~\ref{app:vote-bands}). A stricter rule turns the same disagreement into abstention. Under 4 of 5 the selected ensembles answer fewer questions and are right as often when they answer, so their wrong answers are not lower at the same coverage. Selection on semantic distance removes fewer wrong answers, and selection on the pair-specific IGP score acts in the same direction (Appendix~\ref{app:vote-pair-specific}).}

\section{Discussion}
\label{sec:discussion}

\begin{addblock}
Estimated from single-turn responses to open-ended chat prompts \citep{jiang_artificial_2025}, \rev{inferred generative-process diversity} predicts which model pairs choose the same wrong answer on \rev{ten} benchmarks beyond semantic similarity and capability (Figure~\ref{fig:dissociation})\rev{, which} points toward it capturing a persistent property of generation rather than of the specific tasks. \rev{Ensembles selected on it at fixed semantic distance converge on one wrong answer less often than capability-matched ensembles, whereas the reverse selection does not (Figure~\ref{fig:ensemble}). The association is larger once each model's own level is removed (Table~\ref{tab:main-measures}), so this diversity is a relation between models and cannot be read from the properties of any one of them. Its benefit depends on the decision rule (Figure~\ref{fig:rev-vote-rules}): selection removes about one wrong answer in ten under majority vote, which can outvote a shared error, and costs coverage under rules that require near-unanimity.}

\rev{The IGP score and the IGP distance answer different questions. The score compares two models directly. The distance compares how each relates to the whole fleet, so two models are close when their scores against all other models agree. It is therefore a metric and the natural input to ensemble selection through the Vendi kernel, and it gives larger estimates that vary less across embeddings (Table~\ref{tab:main-measures}), at the cost of depending more on which other models are present. We would use the score to audit a pair and the distance to choose an ensemble.}

Residualisation of NCD against the byte-permutation control appears critical to isolating failure-relevant signal. Raw NCD is weakly associated with more correlated failure, \rev{the permutation control} more strongly, and only the \rev{population IGP score} reverses this association (Figure~\ref{fig:components}). This separation parallels the distinction between statistical and algorithmic structure \citep{zenil_low-algorithmic-complexity_2017}, with \rev{length and byte composition} describing the former and shared sequential organisation the latter. One possible explanation is that \rev{byte composition reflects} vocabulary and formatting conventions that models share for reasons unrelated to how they solve problems, and that many distinct processes can produce the same \rev{composition}.

\rev{On its own, with capability controlled, semantic distance predicts correlated failure weakly or not at all: pairs that differ more in meaning share about as many wrong answers under MiniLM and slightly fewer under the two stronger embeddings. Once the IGP score is also held fixed, semantic distance no longer predicts fewer shared wrong answers, and under MiniLM it predicts more (Figure~\ref{fig:dissociation}, Appendix~\ref{app:embeddings}).} This is consistent with semantic distance carrying failure-relevant information only through the variation it shares with the \rev{score, so} assessments of model homogeneity based on output meaning \citep{jiang_artificial_2025} do not by themselves capture the variation associated with independent failure. \rev{Selecting ensembles on semantic diversity did not increase convergent failure in this fleet, and under majority vote it removed some wrong answers, though fewer than selection on the IGP score. That benefit appears to come from the overlap between the two measures. With the IGP score held within a band, semantic selection converges on the same wrong answer slightly more often than matched ensembles in every band (Figure~\ref{fig:ensemble}b). In fleets where meaning and process diverge more, selecting on meaning alone could therefore assemble ensembles that fail together.}

The association between inferred generative-process diversity and correlated failure is modest partly because there is little diversity to exploit. \rev{Across 20 providers, semantic} distance occupies a narrow range (Figure~\ref{fig:views}), and the process diversity within it is modest. Our intuition is that both reflect convergence in training data, architectures, pipelines and potentially distillation between models. If the resilience framing is taken seriously, this narrowness is itself a risk. If we wish to exploit diversity to improve the performance and resilience of collective systems \citep{galesic_beyond_2023,tieman_landscape_2025,holling_resilience_1973,scheffer_anticipating_2012}, a greater number of diverse models are required, along with measures that assess diversity in failure-relevant terms.

\rev{Several} limitations remain. Generation cost constrained sampling depth and benchmark breadth, so failure outcomes rest on five epochs per benchmark, and several benchmarks were saturated for the stronger models, which limits usable CWA and CFR and attenuates associations toward zero. \rev{Our results concern ensemble reliability on closed-form benchmarks. Whether the association carries over to multi-agent systems is a hypothesis for future work. There a model's outputs return as its inputs, through code and through messages between agents, so output biases that models share can compound.} \rev{Next} steps are to apply the measure to action--observation and tool-use traces, treating model, harness, and tools as one input--output unit, and to use the eigenstructure of the ensemble kernel rather than a single Vendi score to examine how diversity is distributed across an ensemble.
\end{addblock}

\section{Conclusion}
\label{sec:conclusion}
Multi-model systems rely on redundancy for robustness, and ecological resilience research indicates that diversity must be judged against the function and failure modes of the system. We introduced inferred generative-process diversity, estimated as the residual of NCD after a byte-permutation control. Across 38 models it predicts correlated failure on ten benchmarks beyond semantic distance and capability, and ensembles selected on it at fixed semantic distance converge on the same wrong answer less often than capability-matched ensembles. \rev{Under majority vote, ensembles selected on it give more correct and fewer wrong answers.} \rev{As systems combine more models, what matters is not how many models they contain but whether those models fail independently. In the narrow fleet we studied, models that produce their outputs in the same way tended to be wrong in the same way. Measuring that sameness from outputs alone, model-free and parameter-free, is a first step toward collective systems whose redundancy is real.}

\subsection*{Reproducibility statement}
Section~\ref{sec:method} defines every statistic used in the main text, and the appendices give additional detail required to recompute them. Appendix~\ref{app:data} describes the response corpus, sampling settings, filtering, benchmark panel, question caps, and pair estimability. Appendix~\ref{app:compression} gives details on the compressor, the permutation residualisation, and robustness checks over compressors, permutation counts, and aggregation. Appendix~\ref{app:outcomes} constructs the pairwise outcome and its chance baselines. Appendix~\ref{app:ensemble} specifies ensemble enumeration, the ensemble outcome, the kernel and Vendi score, pin bands, capability matching, inference, and full results tables and figures. Appendix~\ref{app:statistics} details the partial rank correlations, model-node bootstrap, and paired comparisons. \rev{Appendices~\ref{app:igp-measures} to~\ref{app:phylolm} define the four IGP measures and report the analysis under three embeddings and against PhyloLM. Appendix~\ref{app:ensemble-votes} gives the design and the results of the vote and abstention outcomes.} All generation used public model endpoints under a fixed sampling policy, and benchmark evaluation used the Inspect AI framework. 

\subsection*{AI use statement}
In this work, we used generative AI tools to provide feedback on research methodology and experiments, implement methods, formulate mathematical claims, and interpret results. We have not used generative AI tools to generate synthetic data sets (the analysed model outputs are the object of study), help develop theoretical models or conceptual frameworks, or propose or refine hypotheses, and assistance with translation and support for qualitative and thematic data analysis are not applicable to this work. Additionally, we used generative AI tools to create or edit code, create or modify scientific figures, assist in drafting parts of the paper, edit the paper to improve readability, and suggest a structure for parts of the paper. We have reviewed all AI-assisted work. The authors reviewed and tested all AI-assisted code and checked reported numbers against the analysis outputs. We take responsibility for the final content of this work, including text, claims, or artifacts produced with the aid of generative AI.
\bibliographystyle{iclr2027_conference}
\bibliography{references}

\appendix

\begingroup
\renewcommand{\contentsname}{Appendices}%
\setcounter{tocdepth}{-1}%
\rev{\tableofcontents}
\endgroup
\setcounter{tocdepth}{3}
\addtocontents{toc}{\protect\setcounter{tocdepth}{2}}
\clearpage

\section{Data Generation}
\label{app:data}

Across both the predictor corpus and benchmark evaluations, 29 of the 38 deployed model configurations were served through OpenRouter. Four configurations used the OpenAI API directly, four used the Anthropic API directly, and one used the Google API directly.

\subsection{Response corpus}
\label{app:response-corpus}

The predictor corpus uses the 100 open-ended prompts from the Infinity-Chat prompts introduced for the Artificial Hivemind study \citep{jiang_artificial_2025}. For each prompt, we sampled 50 responses from each of 38 deployed model configurations, giving 190,000 generations. Sampling used temperature 1.0, top-$p$ 0.9, no minimum-$p$, and a 32,768-token output budget. Compression and semantic analyses use only the stored visible assistant response. The prompt, request metadata, and separately recorded reasoning fields do not enter either representation. To prevent language choice from dominating the byte- and embedding-based comparisons, we excluded responses containing more than 20 CJK ideographs. This removed 75 generations (0.04\%) and left 189,925 responses; per-prompt computations use the available responses in each model--prompt cell.

\subsection{Benchmark panel}

Benchmark generation and logging used the UK AI Security Institute's Inspect AI framework \citep{UK_AI_Security_Institute_Inspect_AI_Framework_2024}. Each selected question was evaluated in five independently sampled epochs under the shared generation policy, and the resulting Inspect logs were converted to a common per-model record format while retaining all five responses.

We distinguish task execution, answer-space-specific analysis, and statistical reporting. At execution time, one benchmark family may comprise several Inspect tasks. AGIEval and WorldSense each expand to six tasks, while MuSR and AIME each expand to three. At analysis time, BBEH-mini is divided into multiple-choice, numeric, and other short-answer keys because these answer spaces require different equivalence rules. This gives 12 matrix keys for the ten benchmark families. For reporting, the sufficient counts from the three BBEH-mini keys are pooled before CWA and its baseline are calculated, yielding one BBEH-mini estimate and preventing that benchmark from receiving three votes in cross-benchmark summaries. The reporting panel therefore contains the ten benchmark families used in the main text. BBH is not included because BBEH was designed as its harder successor.

Question caps were applied per Inspect task for API cost reduction. They were 800 for GSM8K, 750 for MMLU-Pro, 300 for HLE-MC, 150 for each WorldSense task, 117 for each MuSR domain, and 100 for each AGIEval task. AIME, BBEH-mini, GPQA-Diamond, and TruthfulQA used their full configured datasets.

\subsection{Model population and estimability}

The 38-model population gives $\binom{38}{2}=703$ candidate pairs. Under the requirement of at least 30 questions with a nonzero either-wrong denominator, 703 pairs are estimable on seven keys, 701 on WorldSense, 699 on GPQA-Diamond, and 562 on AIME. AIME is limited by its 90-question cap. Predictor and outcome texts are disjoint, but the same models generate both and capability controls come from the correctness matrices used for the outcome.

\section{Compression Distance}
\label{app:compression}

\subsection{Compressor choice and robustness}
\label{app:compressor-choice}

\rev{Unless stated otherwise, reported} distances use PPMd variant I through \texttt{pyppmd} at library-default memory settings. PPMd is a sequential statistical compressor. It maintains variable-order byte contexts, updates its conditional next-byte distribution as the stream is read, and encodes the resulting predictions with a range coder \citep{shkarin_ppm_2002}. Context Tree Weighting (CTW) is also sequential, recursively weighting bounded-memory tree sources \citep{willems_context-tree_1995}. \rev{Its binary-decomposition form codes each byte as a sequence of binary decisions, each with its own context tree \citep{tjalkens_context-tree_1997,begleiter_superior_2006,franken_overview_2002}.} By contrast, gzip and LZMA belong to the Lempel--Ziv dictionary family, whose reuse of information across a concatenation is expressed principally through references to matching phrases \citep{ziv_universal_1977,ferragina_compressing_2010}. Gzip additionally has a finite 32-kB history window, which can make NCD depend on object length once relevant content falls outside that window \citep{cebrian_common_2005}. \rev{bzip2 is a block-sorting compressor. It applies the Burrows--Wheeler transform to each block and codes the result \citep{burrows_block-sorting_1994}.} PPM compressors have performed consistently among the strongest compressors in compression-based classification, whereas gzip provides a faster but sometimes less discriminative alternative \citep{ferragina_compression-based_2007}.

For an adaptive compressor, the information transferred across a concatenation is directional. After reading $x$, the incremental code length of $y$ is $C(xy)-C(x)$. We express its reduction relative to coding $y$ alone as
\begin{equation}
g^C_{x\rightarrow y}
=1-\frac{C(xy)-C(x)}{C(y)}
=\frac{C(x)+C(y)-C(xy)}{C(y)},
\qquad
T_C(x,y)=\frac{g^C_{x\rightarrow y}+g^C_{y\rightarrow x}}{2},
\label{eq:directional-transfer}
\end{equation}
where $g^C_{y\rightarrow x}$ is defined analogously from $C(yx)$. PPMd and CTW transfer an adaptive conditional context model from the first sequence to the second; the dictionary compressors transfer a phrase dictionary. Both mechanisms can therefore be directional, and $C(xy)$ need not equal $C(yx)$. The symmetric gain $T_C$ is used only for the diagnostic below; the reported NCD matrices retain the fixed orientation in Equation~\ref{eq:ncd}.

\begin{figure}[htbp]
  \centering
  \includegraphics[width=0.74\linewidth]{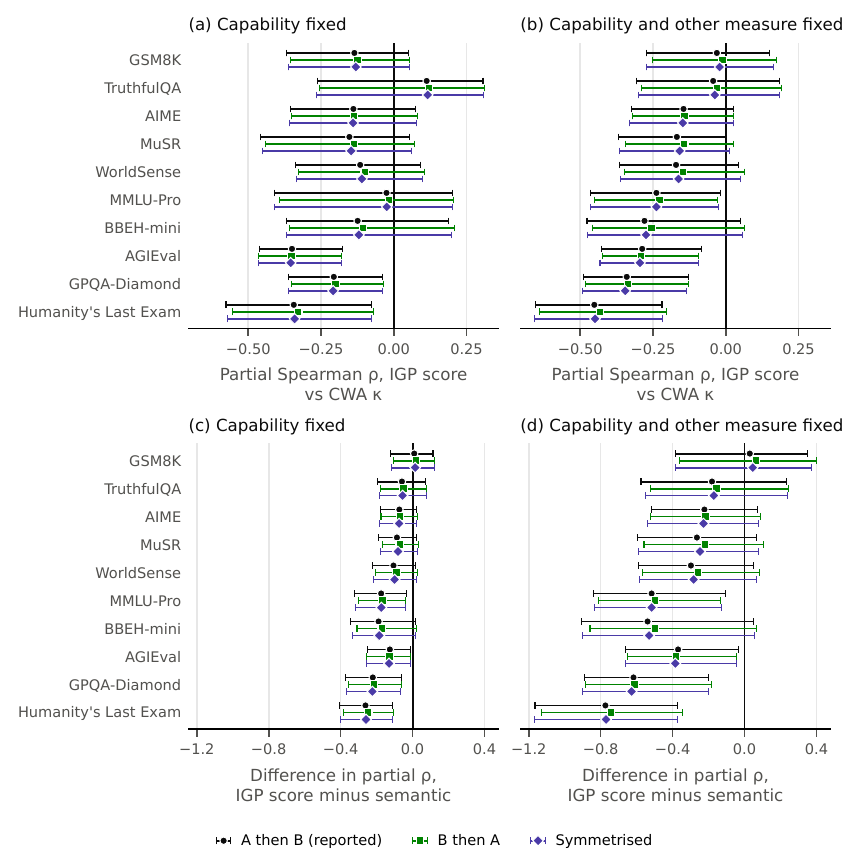}
  \caption{\revcolor{}\textbf{The dissociation under both concatenation orders and the symmetrised distance.} Each estimate is shown under the reported order (A then B), the reversed order (B then A) and the symmetrised distance, the mean of the two. (a, b) Partial Spearman correlation between the population IGP score and CWA $\kappa$, the IGP series of Figure~\ref{fig:dissociation}, with capability fixed (a) and with capability and semantic distance fixed (b). (c, d) The paired difference, IGP score minus semantic distance. Bars are 95\% model-node bootstrap intervals. No estimate changes sign between the three versions, and an interval that excludes zero under one excludes it under all three.}
  \label{fig:rev-order-dissociation}

  \vspace{0.2em}
  \includegraphics[width=0.76\linewidth]{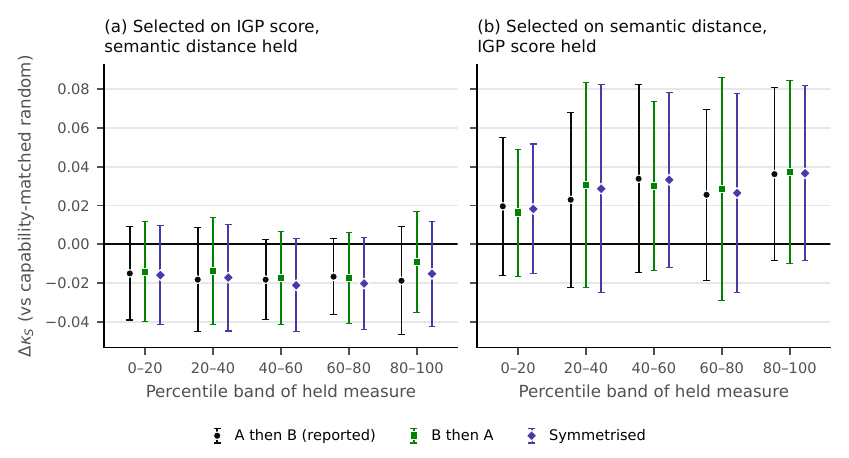}
  \caption{\revcolor{}\textbf{The ensemble result under both concatenation orders and the symmetrised distance.} Figure~\ref{fig:ensemble}b under the three versions, marked as in Figure~\ref{fig:rev-order-dissociation}. Each marker is $\Delta\kappa_S$, the difference in CFR $\kappa$ between the 200 ensembles with the highest Vendi score on one measure and capability-matched random ensembles\revr{, after the capability adjustment of Appendix~\ref{app:ensemble-inference}}, averaged over nine benchmarks, with the other measure held within one percentile band. Bars are 95\% $t$-intervals. (a) Ensembles selected on the population IGP score, semantic distance held. (b) Ensembles selected on semantic distance, population IGP score held. Every band mean keeps its sign under all three versions, and no band mean moves by more than 0.010.}
  \label{fig:rev-order-ensemble}
\end{figure}

\begin{revblock}
The fixed orientation does not drive the results. We recomputed every NCD, its permutation control and the within-model diagonal with the two responses concatenated in the reverse order, and formed the symmetrised distance, the mean of the two orders. Figures~\ref{fig:rev-order-dissociation} and~\ref{fig:rev-order-ensemble} repeat the main results under all three versions. With capability and semantic distance controlled, the mean partial correlation between the population IGP score and CWA $\kappa$ is $-0.216$ under the reported order, $-0.202$ under the reversed order and $-0.213$ for the symmetrised distance, and it is negative on all ten benchmarks under each. With semantic distance held within a band, ensembles selected on the IGP Vendi score stay below capability-matched ensembles in every band under all three versions.
\end{revblock}

We first compared PPMd with byte-level CTW (alphabet size 256 and depth 12), \rev{binary-decomposition CTW (depth 6, $\alpha=1/16$), bzip2,} LZMA, and gzip on raw model responses. Because CTW is costly on byte strings, this diagnostic uses a deliberately heavy subsample comprising 16 of the 38 models, 15 of the 100 prompts, and one response position per prompt shared across models. Responses are not truncated; four model--prompt responses above the 8-KiB CTW limit were excluded from every compressor. This leaves 1,746 pair--prompt observations and 120 model-pair means. Table~\ref{tab:compressor-robustness} reports agreement with the PPMd ordering and finite-length saturation.

\begin{table}[htbp]
  \caption{\textbf{Raw-response compressor robustness.} Rank agreement is Spearman's $\rho$ between the 120 model-pair mean distances under PPMd and each alternative compressor. The remaining columns use all 1,746 pair--prompt observations. Absolute NCD levels are compressor-specific and are not a common calibrated scale. \rev{CTW is the byte-level form at depth 12 and binary CTW the binary-decomposition form at depth 6 with $\alpha=1/16$.}}
  \label{tab:compressor-robustness}
  \centering
  \begin{tabular}{lrrr}
    \toprule
    Compressor & $\rho$ with PPMd & Mean NCD & NCD $\geq 0.95$ \\
    \midrule
    PPMd & --- & \rev{0.795} & \rev{3.0\%} \\
    CTW  & \rev{0.891} & \rev{0.934} & \rev{39.4\%} \\
    \rev{Binary CTW} & \rev{0.958} & \rev{0.843} & \rev{8.8\%} \\
    \rev{bzip2} & \rev{0.942} & \rev{0.737} & \rev{0.5\%} \\
    LZMA & 0.918 & \rev{0.686} & \rev{0.1\%} \\
    gzip & \rev{0.951} & \rev{0.779} & \rev{1.3\%} \\
    \bottomrule
  \end{tabular}
\end{table}

The model-pair ordering is therefore broadly preserved under all \rev{five} alternatives, most closely under \rev{binary CTW and} gzip. The mean distance itself is not a quality score. LZMA's lower mean, for example, reflects a different finite-length scale. Saturation is the relevant failure mode for discrimination. \rev{Byte-level} CTW places \rev{39.4\%} of observations at or above 0.95, compared with \rev{3.0\%} for PPMd \rev{and 8.8\% for binary CTW}, leaving much less variation among raw response pairs.

\begin{figure}[htbp]
  \centering
  \includegraphics[width=\linewidth]{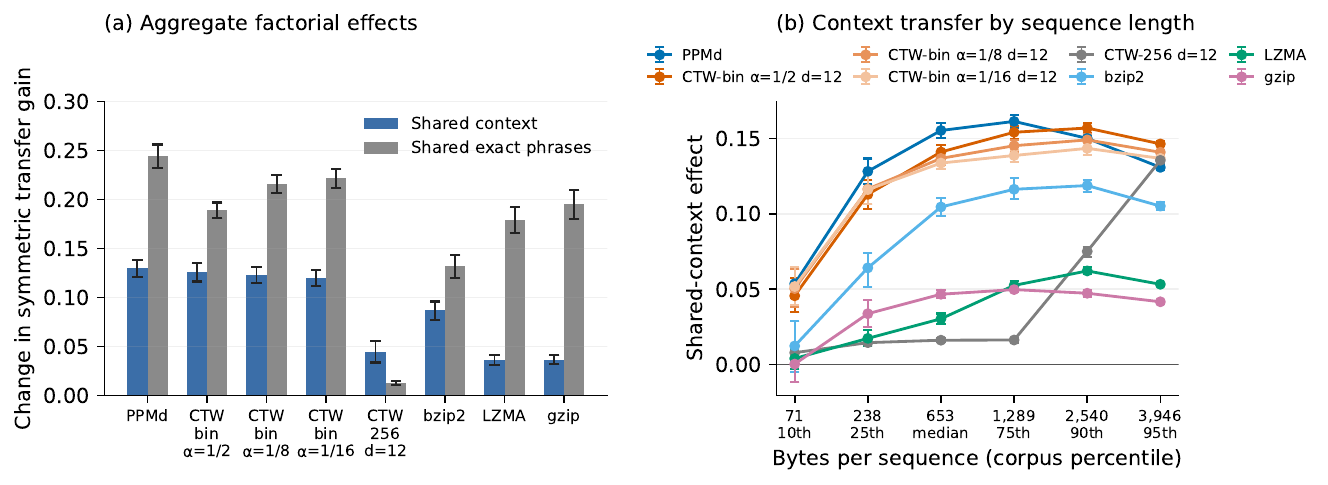}
  \caption{\textbf{Sequential-context and exact-phrase transfer.} (a) Main effects of sharing a first-order transition law or exact byte blocks on the symmetric transfer gain in Equation~\ref{eq:directional-transfer}\rev{, for PPMd, binary-decomposition CTW (CTW-bin) at depth 12 with $\alpha=1/2$, $1/8$ and $1/16$, byte-level CTW (CTW-256) at depth 12, bzip2, LZMA and gzip}. Bars are means over 72 paired units; intervals are 95\% bootstrap intervals. (b) The shared-context effect at six sequence lengths drawn from the empirical response-length distribution. Each of $x$ and $y$ has the displayed length, so the concatenation has approximately twice as many bytes. Intervals resample the 12 replicates at each length. PPMd \rev{and binary-decomposition CTW transfer} substantially more shared context over the central range of the corpus; \rev{byte-level} CTW approaches \rev{them} only near the 95th percentile. \rev{The correlated-failure analysis of Appendix~\ref{app:compressor-failure} uses binary-decomposition CTW at depth 6 with $\alpha=1/16$. Its shared-context effect is 0.120, and the effect is 0.119 to 0.126 across the nine settings of depth (3, 6, 12) and $\alpha$ ($1/2$, $1/8$, $1/16$), so the three series drawn at depth 12 represent that setting.}}
  \label{fig:rev-compressor-factorial}
\end{figure}

We next separated transfer of sequential statistics from reuse of literal substrings in a paired $2\times2$ factorial experiment \rev{(Figure~\ref{fig:rev-compressor-factorial})}. Each sequence pair either shared or did not share a first-order byte-transition law (\emph{shared context}), and independently contained the same or disjoint randomly generated byte blocks (\emph{shared exact phrases}). The two sequences were generated independently, and shared blocks were inserted at different positions and in different orders. Phrase lengths were \rev{8} bytes at the shortest sequence length, \rev{29} bytes at \rev{238} bytes, and 32 bytes thereafter; phrase coverage was \rev{23\% to 25\%}. The six sequence lengths---\rev{71, 238, 653, 1,289, 2,540, and 3,946} bytes---are the 10th, 25th, 50th, 75th, 90th, and 95th percentiles of \rev{response length among the stored open-ended responses}. We used 12 paired replicates at each length. A factorial main effect is the change in $T_C$ when one factor is shared, averaged over the two levels of the other factor.

PPMd's aggregate shared-context effect is \rev{0.130} (95\% interval \rev{$[0.121,0.138]$}), compared with \rev{0.119 to 0.126 for binary-decomposition CTW across its nine settings of depth and $\alpha$ (0.120 at depth 6 with $\alpha=1/16$), 0.087 for bzip2, 0.044 for byte-level CTW, 0.037 for LZMA, and 0.036 for gzip. The paired PPMd contrasts with byte-level CTW, bzip2, LZMA and gzip} remain significant after Holm correction \rev{over the 15 contrasts ($p=7.5\times10^{-4}$}; Appendix~\ref{app:rank-and-multiplicity}). PPMd also has the largest exact-phrase effect (\rev{0.244}), so the result is not that PPMd ignores literal reuse. Rather, it combines phrase reuse with substantially stronger transfer of non-verbatim sequential statistics than the dictionary compressors. \rev{Byte-level} CTW exhibits the strongest context preference relative to its own phrase effect, but both effects are small at typical response lengths. At \rev{1,289} bytes, \rev{its} context effect is 0.016 against PPMd's \rev{0.161}; at \rev{2,540} bytes it is \rev{0.075} against \rev{0.150}; only at \rev{3,946} bytes does \rev{it} reach 0.136 against PPMd's \rev{0.131}. CTW can therefore detect the controlled transition structure, but requires substantially longer strings to do so under the byte-level configuration used here. \rev{Binary-decomposition CTW transfers almost as much sequential context as PPMd at the response lengths of the corpus.}

The compressor choice is also coupled to the aggregation protocol. We compress position-paired responses separately and average their distances, rather than concatenate all 50 responses in a model--prompt cell into one stream. The latter construction drove raw PPMd distances towards the upper boundary as cell length increased in our development diagnostics; response-level pairing retained substantially more variation and confines directional transfer to one response pair. Taken together, the preserved model-pair ordering, limited saturation, and finite-length context transfer support PPMd for the present response-level analysis. The comparison is nevertheless descriptive and heavily subsampled. \rev{Appendix~\ref{app:compressor-failure} repeats the correlated-failure analysis under each compressor.}

\subsection{Correlated-failure analysis under four compressors}
\label{app:compressor-failure}

\begin{figure}[htbp]
  \centering
  \includegraphics[width=0.9\linewidth]{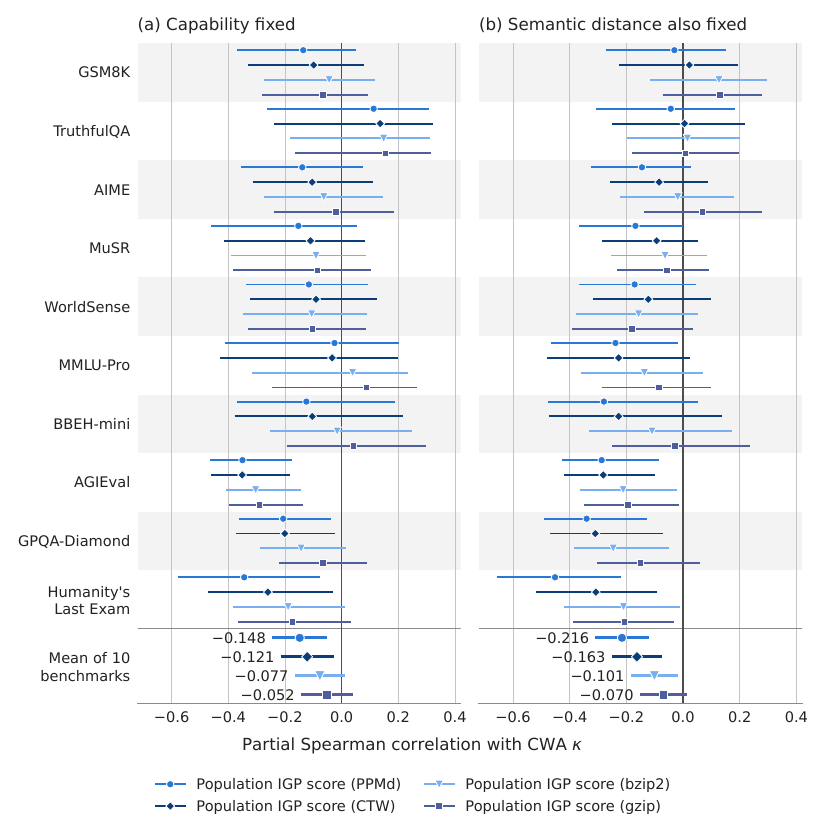}
  \caption{\revcolor{}\textbf{Population IGP score against CWA $\kappa$ under four compressors, MiniLM.} Each point is the partial Spearman correlation on one benchmark, across the model pairs of the 38-model fleet. (a) Capability level and gap controlled. (b) Semantic distance also controlled. Thin bars are 95\% model-node bootstrap intervals. The last row is the mean over the ten benchmarks with its 95\% $t$-interval, and the number beside it is that mean. The layout follows Figure~\ref{fig:dissociation}.}
  \label{fig:rev-compressor-forest}
\end{figure}

\begin{revblock}
We repeat the analysis of Section~\ref{sec:results} with CTW, bzip2 and gzip in place of PPMd, with every other step unchanged. CTW is the binary-decomposition form at depth 6 with $\alpha=1/16$ and a zero-redundancy estimator \citep{tjalkens_context-tree_1997,franken_overview_2002}. Table~\ref{tab:rev-compressors-bench} and Figure~\ref{fig:rev-compressor-forest} give the benchmark-level estimates for the population IGP score under MiniLM, and Table~\ref{tab:rev-compressors-measures} the cross-benchmark estimates for all four IGP measures under all three embeddings.

For the population IGP score, the association is negative under PPMd and CTW for all three embeddings, and under bzip2 and gzip for MiniLM only. The order of the four compressors is the order of their shared-context effect in Appendix~\ref{app:compressor-choice}: a compressor that transfers less sequential context gives a weaker association with correlated failure. The order is the same with no semantic distance in the model. There the population IGP score gives $-0.148$ $[-0.245,-0.050]$ under PPMd, $-0.121$ $[-0.215,-0.028]$ under CTW, $-0.077$ $[-0.165,+0.012]$ under bzip2 and $-0.052$ $[-0.144,+0.039]$ under gzip, with 9, 9, 8 and 7 of the 10 benchmarks negative. The permutation control of PPMd alone gives $+0.150$ $[+0.050,+0.249]$, the opposite sign (Figure~\ref{fig:components}). The compressors nonetheless rank model pairs alike, with Spearman correlations against the PPMd scores of 0.96 for CTW, 0.92 for bzip2 and 0.91 for gzip.

With the other three IGP measures of Appendix~\ref{app:igp-measures}, every compressor gives a negative estimate under every embedding, every interval excludes zero, and 8 to 10 of the 10 benchmarks are negative in every cell.
\end{revblock}

\begin{table}[htbp]
  \revcolor
  \caption{\textbf{Benchmark-level estimates under four compressors, MiniLM.} Partial Spearman correlation of the population IGP score with CWA $\kappa$, controlling capability level, capability gap and semantic distance.}
  \label{tab:rev-compressors-bench}
  \centering
  \begin{tabular}{lrrrr}
    \toprule
    Benchmark & PPMd & CTW & bzip2 & gzip \\
    \midrule
    TruthfulQA & $-0.043$ & $+0.005$ & $+0.015$ & $+0.008$ \\
    MMLU-Pro & $-0.238$ & $-0.228$ & $-0.136$ & $-0.085$ \\
    WorldSense & $-0.171$ & $-0.123$ & $-0.157$ & $-0.180$ \\
    BBEH-mini & $-0.279$ & $-0.227$ & $-0.109$ & $-0.028$ \\
    GSM8K & $-0.031$ & $+0.022$ & $+0.127$ & $+0.129$ \\
    AIME & $-0.145$ & $-0.084$ & $-0.018$ & $+0.068$ \\
    MuSR & $-0.168$ & $-0.093$ & $-0.064$ & $-0.058$ \\
    GPQA-Diamond & $-0.340$ & $-0.310$ & $-0.246$ & $-0.150$ \\
    Humanity's Last Exam & $-0.452$ & $-0.308$ & $-0.210$ & $-0.207$ \\
    AGIEval & $-0.287$ & $-0.282$ & $-0.211$ & $-0.194$ \\
    \midrule
    Mean & $-0.216$ & $-0.163$ & $-0.101$ & $-0.070$ \\
    \bottomrule
  \end{tabular}
\end{table}

\begin{table}[htbp]
  \revcolor
  \caption{\textbf{The four IGP measures under four compressors.} Mean partial Spearman correlation with CWA $\kappa$ over ten benchmarks, controlling capability and semantic distance, with its 95\% interval and the number of benchmarks with a negative estimate. The pair-specific rows use member-adjusted CWA with the accuracy-product control. The PPMd rows repeat Table~\ref{tab:main-measures}.}
  \label{tab:rev-compressors-measures}
  \centering
  \resizebox{\linewidth}{!}{%
  \begin{tabular}{llll}
    \toprule
    Compressor and measure & MiniLM & TE3-small & Qwen3 \\
    \midrule
    PPMd, population score & $-0.216$ $[-0.309,-0.122]$ 10/10 & $-0.128$ $[-0.207,-0.048]$ 9/10 & $-0.106$ $[-0.199,-0.013]$ 8/10 \\
    PPMd, population distance & $-0.227$ $[-0.297,-0.158]$ 10/10 & $-0.195$ $[-0.252,-0.139]$ 10/10 & $-0.188$ $[-0.252,-0.123]$ 10/10 \\
    PPMd, pair-specific score & $-0.257$ $[-0.367,-0.147]$ 9/10 & $-0.279$ $[-0.400,-0.157]$ 9/10 & $-0.248$ $[-0.353,-0.144]$ 10/10 \\
    PPMd, pair-specific distance & $-0.317$ $[-0.465,-0.169]$ 9/10 & $-0.338$ $[-0.493,-0.182]$ 9/10 & $-0.305$ $[-0.444,-0.165]$ 9/10 \\
    \midrule
    CTW, population score & $-0.163$ $[-0.252,-0.073]$ 8/10 & $-0.078$ $[-0.157,+0.002]$ 8/10 & $-0.064$ $[-0.151,+0.024]$ 8/10 \\
    CTW, population distance & $-0.168$ $[-0.226,-0.111]$ 10/10 & $-0.128$ $[-0.171,-0.084]$ 10/10 & $-0.119$ $[-0.170,-0.068]$ 10/10 \\
    CTW, pair-specific score & $-0.225$ $[-0.320,-0.130]$ 9/10 & $-0.249$ $[-0.358,-0.140]$ 9/10 & $-0.223$ $[-0.319,-0.128]$ 9/10 \\
    CTW, pair-specific distance & $-0.219$ $[-0.331,-0.107]$ 9/10 & $-0.246$ $[-0.371,-0.121]$ 8/10 & $-0.209$ $[-0.314,-0.103]$ 8/10 \\
    \midrule
    bzip2, population score & $-0.101$ $[-0.185,-0.017]$ 8/10 & $+0.006$ $[-0.081,+0.093]$ 4/10 & $+0.009$ $[-0.077,+0.095]$ 3/10 \\
    bzip2, population distance & $-0.137$ $[-0.189,-0.085]$ 10/10 & $-0.086$ $[-0.128,-0.044]$ 10/10 & $-0.076$ $[-0.128,-0.023]$ 8/10 \\
    bzip2, pair-specific score & $-0.219$ $[-0.312,-0.127]$ 9/10 & $-0.244$ $[-0.350,-0.137]$ 8/10 & $-0.213$ $[-0.303,-0.123]$ 9/10 \\
    bzip2, pair-specific distance & $-0.199$ $[-0.309,-0.089]$ 9/10 & $-0.227$ $[-0.349,-0.106]$ 9/10 & $-0.188$ $[-0.290,-0.085]$ 8/10 \\
    \midrule
    gzip, population score & $-0.070$ $[-0.152,+0.013]$ 7/10 & $+0.059$ $[-0.030,+0.148]$ 2/10 & $+0.067$ $[-0.017,+0.150]$ 3/10 \\
    gzip, population distance & $-0.145$ $[-0.203,-0.087]$ 9/10 & $-0.094$ $[-0.140,-0.048]$ 8/10 & $-0.082$ $[-0.136,-0.028]$ 8/10 \\
    gzip, pair-specific score & $-0.221$ $[-0.326,-0.116]$ 8/10 & $-0.246$ $[-0.365,-0.128]$ 8/10 & $-0.212$ $[-0.311,-0.112]$ 9/10 \\
    gzip, pair-specific distance & $-0.233$ $[-0.364,-0.102]$ 9/10 & $-0.258$ $[-0.399,-0.117]$ 8/10 & $-0.218$ $[-0.339,-0.098]$ 8/10 \\
    \bottomrule
  \end{tabular}}
\end{table}

\subsection{The \texorpdfstring{$V_p$}{Vp} family on sets and sequences}

The finite-set construction makes the relation among $V_1$, $V_2$, and $V_\infty$ exact. For finite sets $A$ and $B$, let
\begin{equation}
\Delta_p(A,B)=
\left(|B\setminus A|^p+|A\setminus B|^p\right)^{1/p},
\qquad
V_p(A,B)=\frac{\Delta_p(A,B)}{|A\cap B|+\Delta_p(A,B)},
\label{eq:vp-set}
\end{equation}
for $1\leq p<\infty$, with $\Delta_\infty(A,B)=\max\{|B\setminus A|,|A\setminus B|\}$ and $V_p(\varnothing,\varnothing)=0$. The endpoints are
\begin{equation}
V_1(A,B)=1-\frac{|A\cap B|}{|A\cup B|},
\qquad
V_\infty(A,B)=
\frac{\max\{|B\setminus A|,|A\setminus B|\}}
{\max\{|A|,|B|\}}.
\label{eq:vp-set-endpoints}
\end{equation}
Thus $V_1$ is the Jaccard distance, $V_\infty$ is the set analogue of Normalised Information Distance (NID), and $V_2$ uses the Euclidean norm of the two directional differences. In the symmetric Tversky family, these are the endpoints $V_1=D_{1/2,2}$ and $V_\infty=D_{0,1}$ \citep{tversky_features_1977,kjos_hanssen_interpolating_2022}.

Kjos-Hanssen proves that $V_p$ is a metric for every $p\in[1,\infty]$ \citep{kjos_hanssen_interpolating_2022}. The main step can be seen directly. Set containment gives
\[
|B\setminus A|\leq |B\setminus C|+|C\setminus A|,
\qquad
|A\setminus B|\leq |A\setminus C|+|C\setminus B|.
\]
Applying Minkowski's inequality to these two coordinates yields
\[
\Delta_p(A,B)\leq \Delta_p(A,C)+\Delta_p(C,B).
\]
The normalisation in Equation~\ref{eq:vp-set} also preserves the triangle inequality because $|B\setminus A|\leq\Delta_p(A,B)$ supplies the condition required by the ratio-normalisation lemma of \citet{kjos_hanssen_interpolating_2022}. Non-negativity, symmetry, and identity follow from the two set differences. This proves the metric result without selecting a special value of $p$.

The ordering follows from the standard ordering of norms on $\mathbb{R}^2$.
\begin{equation}
\Delta_\infty\leq\Delta_2\leq\Delta_1
\quad\Longrightarrow\quad
V_\infty\leq V_2\leq V_1.
\label{eq:vp-ordering}
\end{equation}
The implication holds because $t/(|A\cap B|+t)$ is increasing in $t$. If one directional difference is zero, all three values coincide. If the differences are balanced at $(t,t)$, their unnormalised values are $t$, $\sqrt{2}t$, and $2t$ for $p=\infty,2,1$, respectively. The members therefore differ most when each object contains substantial information absent from the other.

The set result also suggests a direct route to sequence distances. One can map a sequence $s$ to its Lempel--Ziv phrase dictionary $F(s)=\operatorname{LZSet}(s)$ and pull the set metric back as $V_p(F(s),F(t))$; at $p=1$, this recovers the Lempel--Ziv Jaccard distance \citep{raff_alternative_2017,kjos_hanssen_interpolating_2022}. The construction is exact but inherits the representation chosen by $F$. The map is parser-specific and non-injective, so distinct sequences with the same phrase set receive distance zero. This limitation motivates an object-level interpolation that does not first reduce each sequence to an explicit feature set.

\subsection{Extension to strings and finite objects}

For binary strings $x$ and $y$, let $K(x)$ be prefix Kolmogorov complexity, $K(x,y)$ the complexity of a fixed effective pairing, and
\[
I_K(x;y)=K(x)+K(y)-K(x,y).
\]
Using the directional conditional complexities $k_{x\mid y}=K(x\mid y)$ and $k_{y\mid x}=K(y\mid x)$, define
\begin{equation}
\Delta_p^K(x,y)=\left(k_{x\mid y}^p+k_{y\mid x}^p\right)^{1/p},
\qquad
V_p^K(x,y)=\frac{\Delta_p^K(x,y)}{I_K(x;y)+\Delta_p^K(x,y)}.
\label{eq:vp-k}
\end{equation}
Let $N=K(x,y,z)+2$ and $\lambda_N=O(\!\log N)$. Symmetry of information gives
\[
K(x,y)=K(x)+K(y\mid x)+O(\lambda_N)
=K(y)+K(x\mid y)+O(\lambda_N)
\]
\citep{levin_laws_1974,gacs_symmetry_1974}. Substitution into Equation~\ref{eq:vp-k} gives the two endpoint identities
\begin{align}
V_1^K(x,y)
  &=1-\frac{I_K(x;y)}{K(x,y)}
    +O\!\left(\frac{\lambda_N}{K(x,y)}\right),
    \label{eq:algorithmic-jaccard}\\
V_\infty^K(x,y)
  &=\frac{K(x,y)-\min\{K(x),K(y)\}}
  {\max\{K(x),K(y)\}}
    +O\!\left(\frac{\lambda_N}{\max\{K(x),K(y)\}}\right).
    \label{eq:vp-nid}
\end{align}
We refer to Equation~\ref{eq:algorithmic-jaccard} as the \emph{algorithmic Jaccard distance}; Equation~\ref{eq:vp-nid} is NID \citep{li_similarity_2004}. At the conditional-complexity level, $V_1^K=D_{1/2,2}^K$ and $V_\infty^K=D_{0,1}^K$ are exact. Only their reduction to the displayed joint-complexity forms incurs logarithmic slack.

\paragraph{Universality and the information retained at finite $p$.}
Let
\[
M_K=\max\{K(x\mid y),K(y\mid x)\},\qquad
m_K=\min\{K(x\mid y),K(y\mid x)\}.
\]
Then $\Delta_\infty^K=M_K$ and, for finite $p$,
$\Delta_p^K=(M_K^p+m_K^p)^{1/p}$. Norm equivalence in two dimensions gives
\begin{equation}
\Delta_\infty^K\leq \Delta_p^K
\leq 2^{1/p}\Delta_\infty^K.
\label{eq:vp-norm-equivalence}
\end{equation}
After absorbing the $O(\lambda_N)$ possible negativity of $I_K$ into the symmetry-of-information slack, the same comparison holds after normalisation.
\begin{equation}
V_\infty^K(x,y)\leq V_p^K(x,y)
\leq 2^{1/p}V_\infty^K(x,y)
+O\!\left(\frac{\lambda_N}{H}\right),
\label{eq:vp-nid-factor}
\end{equation}
where $H$ is the denominator scale and $2^{1/\infty}=1$. NID minorises every admissible upper-semicomputable normalised distance $d$ satisfying the density condition \citep{li_similarity_2004,vitanyi_normalized_2009}. Equation~\ref{eq:vp-nid-factor} therefore implies
\begin{equation}
V_p^K(x,y)\leq 2^{1/p}d(x,y)
+O\!\left(\frac{\lambda_N}{H}\right).
\label{eq:vp-factor-universality}
\end{equation}
Thus $V_2^K$ and $V_1^K$ inherit NID's minorisation property within factors $\sqrt{2}$ and $2$, respectively, while only $V_\infty^K$ retains the sharp coefficient one. This is a difference in worst-case universality, not an ordering of usefulness for a particular outcome.

Finite $p$ remains sensitive to a quantity absent from NID. Up to logarithmic slack,
\[
V_\infty^K=\frac{M_K}{I_K+M_K},\qquad
V_1^K=\frac{M_K+m_K}{I_K+M_K+m_K},
\]
and hence
\begin{equation}
V_1^K-V_\infty^K
=\frac{I_Km_K}
{(I_K+M_K)(I_K+M_K+m_K)}
+O\!\left(\frac{\lambda_N}{H}\right).
\label{eq:vp-retained-direction}
\end{equation}
At fixed shared information $I_K>0$ and larger directional cost $M_K$, every finite-$p$ member varies with the smaller cost $m_K$, whereas NID does not. The members agree when one directional cost vanishes and separate as the two costs become more balanced. Retaining this second direction may be useful when the target depends on reciprocal novelty---neither object can be described from the other by a short program---rather than on the larger one-sided difference alone. This does not imply that a finite-$p$ member must predict correlated failure better. That question is outcome-specific and empirical. Moreover, the universality statements above concern the ideal $K$-level quantities; a real-compressor approximation, and especially its signed residual after projection, does not inherit them automatically.

The metric proof also transfers, with the same qualification. Conditional descriptions compose according to
\[
K(x\mid y)\leq K(x\mid z)+K(z\mid y)+O(\lambda_N),
\]
and the analogous inequality holds in the opposite direction. Minkowski's inequality therefore gives
\begin{equation}
\Delta_p^K(x,y)\leq
\Delta_p^K(x,z)+\Delta_p^K(z,y)+O(\lambda_N).
\label{eq:vp-k-triangle}
\end{equation}
For the ratio normalisation, symmetry of information and the same composition inequality imply
\[
I_K(x;z)-I_K(x;y)
\leq K(z\mid y)+O(\lambda_N)
\leq\Delta_p^K(y,z)+O(\lambda_N).
\]
Applying the set-level ratio argument then yields the triangle inequality for $V_p^K$ up to relative $O(\!\lambda_N/H)$ slack, where $H$ is the scale of the denominators. Identity is approximate rather than literal. Even $K(x\mid x)$ is $O(1)$, and distinct strings related by a fixed reversible procedure can have $K(x\mid y),K(y\mid x)=O(1)$. More precisely, for a family of pairs $(x_n,y_n)$, their mutual description lengths are \emph{uniformly bounded} if there is one constant $c$, independent of $n$, such that
\[
\max\{K(x_n\mid y_n),K(y_n\mid x_n)\}\leq c
\quad\text{for every }n.
\]
If the denominator scale $H_n$ grows, then $V_p^K(x_n,y_n)=O(1/H_n)$ and tends to zero. Thus $V_p^K$ separates growing algorithmic objects only up to uniformly bounded mutual description length; it is a logarithmic pseudometric rather than an exact metric on literal finite strings.

The same result applies to any finite object supplied with an effective self-delimiting encoding. A computable reversible change of encoding alters Kolmogorov complexity by at most an additive constant, which is absorbed by the logarithmic term. The object-level statement is therefore not tied to one byte representation, although any computable approximation using a real compressor remains representation-dependent.

\subsection{Compressed-length interpolation}

Let $c_x=C(x)$, $c_y=C(y)$, and $c_{xy}=C(xy)$, and define the estimated directional complexities
\[
u_C=c_{xy}-c_y,\qquad v_C=c_{xy}-c_x,
\qquad I_C=c_x+c_y-c_{xy}.
\]
Replacing $K$ by a normal compressor $C$ in Equation~\ref{eq:vp-k} gives $\Delta_p^C=\|(u_C,v_C)\|_p$ and $V_p^C=\Delta_p^C/(I_C+\Delta_p^C)$. The normal-compressor axioms---idempotence, monotonicity, symmetry, and distributivity up to $\varepsilon(n)=O(\!\log n)$---supply the compressor analogue of conditional composition,
\[
C(x\mid z)\leq C(x\mid y)+C(y\mid z)+\varepsilon(n),
\]
so the proof above carries through with $O(\!\varepsilon(n)/H)$ relative slack \citep{cilibrasi_clustering_2005}. When $u_C,v_C\geq0$, the endpoints simplify to
\begin{equation}
V_1^C=\frac{2c_{xy}-c_x-c_y}{c_{xy}},
\qquad
V_\infty^C=
\frac{c_{xy}-\min\{c_x,c_y\}}{\max\{c_x,c_y\}},
\label{eq:vp-compressor-endpoints}
\end{equation}
so $V_1^C$ is the computable algorithmic Jaccard distance and $V_\infty^C$ is NCD. The middle member uses $\Delta_2^C=(u_C^2+v_C^2)^{1/2}$.

A real compressor need not satisfy the normality axioms exactly. PPMd is left-to-right and $C(xy)$ need not equal $C(yx)$; the reported statistic uses the fixed orientation shown in Equation~\ref{eq:ncd}. We therefore treat all three members as empirical compression dissimilarities, not exact metrics. The implementation clips negative directional estimates to zero for $V_2$ and clips all reported values to $[0,1]$. Under compressor normality these adjustments are within the same $O(\!\varepsilon(n)/H)$ approximation.

\subsection{Residualisation as a within-prompt projection}
\label{app:residualisation-projection}

For a prompt $q$, collect the $n=\binom{M}{2}=703$ raw pair distances into the vector $y_q\in\mathbb{R}^n$ and the corresponding permutation-control distances into $p_q\in\mathbb{R}^n$. Let $\iota_n=(1,\ldots,1)^\top$ and
\begin{equation}
X_q=\begin{bmatrix}\iota_n&p_q\end{bmatrix},
\qquad
P_q=X_q(X_q^\top X_q)^{-1}X_q^\top.
\label{eq:resid-projector}
\end{equation}
The fitted component and residual are
\begin{equation}
\widehat{y}_q=P_qy_q,
\qquad
r_q=(I_n-P_q)y_q,
\qquad
\spop=\frac{1}{Q}\sum_{q=1}^{Q}r_q.
\label{eq:resid-projection}
\end{equation}
The matrix $P_q$ is symmetric and idempotent, $P_q^\top=P_q$ and $P_q^2=P_q$, and therefore projects onto the span of the intercept and the prompt-specific control. The normal equations give
\begin{equation}
X_q^\top r_q=0.
\label{eq:resid-orthogonality}
\end{equation}
The residuals consequently have mean zero and zero sample covariance with the permutation control within each prompt. Positive residuals are pairs whose observed distance is greater than \rev{their permutation control predicts} for that prompt; negative residuals are closer than predicted. Zero is a fitted reference point, not the absence of diversity.

\begin{figure}[ht]
  \centering
  \includegraphics[width=0.68\linewidth]{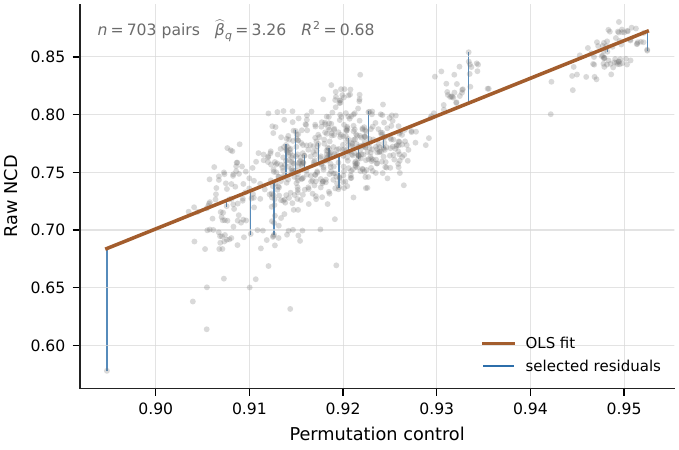}
  \caption{\textbf{Within-prompt residualisation.} One representative prompt with 703 model pairs. The line is the within-prompt OLS fit and the blue segments show a subset of residuals.}
  \label{fig:supp-residualisation-projection}
\end{figure}

The regression is fitted separately for each prompt because pair comparisons are matched on prompt content and the scale of the control varies across prompts. \rev{Slopes} range from 1.03 to 7.65, with mean 3.13. Averaging $r_q$ then gives every prompt equal weight. Orthogonality is a within-prompt property. It does not require the averaged residual vector to be exactly orthogonal to the averaged control vector, because cross-prompt products remain. Their Pearson correlation \rev{across model pairs} is $-0.048$.

\begin{figure}[t]
  \centering
  \includegraphics[width=\linewidth]{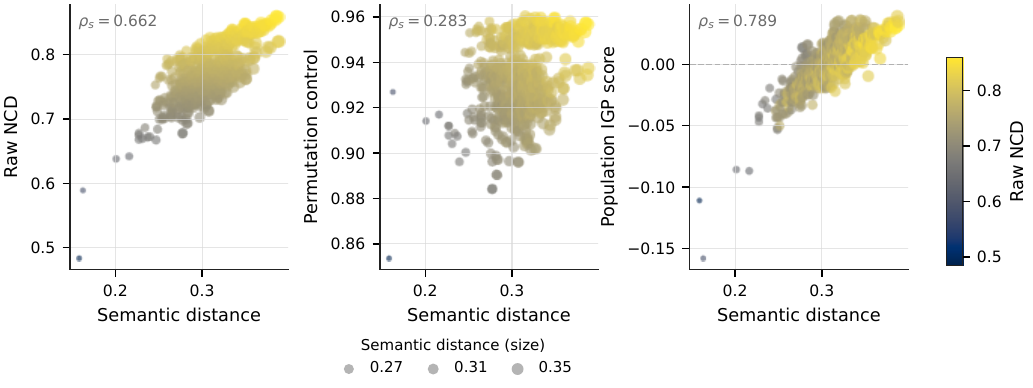}
  \caption{\textbf{Pairwise effect of residualisation.} All 703 model pairs are plotted against semantic distance. Point colour records raw NCD and point size also records semantic distance; both encodings are held fixed across panels. The panels show raw NCD, its permutation control, and the \rev{population IGP score}. Rank associations with semantic distance are printed within each panel.}
  \label{fig:supp-residualisation-colour}
\end{figure}

Figure~\ref{fig:supp-residualisation-colour} shows what the projection changes at the pair level. The point colour is fixed to raw NCD and the point size to semantic distance in all three panels. Raw NCD is strongly ordered by its own colour, as it must be. The permutation panel retains much of that ordering. After projection, the colours are rearranged vertically. At a fixed semantic distance, pairs with larger raw NCD need not have \rev{a larger population IGP score (Section~\ref{sec:permutation-residual})}. The partial Spearman association between raw NCD and the \rev{population IGP score}, controlling for semantic distance, is $-0.409$. This conditional reordering is distinct from the positive marginal association between the \rev{score} and semantic distance ($\rho_s=0.789$, against $0.662$ for raw NCD, printed in the panels). \rev{The population IGP score is therefore strongly collinear with semantic distance. Since $1-\rho_s^2\approx0.38$, only about 38\% of the rank variance of semantic distance is not shared with the score, and the cross-controlled semantic estimate of Section~\ref{sec:results} describes that part.}

\begin{figure}[t]
  \centering
  \includegraphics[width=0.9\linewidth]{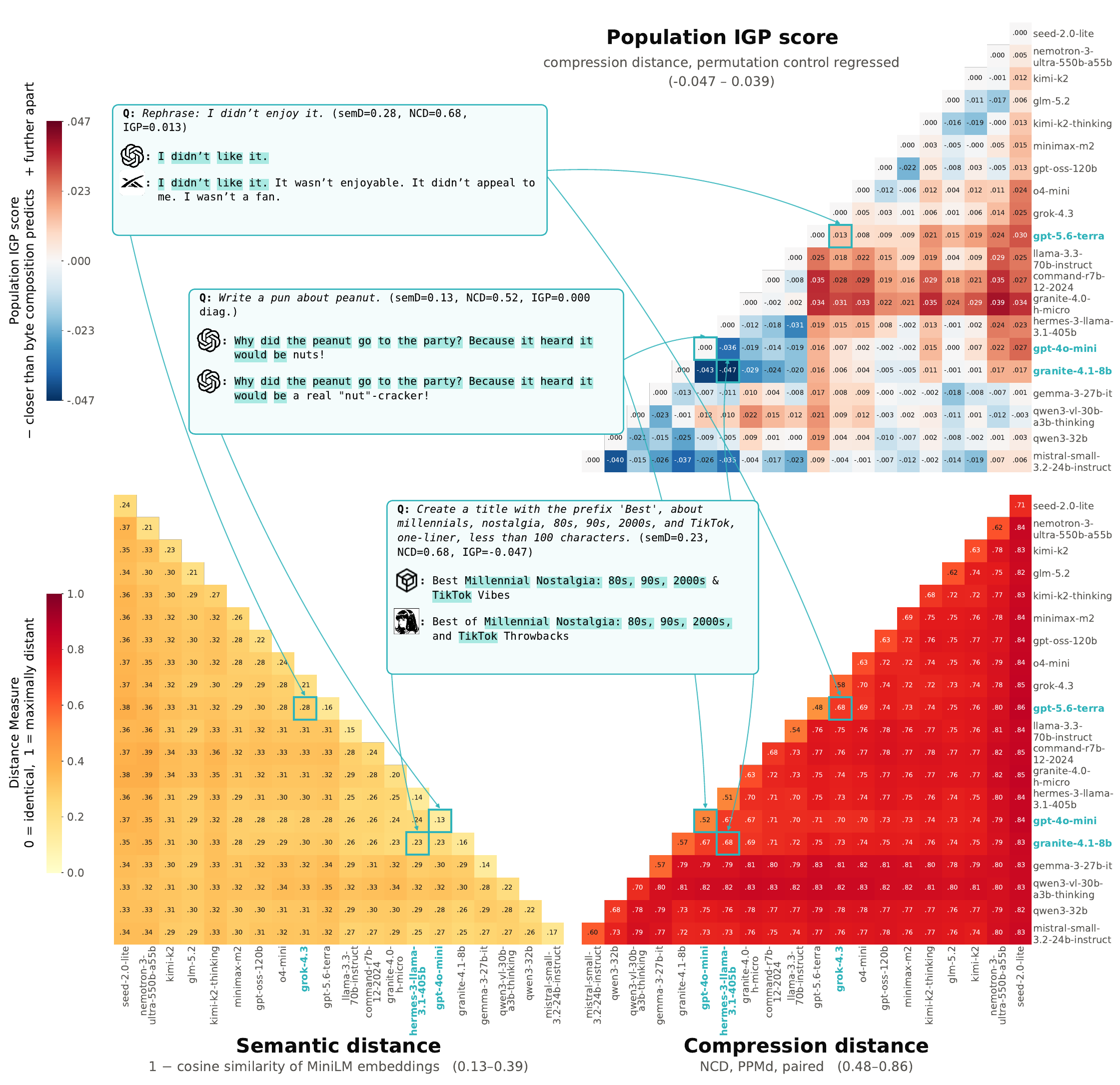}
  \caption{\addtext{\textbf{\rev{Semantic distance, raw NCD, and the population IGP score for the same responses.}} The lower matrices show semantic distance (left) and raw paired PPMd NCD (right). The upper-right matrix shows the \rev{population IGP score,} obtained by residualising NCD against the within-prompt byte-permutation control and averaging across prompts. 20 of 38 models are displayed; reported matrix associations use the 703 off-diagonal pairs. Annotated boxes contain response excerpts and give the semantic distance, raw NCD, and \rev{population IGP score} for a positive, near-zero, and negative \rev{score}.}}
  \label{fig:supp-views-stacked}
\end{figure}

\addtext{Figure~\ref{fig:supp-views-stacked} shows the raw NCD matrix beside the \rev{semantic distance and population IGP score} matrices of Figure~\ref{fig:views}, with the same annotated pairs. Across all 703 pairs raw NCD spans $[0.484,0.859]$, against $[0.159,0.388]$ for semantic distance, and the two agree on coarse population structure but not on many individual pairs (Spearman $\rho=0.662$, Pearson $r=0.745$). The Granite--Hermes and GPT-5.6--Grok pairs have raw NCD values of 0.675 and 0.678, indistinguishable within that range, yet \rev{population IGP scores} of $-0.047$ and $+0.013$.}

\begin{figure}[t]
  \centering
  \includegraphics[width=\linewidth]{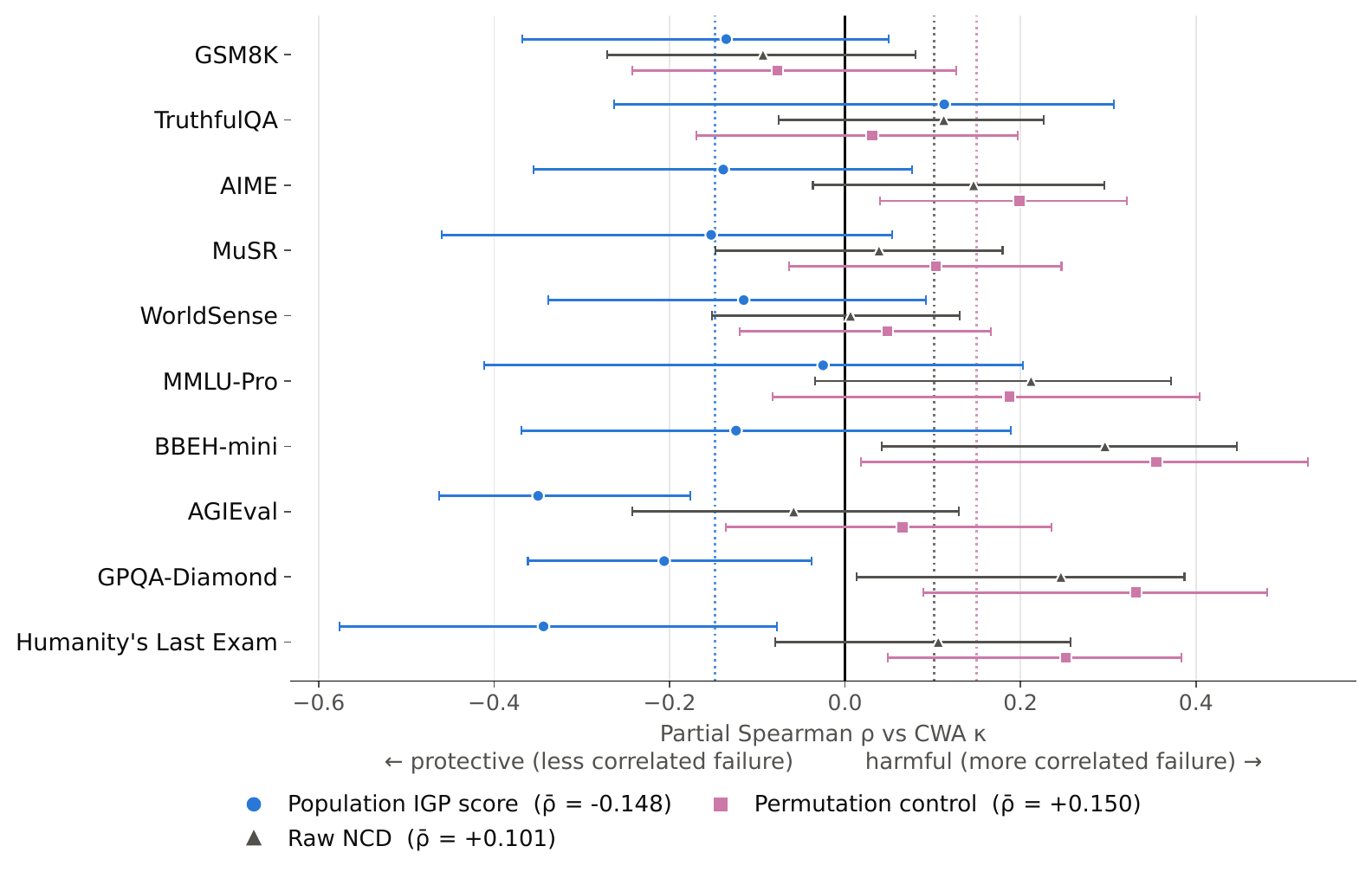}
  \caption{\textbf{Compression components against correlated failure.} Points are benchmark-specific partial Spearman correlations with capability level and gap held fixed; bars are model-level bootstrap intervals and dotted lines are cross-benchmark means. Negative values denote \rev{lower CWA $\kappa$}. Raw NCD ($\bar\rho=+0.101$) and the \rev{permutation control} ($+0.150$) are associated with more correlated failure on average, whereas the \rev{population IGP score} is associated with less ($-0.148$).}
  \label{fig:components}
\end{figure}

Figure~\ref{fig:components} decomposes compression distance while holding capability level and gap fixed. The components differ in sign. Raw NCD is mildly associated with \emph{more} correlated failure ($\bar\rho=+0.101$), and the \rev{permutation control} is more positive ($+0.150$). Only the \rev{population IGP score} is associated with \emph{less} correlated failure ($-0.148$). \rev{Raw NCD} therefore does not merely add noise. It suggests the opposite conclusion. We therefore define \rev{IGP diversity} using the permutation residual. Only after removing \rev{what the permutation control predicts} does the measure identify pairs with less correlated failure.

\subsection{Within-model response variability}
\label{app:within-model}

The Artificial Hivemind characterises open-ended homogeneity at two levels. These are intra-model repetition among responses repeatedly sampled from one model and inter-model similarity across models \citep{jiang_artificial_2025}. Our primary analysis extends the inter-model comparison. To complete the comparison on the same prompt taxonomy, we apply the corresponding within-model semantic construction to our 38-model response corpus and place it beside within-model compression variability. The construction is analogous rather than a numerical replication because we use MiniLM embeddings, but both analyses average pairwise embedding similarity among repeated responses to the same prompt.

This extension uses NCD at a second scale. Between models, NCD compares the sequential organisation expressed by different deployed configurations. Within a model and prompt, it compares repeated draws from the same conditional response distribution and therefore characterises realised stochastic response variability under the fixed decoding policy. It is not an estimator of an intrinsic entropy or randomness parameter. The measured variation includes the effects of the prompt, model, interface, and sampling procedure. Nor does it imply that a model changes its generative process across prompts. Each prompt conditions the same deployed configuration on a different input.

For model $m$ and prompt $q$, within-model semantic distance $d^{\mathrm{sem}}_{mq}$ is one minus the mean cosine similarity over all distinct pairs among the 50 response embeddings. Within-model compression distance $d^{\mathrm{NCD}}_{mq}$ pairs response $r$ with response $r+25$ and averages the resulting 25 disjoint NCD values. The permutation control $d^{\mathrm{perm}}_{mq}$ is computed from the same 25 response pairs. Because this is a separate estimand from the cross-model analysis, we fit new coefficients across the 38 within-model observations for each prompt as follows.
\begin{equation}
\begin{aligned}
(\widehat{\alpha}^{\mathrm{intra}}_q,
 \widehat{\beta}^{\mathrm{intra}}_q)
&=\operatorname*{arg\,min}_{a,b}\sum_{m=1}^{M}
\left(d^{\mathrm{NCD}}_{mq}-a-bd^{\mathrm{perm}}_{mq}\right)^2,\\
\widehat{\varepsilon}^{\mathrm{intra}}_{mq}
&=d^{\mathrm{NCD}}_{mq}
-\left(\widehat{\alpha}^{\mathrm{intra}}_q+
\widehat{\beta}^{\mathrm{intra}}_q d^{\mathrm{perm}}_{mq}\right).
\end{aligned}
\label{eq:intra-resid}
\end{equation}
The resulting residual is relative to the model population for that prompt. Positive values denote greater within-model order-specific variation than predicted from byte composition, and negative values denote less. As in the pairwise construction, the residual is orthogonal to the permutation control within each prompt; it is not residualised against semantic distance.

Models differ under the shared policy. Mean semantic distance across all model--prompt cells is $0.197$, with model means from $0.134$ to $0.273$; mean NCD is $0.614$, with model means from $0.483$ to $0.719$. Treating prompt as a repeated-measures block, model effects are present for semantic distance (Kendall's $W=0.357$), raw NCD ($W=0.538$), and the intra-model order-specific residual ($W=0.422$); all three Friedman tests have $p<10^{-250}$. Appendix~\ref{app:rank-and-multiplicity} defines the coefficient and test. These are differences in the degree of response variability, not evidence of broad semantic disagreement. Repeated responses remain close in the embedding space for every model.

\begin{figure}[p]
  \centering
  \includegraphics[width=\linewidth]{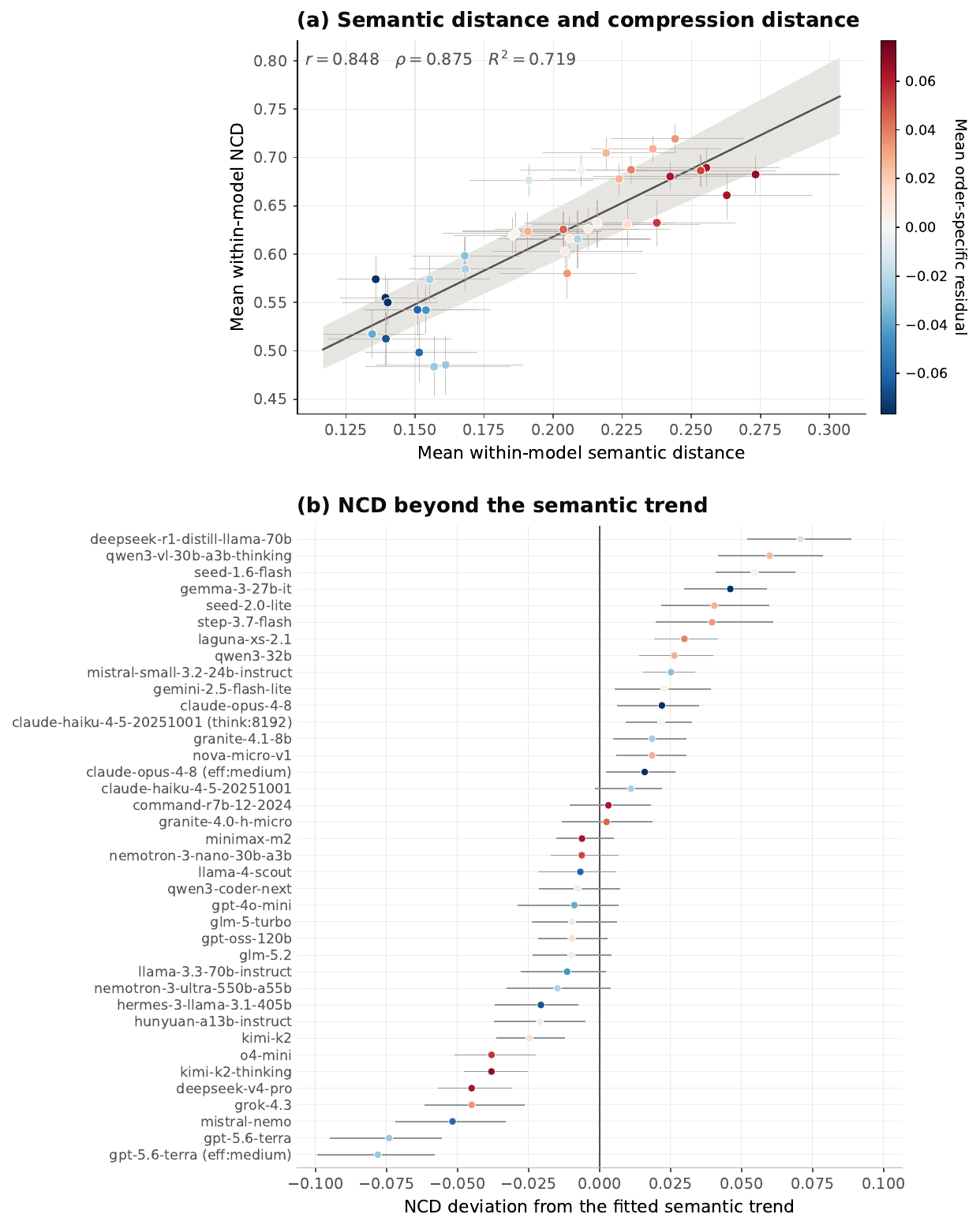}
  \caption{\textbf{Within-model semantic and compression variability.} (a) Each point is one model, positioned by its mean semantic distance and mean raw NCD over 100 prompts. Horizontal and vertical bars are 95\% prompt-bootstrap intervals. The line is the OLS fit across the 38 model means, the band refits that relationship in each prompt-bootstrap sample, and colour records the separately fitted mean intra-model order-specific residual from Equation~\ref{eq:intra-resid}. (b) Model-specific NCD deviation $\delta_m$ from the fitted semantic trend. Intervals resample prompts jointly across models and refit the trend in every replicate. Positive values indicate more compression variability than semantic distance predicts; negative values indicate less. The figure therefore displays both the shared response-variability component and the model-specific departures from it.}
  \label{fig:supp-intra-model}
\end{figure}

The measures share substantial variation. Across the 38 model means, semantic distance and raw NCD have Pearson $r=0.848$, Spearman $\rho_s=0.875$, and $R^2=0.719$ under the linear fit in Figure~\ref{fig:supp-intra-model}. The order-specific residual follows the same broad pattern. Its association with semantic distance across model means is $r=0.931$ and $\rho_s=0.917$. This does not indicate a failure of the projection. The permutation control is only weakly associated with semantic distance across model means ($r=0.267$), so removing the composition-predicted component leaves most of the semantic-associated sequential variation. At the prompt level, the mean Pearson association across models decreases from $0.732$ for raw NCD to $0.702$ for the residual; averaging over prompts exposes stable model-level differences that both measures track.

The association is strong but not exact. The linear semantic trend leaves $28.1\%$ of between-model NCD variance unexplained, and rankings by the two model means disagree for 115 of the 703 pairwise model orderings. To show where the measures differ, Figure~\ref{fig:supp-intra-model} defines
$\delta_m=\bar d^{\mathrm{NCD}}_m-(\widehat a+\widehat b\bar d^{\mathrm{sem}}_m)$,
where $(\widehat a,\widehat b)$ is fitted across the 38 model means. DeepSeek-R1-Distill-Llama-70B, Qwen3-VL-Thinking, and Seed-1.6-Flash have more within-model NCD than their semantic distance predicts. Both GPT-5.6-Terra configurations and Mistral-Nemo have less. These departures are descriptive differences between the measures; they do not by themselves establish which component predicts a separate outcome.

The intra-model analysis therefore completes the comparison with the Artificial Hivemind while giving NCD a complementary interpretation. The same output-derived statistic characterises separation between deployed model configurations and stochastic response variability within a configuration. At the within-model scale, semantic and compression variability are related but not interchangeable. Semantic distance records changes in expressed meaning, whereas NCD also responds to how repeated draws vary in their sequential organisation. Accordingly, this analysis characterises how the two measures covary under repeated sampling, but does not independently establish that compression captures information beyond semantic distance. Evidence that compression carries outcome-relevant information beyond semantic distance comes instead from the held-out correlated-failure analysis, where semantic distance and capability are controlled directly. We hypothesise that NCD may capture non-trivial variation in stochastic response variability beyond semantic distance; testing this possibility requires a dedicated analysis and is left to future work.

\subsection{Permutation count and aggregation}

\begin{figure}[ht]
  \centering
  \includegraphics[width=\linewidth]{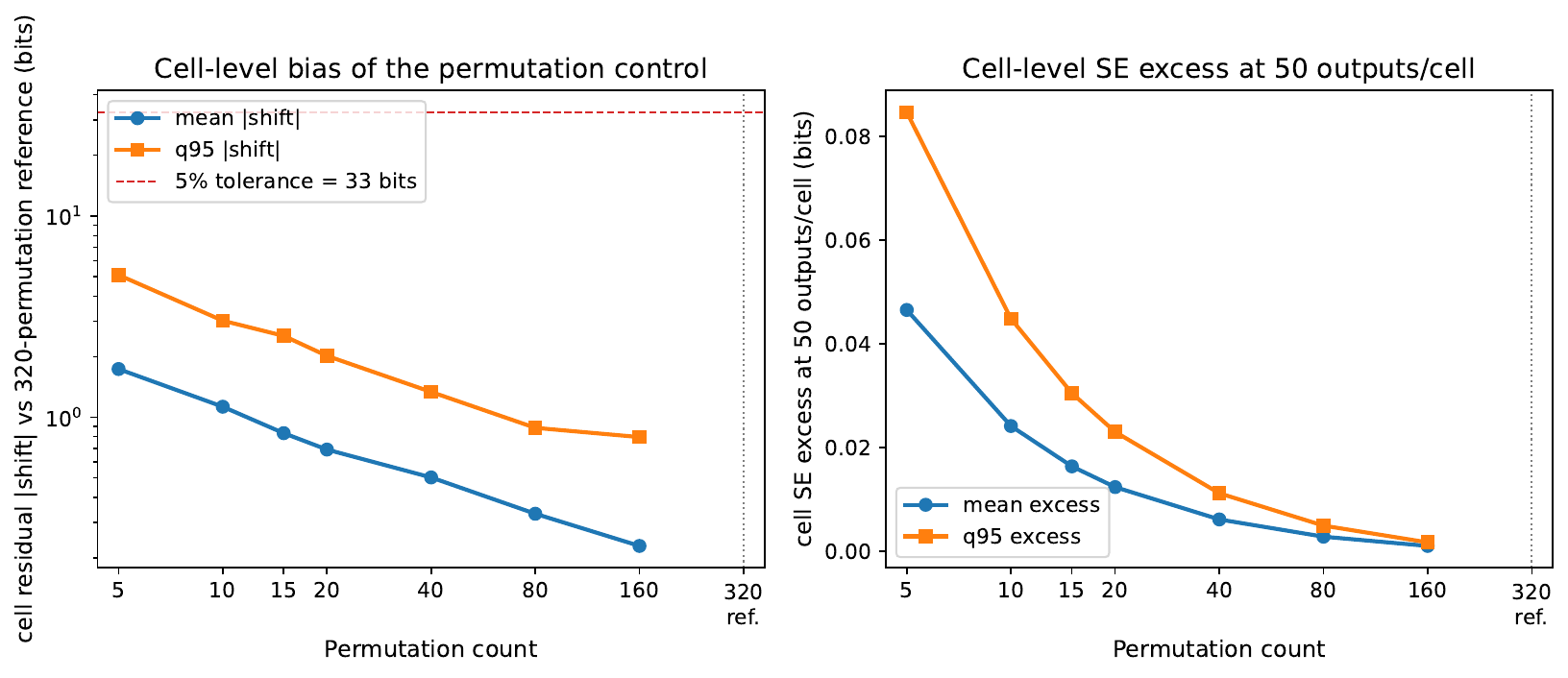}
  \caption{\textbf{Permutation-count sensitivity for 50-output cells.} Left: mean and 95th-percentile absolute shifts in cell residuals at each tested permutation count relative to \rev{$P=320$}; the dashed line is the 5\% tolerance of 33 bits. Right: mean and 95th-percentile excess standard error for a 50-output cell relative to \rev{$P=320$}. The \rev{$P=320$} reference is marked on both axes but omitted as a data point because both quantities are zero by construction. At the chosen value \rev{$P=20$}, the mean shift is 0.7 bits and the excess standard error is 0.01 bits.}
  \label{fig:supp-permutation}
\end{figure}

Each response byte stream is permuted $P=20$ times with deterministic seeds derived from the master seed, query identifier, and label. A permutation-count sensitivity analysis over \rev{$P\in\{5,\ldots,320\}$} on 400 outputs found 2.5 bits of per-output drift between \rev{$P=20$} and \rev{$P=320$}, 0.4\% of a typical 655.6-bit residual, with mean drift $+0.15$ bits. For a cell containing 50 outputs, the \rev{$P=20$} control shifted the residual by 0.7 bits and added 0.01 bits of standard error \rev{(Figure~\ref{fig:supp-permutation})}. All tested permutation counts \rev{passed both the per-output and the cell-level criteria}, so \rev{$P=20$} is conservative.

Within a model-pair/query cell, $K=\min(R_i,R_j,50)$ position-paired responses are averaged. Queries then receive equal weight, so a query contributing 25 response pairs counts as much as one contributing 50. Reconstruction from the per-query cache reproduces the stored matrices exactly.

\subsection{Robustness within the raw-byte family}

\begin{figure}[t]
  \centering
  \includegraphics[width=\linewidth]{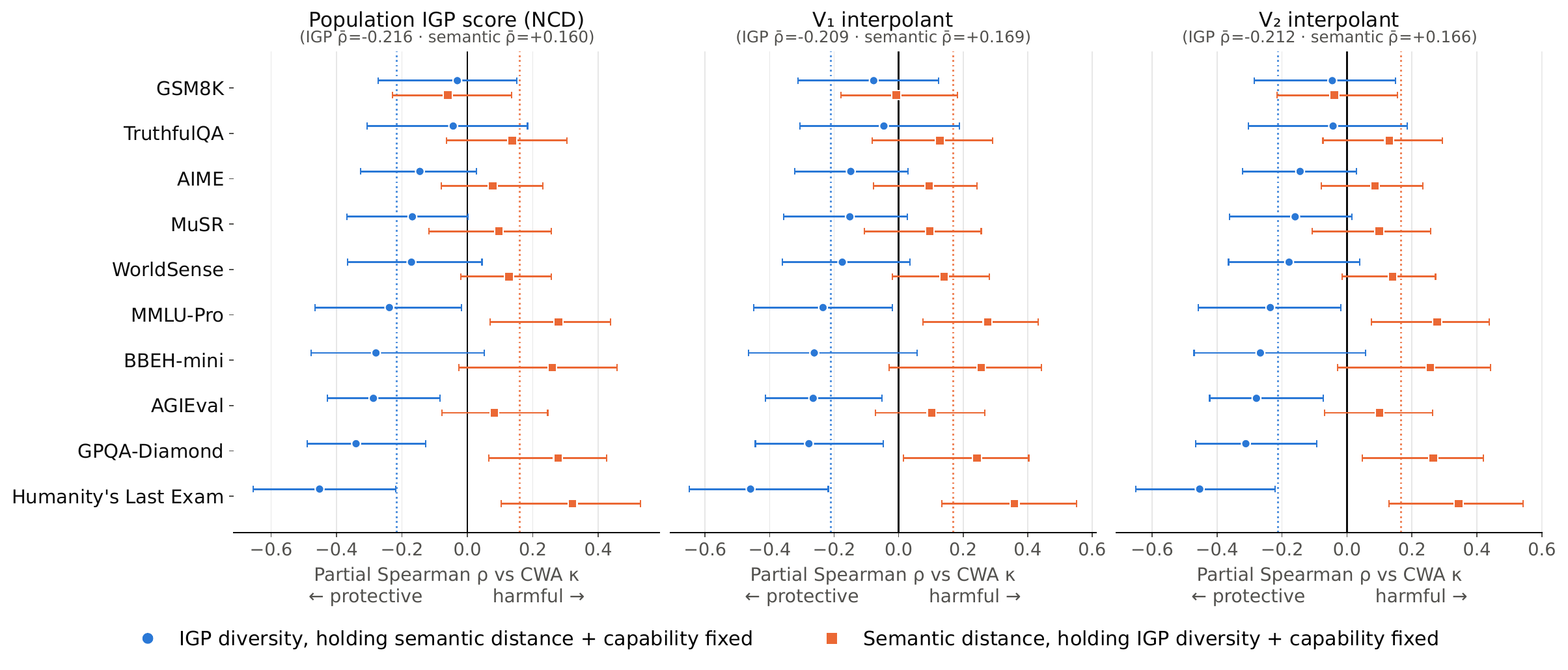}
  \caption{\textbf{\rev{The association with correlated failure} is stable across the $V_p$ family.} Panels show the \rev{population IGP score computed from NCD ($V_\infty$)}, $V_1$, and $V_2$. Blue circles are partial Spearman correlations between \rev{the population IGP score and CWA $\kappa$}, holding semantic distance and capability fixed; orange squares reverse the cross-control, estimating semantic distance while holding \rev{the population IGP score} and capability fixed. Horizontal bars are 95\% model-node-bootstrap intervals, solid vertical lines mark zero, and dotted lines mark the cross-benchmark means reported above each panel. \rev{Estimates for the population IGP score} are negative on all ten benchmarks for every \rev{member of the family} and vary little across panels relative to their uncertainty; semantic estimates are positive on nine of ten.}
  \label{fig:supp-interpolants}
\end{figure}

The finite-$p$ members test whether retaining the smaller directional compression increment changes the association with correlated failure. This sensitivity may be useful when process separation is reciprocal, but it does not make either member preferable to NCD in advance. Figure~\ref{fig:supp-interpolants} shows that the result is stable across the family. Under the capability-plus-cross-control specification, the cross-benchmark means are $-0.209$ for residualised $V_1$, $-0.212$ for $V_2$, and $-0.216$ for $V_\infty$. \rev{For every member of the family}, \rev{the population IGP score has a negative estimate} on all ten benchmarks, while the cross-controlled semantic estimate is positive on nine of ten. \rev{Its estimates} have Spearman agreement 0.988 across \rev{the three members} and a mean absolute difference of 0.019, compared with a mean node-bootstrap interval width of 0.149. The small shifts between panels therefore support robustness across the family.

\FloatBarrier

\section{Outcome Measures}
\label{app:outcomes}

\subsection{Epoch-native CWA}

\begin{figure}[t]
  \centering
  \includegraphics[width=\linewidth]{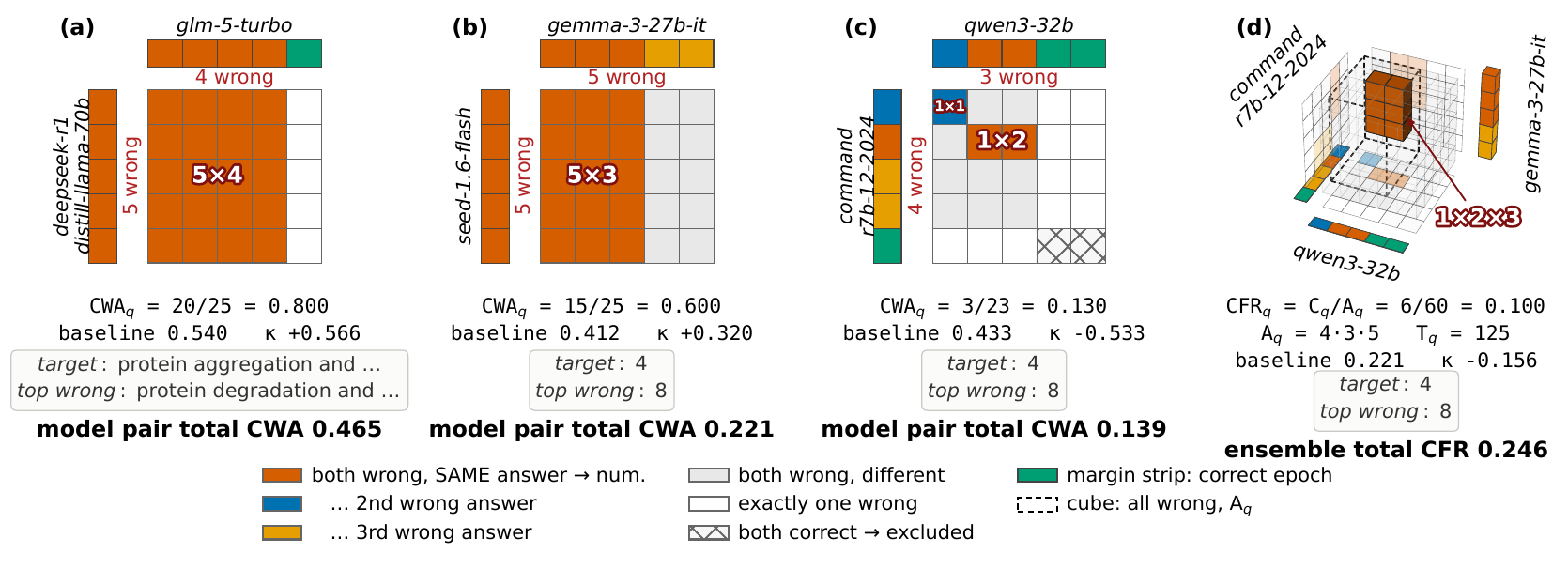}
  \caption{\textbf{Correlated wrong-answer agreement and convergent failure rate on GPQA-Diamond.} (a)--(c): each grid crosses five sampled epochs from the row and column models. Item-level raw rate, per-question leave-pair-out baseline, and chance-corrected $\kappa$ are shown below each grid. (d): Answers in (b)\&(c) extended to a three-model ensemble. Each voxel is one triple of epochs, one from each member ($5^3=125$ triples). The dashed box is the set of triples in which all three members are wrong ($A_q=4\cdot3\cdot5=60$), the solid box the subset that agree on one wrong answer ($C_q=1\cdot2\cdot3=6$). Three member ensemble is shown for legibility, analysis consider 5 member ensembles ($5^5=3{,}125$ epoch tuples per item.)}
  \label{fig:cwa}
\end{figure}

Equation~\ref{eq:cwa} pools question-level numerator and denominator counts into a single model-pair rate. Exactly-one-wrong response pairs contribute to the denominator, while the numerator records shared wrong answers. The epoch cross-product preserves repeated attraction to the same wrong answer through the response counts $c_i(a)c_j(a)$. \rev{In Figure~\ref{fig:cwa}, the} examples span fixed quantiles of pooled \rev{pairwise} CWA among 700 non-sibling pairs and illustrate why raw agreement must be read against item-specific distractor attraction. \rev{In panel (d), adding} a member can only remove triples from the convergent set.

\begin{figure}[t]
  \centering
  \includegraphics[width=\linewidth]{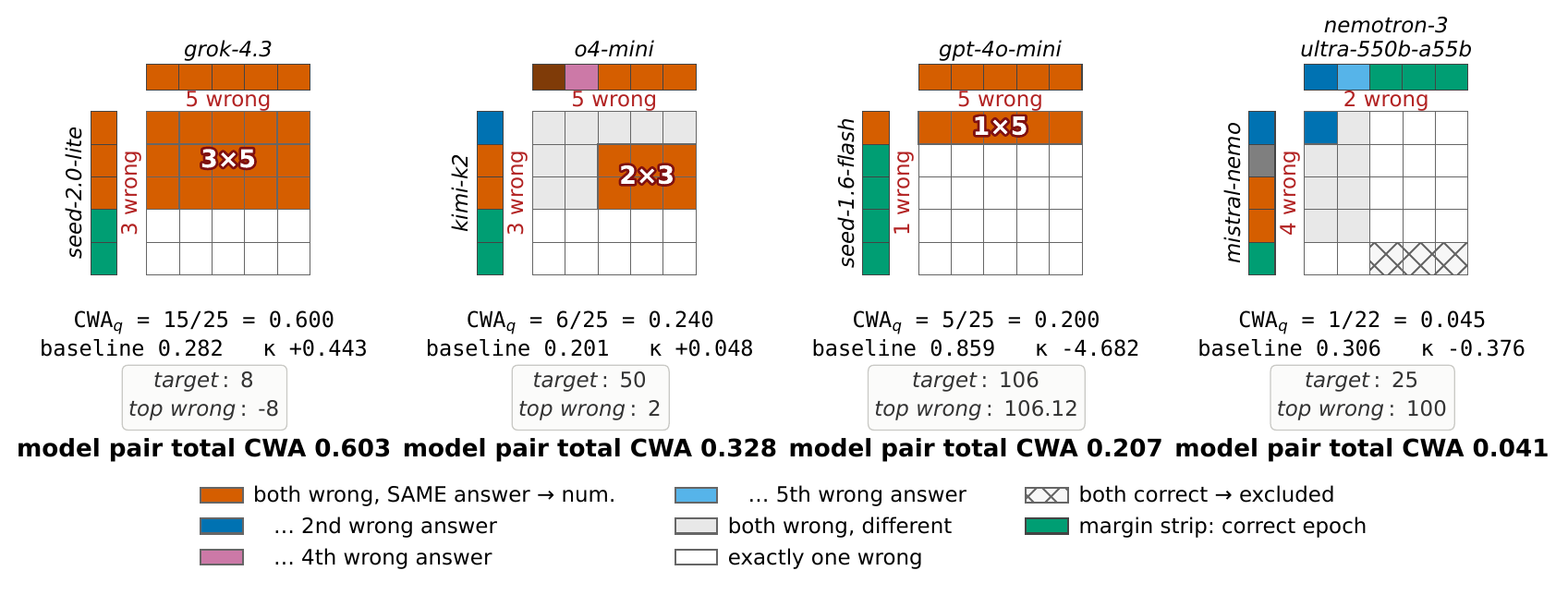}
  \caption{\textbf{Epoch-native CWA construction on GSM8K, complementing the GPQA-Diamond example in Figure~\ref{fig:cwa}.}}
  \label{fig:supp-cwa-gsm8k}
\end{figure}

\FloatBarrier

\subsection{Chance baselines and relation to CAPA}
\label{app:capa}

Four baselines are computed. The per-question leave-pair-out baseline used in the primary analysis estimates each item's wrong-answer collision rate from the remaining models and then aggregates these expectations to the model-pair level. Pair-independence and pair-empirical baselines are comparators; a pooled-population baseline is retained only as a negative control because it does not condition on the question. This per-question-to-pair construction is used throughout the CWA analysis. \rev{The difference-form excess, $\CWA_{ij}-\CWA^{\mathrm{exp}}_{ij}$, is the numerator of $\kappa_{ij}$ without the division by $1-\CWA^{\mathrm{exp}}_{ij}$.} Difference-form excess and chance-corrected $\kappa$ rank pairs almost identically (Spearman 0.988 over benchmark means); eight of ten chance-correction denominators, $1-\CWA^{\mathrm{exp}}$, lie in \rev{$[0.90,0.99]$}, with HLE-MC and MuSR the exceptions. \rev{The denominator is the largest excess above chance that a pair can reach, and it differs across benchmarks. Its mean is 0.65 on HLE-MC, where most models are wrong on most questions, and 0.81 on MuSR, where two wrong answers usually coincide by chance. The same difference-form excess is therefore a larger share of what is possible on these two benchmarks than on the others. Dividing by the denominator puts every benchmark on one scale, from 0 at chance to 1 at complete agreement on wrong answers, so we use $\kappa$ as the primary outcome and the difference form as a companion.}

For completeness, consider a held-out pair $(i,j)$. Let $b_{-ij,q}(a)=\sum_{k\notin\{i,j\}}c_{kq}(a)$ count wrong epochs from the remaining models that select $a\neq y_q$, and let $B_{-ij,q}=\sum_{a\neq y_q}b_{-ij,q}(a)$. On questions for which both held-out models have a wrong epoch and $B_{-ij,q}\geq2$, the collision probability is
\begin{equation}
h_{ijq}=\frac{\sum_{a\neq y_q}b_{-ij,q}(a)[b_{-ij,q}(a)-1]}
{B_{-ij,q}(B_{-ij,q}-1)},
\qquad
\bar h_{-ij}=\frac{1}{|\mathcal{Q}_{ij}|}
\sum_{q\in\mathcal{Q}_{ij}}h_{ijq}.
\end{equation}
The baseline in Equation~\ref{eq:kappa} multiplies $\bar h_{-ij}$ by $p_i p_j/(p_i+p_j-p_i p_j)$, where each $p$ is the fraction of parsed epochs that are wrong across questions with parsed responses from both models.

CAPA applies the same algebraic chance correction but targets overall prediction agreement, counting matches whether models are correct or wrong \citep{goel_great_2025}. CWA instead conditions on at least one error and targets agreement on the same wrong answer. \rev{Here CAPA is computed on the same epoch grid as CWA, counting every pairing of the two models' epochs on a question, with its uniform-distractor baseline, whereas CWA uses the question-specific leave-pair-out baseline.} Figure~\ref{fig:supp-capa} shows that the measures are related but not equivalent. Chance correction increases their rank agreement on seven of the eight fixed-option benchmarks, while HLE-MC moves in the opposite direction. \rev{After correction the two order model pairs almost identically, but CAPA $\kappa$ lies above CWA $\kappa$ for the same pair. Its baseline assumes that wrong answers spread evenly over the $k-1$ wrong options, whereas two different models' wrong answers coincide up to 3.2 times as often as $1/(k-1)$. CWA's leave-pair-out baseline measures this rate, so CWA $\kappa$ is zero when a pair shares wrong answers only as often as the questions' distractors produce. With one modal answer per question, as in CAPA's discrete form, the chance-corrected forms agree at Spearman 0.86 to 0.98, and modal and epoch-native CAPA at 0.90 to 0.99. CWA remains the outcome for three reasons. It needs only a rule for when two answers match, so it also covers the numeric-answer benchmarks, where CAPA is undefined. Its zero rests on the measured collision rate, so its level can be read as excess correlated failure and compared across benchmarks. And it extends to ensembles, since $\CWA=J\times E$ and the CFR of Equation~\ref{eq:cfr} is $E$ for two models (Appendix~\ref{app:joint-agreement}).}

\subsection{Correlated failure across empirical quantiles \rev{of each measure}}
\label{app:dose}

\begin{figure}[t]
  \centering
  \includegraphics[width=0.9\linewidth]{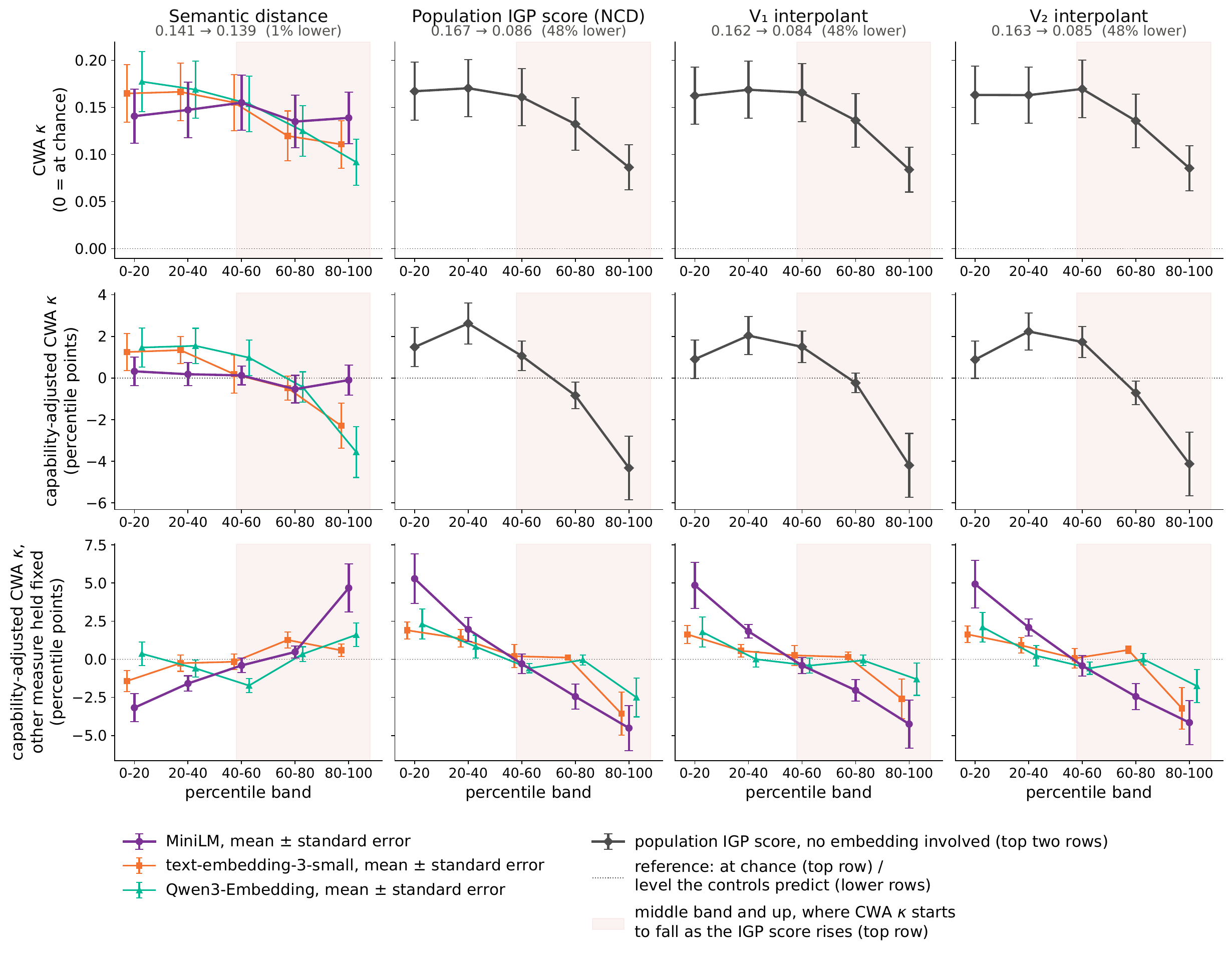}
  \caption{\rev{\textbf{Correlated failure across the range of each measure.} Model pairs are binned into quintiles of semantic distance and of the population IGP score computed from NCD, $V_1$ and $V_2$. Top: mean CWA $\kappa$. Middle: capability-adjusted CWA $\kappa$ (capability level and gap), in percentile points. Bottom: capability-adjusted CWA $\kappa$ with the other measure also held fixed, over bands of each measure with the same controls removed (the population IGP score for semantic distance, semantic distance for each IGP score). Lines show means and standard errors across the ten benchmarks, with one colour per embedding (MiniLM purple, text-embedding-3-small orange, Qwen3-Embedding teal). The IGP panels of the top two rows do not depend on the embedding and are grey. The panels of each row share one axis. Shading marks the middle band and up, where CWA $\kappa$ starts to fall as the IGP score rises (top row). On its own, semantic distance is flat under MiniLM and declines under the two stronger embeddings. With the controls removed it does not decline under any embedding, whereas CWA $\kappa$ falls across the bands of every IGP score in all three rows.}}
  \label{fig:dose}
\end{figure}

\rev{Figure~\ref{fig:dose} bins model pairs into quintiles of each measure within the evaluated model population. On its own (Figure~\ref{fig:dose}, top two rows), the population IGP score tracks less correlated failure. Mean CWA $\kappa$ falls from 0.167 in the lowest band to 0.086 in the highest, 48\% lower, and still declines within the top band. After capability adjustment the shape is the same, with a decline of 5.38 percentile points from the middle band to the highest. The scores computed from $V_1$ and $V_2$ give 0.162 to 0.084 and 0.163 to 0.085, so the pattern does not depend on \rev{the choice of $V_p$}. Bands with low IGP scores remain above the chance line, so low IGP diversity corresponds to excess correlated failure rather than merely the absence of a benefit.}

\rev{Semantic distance on its own depends on the embedding. Under MiniLM it is flat, with mean $\kappa$ 0.141 in the lowest band and 0.139 in the highest. Under text-embedding-3-small and Qwen3-Embedding it falls from 0.165 to 0.111 and from 0.177 to 0.092.}

\rev{The bottom row of Figure~\ref{fig:dose} holds capability and the other measure fixed, as the partial correlations of Section~\ref{sec:conditional} do. From the lowest to the highest band of the population IGP score, the adjusted CWA $\kappa$ falls by 9.8 percentile points under MiniLM (95\% $t$-interval $[-16.2,-3.4]$, negative on 10 of 10 benchmarks), 5.5 under text-embedding-3-small ($[-9.8,-1.2]$, 9 of 10) and 4.8 under Qwen3-Embedding ($[-9.4,-0.3]$, 8 of 10). Across bands of semantic distance it does not fall under any embedding. It rises by 7.9 points under MiniLM ($[+2.5,+13.2]$) and changes by $+2.0$ and $+1.2$ points under the other two, with intervals that include zero. Once the overlap between the two measures is removed, the bands match the partial correlations under every embedding.}

\rev{The range is population-relative. Raw NCD covers only $[0.48,0.86]$ of its nominal $[0,1]$ interval, and the upper band is not an absolute maximum of IGP diversity.}

\subsection{Joint failure and error agreement}
\label{app:joint-agreement}

\begin{revblock}
Pairwise CWA conditions on at least one model being wrong, and the ensemble CFR on every member being wrong. Both come from one construction. Let $w_{iq}=n_{iq}-c_{iq}(y_q)$ count model $i$'s wrong epochs on question $q$. Joint failure $J$ is the probability that both models are wrong given that at least one is, and error agreement $E$ is the probability that they give the same wrong answer given that both are wrong:
\begin{equation}
\begin{gathered}
J_{ij}=\frac{\sum_q w_{iq}w_{jq}}{\sum_q\left[n_{iq}n_{jq}-c_{iq}(y_q)c_{jq}(y_q)\right]},
\qquad
E_{ij}=\frac{\sum_q\sum_{a\neq y_q}c_{iq}(a)c_{jq}(a)}{\sum_q w_{iq}w_{jq}},\\
\CWA_{ij}=J_{ij}\,E_{ij}.
\end{gathered}
\label{eq:joint-agreement}
\end{equation}
The identity is exact. $E$ is the CFR of Equation~\ref{eq:cfr} for two models, and the unconditional convergent-failure rate of Appendix~\ref{app:ensemble-outcome} is $P(\text{all wrong})\times\CFR$. Each conditioning event matches its question. For pairs, the question is whether a second model exposes the first one's errors. For five models, it is the worst case, in which all five are wrong.

\afterpage{%
\begin{landscape}
\begin{figure}[p]
  \centering
  \includegraphics[width=\linewidth]{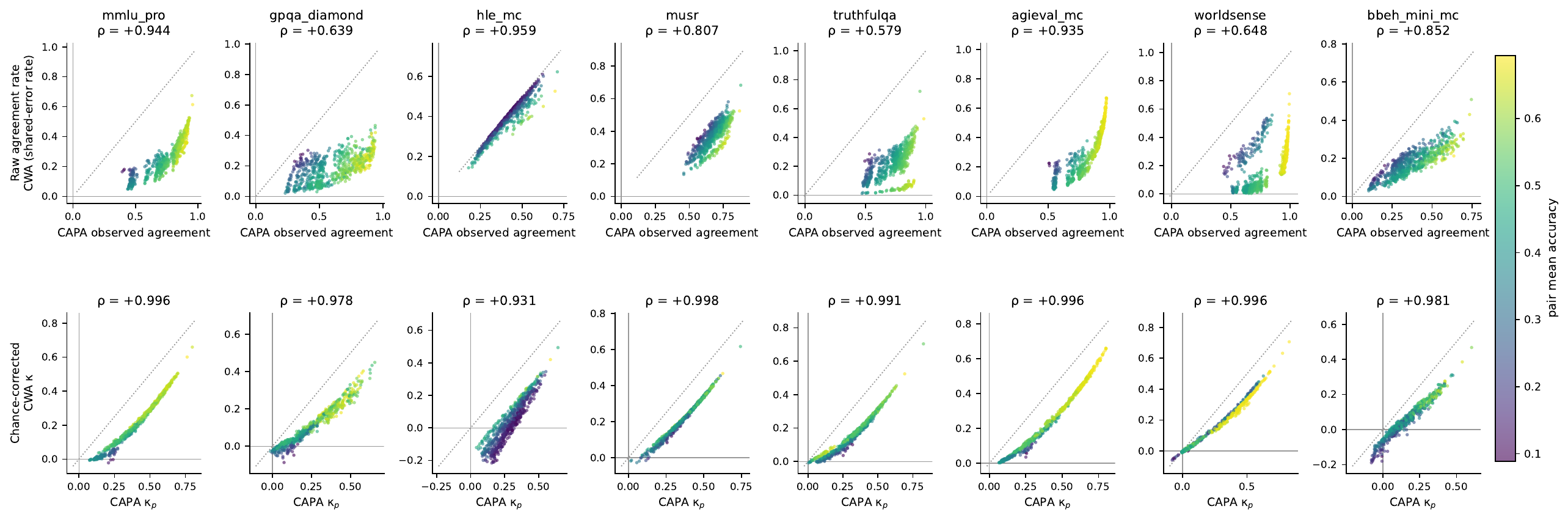}
  \caption{\textbf{CWA and CAPA measure related but distinct forms of agreement.} Each point is a model pair on one of eight fixed-option benchmarks\rev{, with both measures computed over every pairing of the two models' epochs}; numeric-answer benchmarks are omitted because CAPA is undefined. The top row compares raw CAPA agreement with CWA's shared-error rate, and the bottom row compares their chance-corrected forms. Points are coloured by pair mean accuracy, dotted lines show $y=x$, and panel headings report Spearman $\rho$. Raw CWA generally lies below CAPA because it excludes both-correct agreement. Rank agreement spans \rev{$0.58$--$0.96$} before correction and \rev{$0.93$--$1.00$} after correction; HLE-MC is the exception, decreasing from \rev{$0.959$ to $0.931$. After correction CAPA $\kappa$ lies above CWA $\kappa$, because its chance term assumes that wrong answers spread evenly over the options.}}
  \label{fig:supp-capa}
\end{figure}
\end{landscape}
}

Each part has its own chance correction, $\kappa=(\text{observed}-\text{expected})/(1-\text{expected})$. For $J$ the expected value is $p_ip_j/(p_i+p_j-p_ip_j)$, the first factor of $\CWA^{\mathrm{exp}}_{ij}$ in Equation~\ref{eq:kappa}. It assumes that a model's error rate is the same on every question, so question difficulty stays in $J$. For $E$ the expected value is the collision probability $h_{ijq}$ of the other models' wrong epochs (Appendix~\ref{app:outcomes}), averaged over questions with the both-wrong pairings $w_{iq}w_{jq}$ as weights. The $\kappa$ of $E$ needs at least 30 questions on which both models have a wrong epoch and the other models give at least two wrong answers. The row ``$E$, at least 15 questions'' of Table~\ref{tab:rev-joint-agreement} lowers this threshold to 15.

Table~\ref{tab:rev-joint-agreement} gives the association of the population IGP score with each part. For pairs scored under the ensemble's rule, that is on $E$, the estimate is $-0.141$, $-0.110$ and $-0.096$ under MiniLM, TE3-small and Qwen3. For $J$ on the same pairs it is $-0.200$, $-0.152$ and $-0.101$. The sign therefore holds for pairs scored under the ensemble's rule under all three embeddings, and the association is in both parts. CWA is the primary pairwise outcome because its conditioning event keeps the questions on which the second model is right, and conditioning on both models being wrong drops them.
\end{revblock}

\begin{table}[htbp]
  \revcolor
  \caption{\textbf{Population IGP score against the two parts of CWA.} Mean partial Spearman correlation over ten benchmarks, controlling capability level, capability gap and semantic distance, with its 95\% interval and the number of benchmarks with a negative estimate. Each part is chance-corrected. $J$ is computed on the pairs for which $E$ can be estimated.}
  \label{tab:rev-joint-agreement}
  \centering
  \resizebox{\linewidth}{!}{%
  \begin{tabular}{llll}
    \toprule
    Outcome & MiniLM & TE3-small & Qwen3 \\
    \midrule
    CWA $\kappa$ & $-0.216$ $[-0.309,-0.122]$ 10/10 & $-0.128$ $[-0.207,-0.048]$ 9/10 & $-0.106$ $[-0.199,-0.013]$ 8/10 \\
    Joint failure $J$ & $-0.200$ $[-0.312,-0.087]$ 9/10 & $-0.152$ $[-0.260,-0.043]$ 9/10 & $-0.101$ $[-0.240,+0.039]$ 9/10 \\
    Error agreement $E$ & $-0.141$ $[-0.249,-0.033]$ 9/10 & $-0.110$ $[-0.166,-0.053]$ 10/10 & $-0.096$ $[-0.176,-0.016]$ 9/10 \\
    $E$, at least 15 questions & $-0.174$ $[-0.281,-0.067]$ 9/10 & $-0.119$ $[-0.191,-0.046]$ 9/10 & $-0.115$ $[-0.194,-0.035]$ 9/10 \\
    \bottomrule
  \end{tabular}}
\end{table}

\FloatBarrier

\subsection{Chance term with aligned question weights}
\label{app:aligned-kappa}

\begin{revblock}
In Equation~\ref{eq:kappa} the observed CWA pools questions by their number of epoch tuples with at least one model wrong, whereas the chance term $\bar h_{-ij}$ is an unweighted mean over questions. As a sensitivity analysis, we recompute CWA $\kappa$ with the chance term weighted the same way as the observed term, and call the result the aligned CWA $\kappa$. Let $W_{ijq}=n_{iq}n_{jq}-c_{iq}(y_q)c_{jq}(y_q)$, and let $\mathcal{W}_{ij}$ be the questions that the observed CWA pools, those with $W_{ijq}>0$, on which $h_{ijq}$ can be estimated because the other models give at least two wrong answers. Then
\begin{equation}
\bar h^{\mathrm{aligned}}_{-ij}=\frac{\sum_{q\in\mathcal{W}_{ij}}W_{ijq}\,h_{ijq}}{\sum_{q\in\mathcal{W}_{ij}}W_{ijq}},
\label{eq:aligned-h}
\end{equation}
so only the weights and the question set of the collision mean change. The set $\mathcal{W}_{ij}$ needs one of the two models to be wrong, whereas the set $\mathcal{Q}_{ij}$ of Appendix~\ref{app:outcomes} needs both. On AGIEval a pair has 212 questions on average in $\mathcal{W}_{ij}$ and 76 in $\mathcal{Q}_{ij}$. The two versions rank model pairs almost identically, with Spearman correlations of 0.964 to 1.000 across benchmarks, and mean $\kappa$ moves by at most 0.010, on Humanity's Last Exam. The main estimate moves by at most 0.002 under every embedding and the number of negative benchmarks does not change (Table~\ref{tab:rev-aligned-estimate}). With no semantic distance in the model the estimate moves by 0.007, from $-0.148$ to $-0.141$, and the number of negative benchmarks falls from 9 to 8. CWA $\kappa$ of Equation~\ref{eq:kappa} keeps the form of CAPA's chance correction, the marginal error rates times an unweighted question mean of the collision probability, so that the two can be compared directly (Appendix~\ref{app:capa}). Weighting the chance term like the observed term changes no result, so CWA $\kappa$ remains the primary outcome.
\end{revblock}

\begin{table}[htbp]
  \revcolor
  \caption{\textbf{Population IGP score beyond semantic distance, under CWA $\kappa$ and the aligned CWA $\kappa$.} Mean partial Spearman correlation over ten benchmarks with its 95\% interval and the number of benchmarks with a negative estimate.}
  \label{tab:rev-aligned-estimate}
  \centering
  \begin{tabular}{lll}
    \toprule
    Embedding & CWA $\kappa$ & Aligned CWA $\kappa$ \\
    \midrule
    MiniLM & $-0.216$ $[-0.309,-0.122]$ 10/10 & $-0.214$ $[-0.313,-0.116]$ 10/10 \\
    TE3-small & $-0.128$ $[-0.207,-0.048]$ 9/10 & $-0.126$ $[-0.208,-0.044]$ 9/10 \\
    Qwen3 & $-0.106$ $[-0.199,-0.013]$ 8/10 & $-0.106$ $[-0.204,-0.009]$ 8/10 \\
    \bottomrule
  \end{tabular}
\end{table}

\FloatBarrier

\clearpage
\section{Ensemble Selection at Fixed Semantic Distance}
\label{app:ensemble}

\begin{addblock}
This appendix gives the definitions, kernel, matching procedure, and inference behind Section~\ref{sec:ensemble-method} and Figure~\ref{fig:ensemble}, with enough detail to reproduce the analysis from the per-model benchmark records and the two $38\times38$ \rev{measure} matrices. Numbers are for the \rev{population IGP score computed from NCD} ($V_\infty$) unless stated; \rev{results for $V_1$ and $V_2$} are \rev{shown} in Appendix~\ref{app:ensemble-results}. At the ensemble level the chance-corrected $\kappa_S$ of Equation~\ref{eq:cfr} is the primary outcome and the difference form $\CFR_S-\CFR^{\mathrm{exp}}_S$ is the companion, matching the pairwise convention. \rev{In Subsections~\ref{app:ensemble-eligibility} to~\ref{app:ensemble-figures}, controls are capability-matched ensembles, and a pinned arm holds the other measure within a band.}

\subsection{Enumeration and eligibility}
\label{app:ensemble-eligibility}

For $k=5$, every one of the $\binom{38}{5}=501{,}942$ ensembles is enumerated per benchmark key and per \rev{version of the IGP score} and scored on the epoch-native outcomes below. Two estimability rules define the eligible population, and both are applied to the selection universe and to every control pool. First, an ensemble's chance-corrected outcomes are undefined when fewer than 10 items are estimable for it. Second, only ensembles with at least 30 estimable items enter, the same threshold as the pairwise conditioning gate. A key is analysed when at least $10T=2{,}000$ eligible ensembles remain, where $T=200$ is the selection size. Table~\ref{tab:ensemble-eligible} gives the eligible counts, which are identical across the three variants because eligibility depends on outcomes and capability only. Eleven keys are analysed, collapsing to nine source benchmarks; the three BBEH-mini answer spaces pool into one source, question-weighted 180/120/160. GSM8K is excluded by this guard, not by choice, and remains in the ten-benchmark pairwise analysis.

\begin{table}[htbp]
  \caption{\textbf{Eligible five-model ensembles per benchmark key} out of 501,942 enumerated.}
  \label{tab:ensemble-eligible}
  \centering
  \begin{tabular}{lr@{\hspace{2em}}lr}
    \toprule
    Key & Eligible & Key & Eligible \\
    \midrule
    HLE & 501,942 & TruthfulQA & 376,992 \\
    MMLU-Pro & 501,942 & BBEH-mini (numeric) & 205,200 \\
    MuSR & 501,941 & GPQA-Diamond & 97,063 \\
    AGIEval & 501,280 & AIME & 16,813 \\
    BBEH-mini (other) & 500,435 & WorldSense & 14,522 \\
    BBEH-mini (MC) & 490,908 & GSM8K & 1,414 (excluded) \\
    \bottomrule
  \end{tabular}
\end{table}

\subsection{Outcome}
\label{app:ensemble-outcome}

On item $q$, with $n_m$ parsed epochs for member $m$, let $T_q=\prod_m n_m$ be the number of epoch tuples, $A_q$ the tuples in which every member is wrong, and $C_q$ the tuples in which every member is wrong and all give the same wrong answer. The unconditional convergent-failure rate is $\sum_q C_q/\sum_q T_q$ and the conditional rate is $\CFR_S=\sum_q C_q/\sum_q A_q=P(\text{same}\mid\text{all wrong})$. The conditional form is primary for the same reason CWA is conditional at the pairwise level. It separates \emph{how} an ensemble failed from \emph{whether} it failed, so a diverse ensemble is not rewarded merely for being accurate.

On each item where every member has at least one wrong epoch, the leave-ensemble-out population is the pooled wrong epochs of the other $M-k$ models, $N_q$ in total with $n_x$ on wrong answer $x$. The chance that $k$ draws without replacement all land on one wrong answer is the hypergeometric collision probability $b_q$ of Equation~\ref{eq:cfr}. An item is estimable when every member has a wrong epoch, $N_q\geq8$, and $\binom{N_q}{k}>0$. With \rev{$\mathcal{Q}_S$} the estimable set,
\begin{equation}
\CFR_S=\frac{\sum_{q\in\rev{\mathcal{Q}_S}}C_q}{\sum_{q\in\rev{\mathcal{Q}_S}}A_q},
\qquad
\CFR^{\mathrm{exp}}_S=\frac{\sum_{q\in\rev{\mathcal{Q}_S}}A_q\,b_q}{\sum_{q\in\rev{\mathcal{Q}_S}}A_q},
\qquad
\kappa_S=\frac{\CFR_S-\CFR^{\mathrm{exp}}_S}{1-\CFR^{\mathrm{exp}}_S}.
\label{eq:cfr-appendix}
\end{equation}
Observed and expected terms share the item set and the all-wrong tuple weights, so the difference is zero whenever observed convergence equals chance item by item. This alignment is stronger than the pairwise baseline of Equation~\ref{eq:kappa}, whose $\bar h_{-ij}$ is an unweighted item mean. \rev{Appendix~\ref{app:aligned-kappa} gives the pairwise $\kappa$ with the same alignment.} At one epoch per cell the construction reduces to the modal leave-ensemble-out baseline, and at $k=2$ it reduces to the pairwise per-item collision rate $\sum_x c_x(c_x-1)/[N(N-1)]$ of Appendix~\ref{app:outcomes}.

\subsection{Kernel, Vendi score, and log-determinant}
\label{app:ensemble-kernel}

Let $D$ be the $38\times38$ \rev{matrix of the population IGP score} $\spop$, or the non-negative semantic distance matrix $1-\text{similarity}$. With one bandwidth per measure fixed from the median absolute off-diagonal entry,
\begin{equation}
\gamma=\frac{1}{\operatorname{median}_{i<j}|D_{ij}|+\rev{\eta}},
\end{equation}
\rev{where $\eta=10^{-9}$ guards against division by zero; the same $\eta$ keeps the logarithm finite in Equation~\ref{eq:vendi}. Each} model $i$ receives a \rev{feature vector with two channels, each a decaying function of its scores against all $M$ models,} and is normalised to unit length,
\begin{equation}
\phi_i=\Big[e^{-\gamma\max(D_{ij},0)}\Big]_{j=1}^{M}
\;\Vert\;
\Big[e^{-\gamma\max(-D_{ij},0)}\Big]_{j=1}^{M},
\qquad
u_i=\frac{\phi_i}{\lVert\phi_i\rVert_2},
\qquad
K_{ij}=u_i^{\top}u_j,
\label{eq:ensemble-kernel}
\end{equation}
where $\Vert$ denotes concatenation. \rev{We call $u_i$ the feature vector of model $i$ and $K$ the ensemble kernel. The kernel follows the D2KE construction that \citet{hamzi_bridging_2025} apply to NCD, which represents each model by decaying functions of its scores against a set of landmarks, here all $M$ models. Each model's feature vector is a profile of its scores against every model, in two halves. One half records how much closer than expected the two models' responses are, the other how much farther; both read 1 for a pair at its expected value and shrink as the gap grows. Two models have a high $K_{ij}$ when their profiles match, that is, when they are unusually close to the same models and unusually far from the same models. Keeping the halves separate preserves the sign, which carries the meaning. Recording only the size of a score, as $|D|$ or a Gaussian kernel on $D^2$ would, makes a pair much closer than expected look like one much farther than expected, and 329 of 703 pairs are below zero. D2KE needs a non-negative input, and the two halves provide one without moving the zero point away from the expected value, as shifting all scores upward would.} $K$ is positive semidefinite with unit diagonal by construction. It is a chosen geometry for \rev{a signed score}, not a claim that the \rev{IGP score} matrix is itself a similarity. On non-negative semantic distance the negative channel is a constant block and the same construction is used. Because the pool is the full fleet on every key, the kernels are identical across keys: $\gamma_{\IGP}=89.46$ (median $|D|\approx0.011$) and $\gamma_{\mathrm{sem}}=3.23$ (median $\approx0.31$).

For a candidate $S$, take the $5\times5$ principal submatrix $K_S$ with eigenvalues $\lambda_1,\ldots,\lambda_5\geq0$; round-off negatives are clipped and a materially indefinite submatrix is treated as an error. Then
\begin{equation}
\operatorname{logdet}(S)=\sum_r\log(\lambda_r+\rev{\eta}),
\qquad
p_r=\frac{\lambda_r}{\sum_s\lambda_s},
\qquad
\operatorname{Vendi}(S)=\exp\Big(-\sum_r p_r\log p_r\Big),
\label{eq:vendi}
\end{equation}
with $0\log0=0$. Vendi is the exponential of the von Neumann entropy of the trace-normalised kernel, the effective rank of the ensemble in feature space \citep{friedman_vendi_2023}, and lies in $[1,5]$; it receives no ridge. The log-determinant is the determinantal-point-process volume objective \citep{kulesza_determinantal_2012} and is evaluated as an alternative selector only. Nothing is recentred or refitted per candidate, and no outcome enters any score. Every eligible ensemble is scored by exact batched eigendecomposition, and selection is the exact top $T=200$ with ties broken in enumeration order.

Three measured properties qualify the reading of Vendi. First, it is not a monotone function of \rev{the} mean \rev{pairwise value}: Spearman agreement with \rev{the} mean pairwise \rev{value} is 0.37--0.48 for \rev{the IGP score} and 0.22--0.43 for semantic distance across the 33 key-by-variant pools. Second, on this unit-diagonal kernel, uncentred Vendi is rank-equivalent to the trace of the centred submatrix (Spearman 0.98 for \rev{the IGP score}, 1.00 for semantic \rev{distance}). It is therefore a \emph{magnitude} score of how spread the members are, not a score of how evenly that spread is distributed across directions. Third, the semantic kernel is close to rank one under this bandwidth, so semantic Vendi has a narrow range (top-200 cut-offs\rev{, the Vendi score of the 200th selected ensemble,} 1.010--1.020 against control means 1.009--1.014, where \rev{the IGP score's} cut-offs are 1.62--2.10 against 1.52--1.84). The semantic ranking remains well defined, since semantic-Vendi selections reach the 87th--95th percentile of mean semantic distance, and the mean-\rev{value} selector, \rev{whose range is not narrowed in this way}, gives the same sign pattern under the pin design (\rev{Figure~\ref{fig:supp-ensemble-mean}}).

\begin{revblock}
The cut-offs in the paragraph above belong to the design of Figure~\ref{fig:ensemble}, which selects the top 200 within each benchmark's eligible pool (Appendix~\ref{app:ensemble-eligibility}) for each of the three \rev{versions of the IGP score}. For the population IGP score they run from 1.757 on AIME to 2.065 on the four keys whose eligible pool is close to the full enumeration. The vote analysis of Appendix~\ref{app:ensemble-votes} applies no estimability rule. It selects the top 200 among all 501,942 ensembles, so one cut-off per measure holds on every benchmark: 2.065 for the population IGP score, 2.087 for the pair-specific IGP score, and 1.020, 1.025 and 1.031 for semantic distance under MiniLM, TE3-small and Qwen3.

Figure~\ref{fig:rev-vendi-scores} shows each Vendi score over all 501,942 five-model ensembles. No score depends on the benchmark. The scores on the two IGP kernels spread over roughly 1.2 to 2.2, whereas the scores on the semantic kernels stay within a few hundredths of 1 under all three embeddings, which is the near rank-one behaviour described above. \revr{These levels do not compare diversity across measures. The kernel compares models by their distances to all models, so a measure on which every pair is about equally far apart, as with semantic distance, scores close to 1 however distinct the models are. We therefore use each Vendi score only to rank ensembles within its own measure.}
\end{revblock}

\begin{figure}[htbp]
  \centering
  \includegraphics[width=\linewidth]{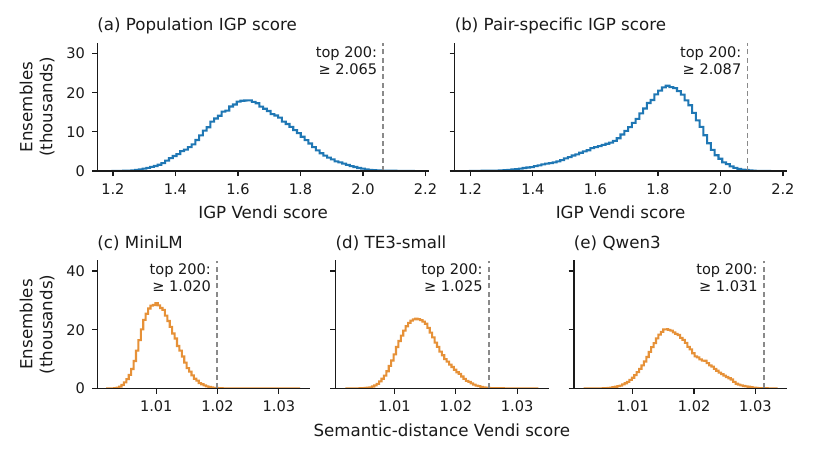}
  \caption{\revcolor{}\textbf{Vendi scores of all ensembles.} Each panel is a histogram of one Vendi score over the 501,942 five-model ensembles. (a), (b) Kernels built on the population and the pair-specific IGP score. (c)--(e) Kernels built on semantic distance under MiniLM, TE3-small and Qwen3. The Vendi score is the effective number of distinct members \revr{under each kernel} and can range from 1 to 5. \revr{Its level is not comparable across measures (Appendix~\ref{app:ensemble-kernel}).} The dashed line marks the 200th highest score, the cut-off for the selected ensembles of Appendix~\ref{app:ensemble-votes}: 2.065 in (a), 2.087 in (b), and 1.020, 1.025 and 1.031 in (c) to (e). Panels (a) and (b) share both axes, as do panels (c) to (e).}
  \label{fig:rev-vendi-scores}
\end{figure}

\subsection{Arms and pin bands}
\label{app:ensemble-arms}

For each key, variant, and selector we run four arms. Two are unconstrained: the top 200 \rev{on the IGP score} and the top 200 \rev{on semantic distance}. Two are pinned: the top 200 \rev{on the IGP score} among ensembles whose \emph{mean semantic distance} lies in a band, and the top 200 \rev{on semantic distance} among ensembles whose \emph{mean \rev{IGP score}} lies in a band. Bands are $[0,20)$, $[20,40)$, $[40,60)$, $[60,80)$, and $[80,100]$ in population percentiles of the eligible pool's mean \rev{pairwise value} on the held measure. The pin is always on \rev{the} mean \rev{pairwise value}, also for spectral selectors, so the held quantity and the dose annotations in Figure~\ref{fig:ensemble} (median percentile of the free measure's mean \rev{pairwise value} among selected ensembles) share one scale. A band with fewer than 2,000 candidates is skipped; none is on the production enumeration.

\subsection{Capability-matched controls}
\label{app:ensemble-matching}

\rev{The ensemble analysis holds fixed the same two capability terms as the pairwise analysis, level and gap, but by matching instead of regression. The two terms are an ensemble's mean member accuracy and the mean absolute accuracy gap between its members, each divided by its SD over the eligible pool. A control is an eligible ensemble within $\tau$ SD of the selected ensemble on both terms. The window $\tau$ starts at 0.5 and grows by a factor of 1.5 until at least 50 controls qualify, but never beyond 2.0. We also match the gap because, in a check, matching on mean accuracy alone attributed to diversity an effect of the spread in accuracy between members. For pinned arms, a control must also lie in the same band, and its mean on the held measure must be within 0.1 population SD of the selected ensemble's. We apply these two conditions before the capability window and never widen them. The selected ensemble meets all of its own conditions, so it is always in its own pool, as the definition of a matched random ensemble requires. This pulls each difference towards zero by a factor $1-1/n$ for a pool of $n$ ensembles, at most about 2.5\% in the smallest pools, on AIME. When no other ensemble matches, the pool holds only the selection itself, so its difference is zero, and we count these cases. We aim for 50 controls but keep a selection that has fewer, so we keep all 200 selections. For each cell we report the smallest, median and largest pool, the number of pools below 50, and the number in which no other ensemble matches. On the held measure, the controls differ from the selected ensembles by at most 0.008 SD in every pinned Vendi cell. On capability they differ by a median of 0.06 to 0.08 SD and at most 0.19 SD. \revr{The difference has a direction. Within a window fewer ensembles lie above a selected ensemble's accuracy gap than below it, so the selected ensembles have the larger gap in 93\% of the pinned Vendi cells (median 0.09 SD). CFR $\kappa$ falls as the gap grows, so we remove the difference by regression, as for the vote outcomes (Appendix~\ref{app:ensemble-inference}).} The median pool for a Vendi selection ranges from 79 ensembles (AIME, 0--20 band) to 6,243 (AGIEval, 60--80 band). Matching \revr{and the adjustment fix} mean member accuracy, accuracy gap and the held measure. The ensembles can still differ in properties that were not measured, so no band contrast identifies a causal effect.}

\subsection{Estimand and inference}
\label{app:ensemble-inference}

For one cell, with $y_s$ the outcome of selection $s$, \revr{$z_s$ its two matching coordinates} and $\mathcal{C}(s)$ its \rev{controls},
\begin{equation}
\revr{\Delta=\frac{1}{T}\sum_{s=1}^{T}\Big[y_s-\frac{1}{|\mathcal{C}(s)|}\sum_{c\in\mathcal{C}(s)}\big\{y_c+\widehat{\beta}^{\top}(z_s-z_c)\big\}\Big],}
\label{eq:ensemble-delta}
\end{equation}
\revr{where $\widehat{\beta}$ holds the two slopes of an ordinary least-squares fit of $y$ on an intercept and $z$ over the selected ensembles and their controls, each counted once. Each control is shifted to its selected ensemble's capability, which removes the part of the difference that is linear in the two matching coordinates and with it the capability difference that matching leaves (Appendix~\ref{app:ensemble-matching}). The estimator follows the bias-corrected matching estimator of \citet{abadie_bias-corrected_2011} with one departure. The selected ensembles enter the fit.}
\rev{A negative $\Delta$ means that the selected ensembles give the same wrong answer less often than capability-matched ensembles at the same level of the held measure. The outcome counts only questions on which all five members are wrong, and it is corrected for chance. \revr{We base inference on the variation across benchmarks.} We first combine the keys of each source benchmark into one $\Delta$, weighting the parts of BBEH-mini by their number of questions. We then average the nine sources with equal weight and report the mean, its 95\% $t$-interval ($t_{0.975,8}=2.31$) and the median. We never pool selector families and exclude no benchmark because of its result. We apply no correction for multiple comparisons across the 90 pinned cells per outcome (5 bands, 3 variants, 2 arms and 3 selectors). We Holm-adjust the vote outcomes of Appendix~\ref{app:ensemble-votes} separately (Appendix~\ref{app:vote-inference}).}

\subsection{\rev{Results by selector and $V_p$}}
\label{app:ensemble-results}
\label{app:ensemble-figures}

\rev{Figures~\ref{fig:supp-ensemble-vendi}--\ref{fig:supp-ensemble-mean} show the Vendi, log-determinant and mean-value selectors in the layout of Figure~\ref{fig:ensemble}, for the population IGP score computed from NCD ($V_\infty$), $V_1$ and $V_2$. Table~\ref{tab:ensemble-vendi} gives the Vendi values. With the other measure held within a band, the IGP arm is below its controls in every band and the semantic arm is above them in every band, under all three selectors and all three versions of the IGP score. The unconstrained arms do not separate the two measures under the mean-value selector. Both are below their controls \revr{($-0.053$ $[-0.112,+0.007]$ and $-0.043$ $[-0.096,+0.010]$ on $V_\infty$), and the semantic arm is below them on six of the nine sources}, as the collinearity of the two measures predicts. The difference form $\CFR_S-\CFR^{\mathrm{exp}}_S$ gives the same sign as $\kappa_S$ in \revr{33} of the 36 means on $V_\infty$ (three selectors, each with two unconstrained arms and ten pinned bands).}

\rev{Per source, the Vendi IGP arm's contrasts in the 60--80 band are \revr{WorldSense $-0.067$, HLE $-0.049$, AGIEval $-0.024$, TruthfulQA $-0.007$, GPQA-Diamond $-0.004$, BBEH-mini $-0.004$, MMLU-Pro $-0.003$, AIME $+0.002$ and MuSR $+0.007$}. In the unconstrained Vendi IGP arm, WorldSense (\revr{$-0.191$}) and HLE (\revr{$-0.107$}) make up \revr{four fifths} of the sum over sources, and the mean over the other seven is \revr{$-0.009$}. WorldSense's contrasts are the largest on this scale. Five wrong answers on WorldSense usually coincide by chance, so its denominator $1-\CFR^{\mathrm{exp}}_S$ is the smallest (mean 0.11 over eligible ensembles, against 0.22 to 0.99 for the other sources). AIME's contrasts are an order of magnitude smaller than the other sources' on both scales.}

\begin{table}[htbp]
  \caption{\textbf{Vendi selector, primary $\kappa_S$ outcome.} Selected-minus-control $\Delta$\revr{ after the capability adjustment of Equation~\ref{eq:ensemble-delta}}, mean over nine sources with 95\% $t$-interval. \rev{The semantic arm's band means are given as a range.}}
  \label{tab:ensemble-vendi}
  \centering
  \resizebox{\linewidth}{!}{%
  \begin{tabular}{llccc}
    \toprule
    & & \multicolumn{3}{c}{\rev{Population IGP score computed from}} \\
    \cmidrule(lr){3-5}
    Arm & Band & \rev{NCD ($V_\infty$)} & $V_1$ & $V_2$ \\
    \midrule
    \rev{IGP} Vendi, semantic held & 0--20 & \revr{$-0.015$ $[-0.039,+0.009]$} & \revr{$-0.018$ $[-0.050,+0.014]$} & \revr{$-0.015$ $[-0.041,+0.012]$} \\
     & 20--40 & \revr{$-0.018$ $[-0.045,+0.008]$} & \revr{$-0.017$ $[-0.046,+0.011]$} & \revr{$-0.016$ $[-0.046,+0.013]$} \\
     & 40--60 & \revr{$-0.018$ $[-0.039,+0.002]$} & \revr{$-0.014$ $[-0.041,+0.013]$} & \revr{$-0.015$ $[-0.037,+0.008]$} \\
     & 60--80 & \revr{$-0.017$ $[-0.036,+0.003]$} & \revr{$-0.014$ $[-0.034,+0.005]$} & \revr{$-0.015$ $[-0.032,+0.002]$} \\
     & 80--100 & \revr{$-0.019$ $[-0.047,+0.009]$} & \revr{$-0.019$ $[-0.049,+0.010]$} & \revr{$-0.020$ $[-0.051,+0.010]$} \\
    Semantic Vendi, \rev{IGP} held & all bands & \revr{$+0.020$ to $+0.036$} & \revr{$+0.014$ to $+0.040$} & \revr{$+0.017$ to $+0.036$} \\
    \rev{IGP} Vendi, unconstrained & --- & \revr{$-0.040$ $[-0.092,+0.011]$} & \revr{$-0.028$ $[-0.067,+0.012]$} & \revr{$-0.034$ $[-0.081,+0.014]$} \\
    Semantic Vendi, unconstrained & --- & \revr{$+0.016$ $[-0.045,+0.078]$} & \revr{$+0.016$ $[-0.045,+0.078]$} & \revr{$+0.016$ $[-0.045,+0.078]$} \\
    \bottomrule
  \end{tabular}}
\end{table}

\FloatBarrier

\begin{figure}[p]
  \centering
  \includegraphics[width=0.85\linewidth]{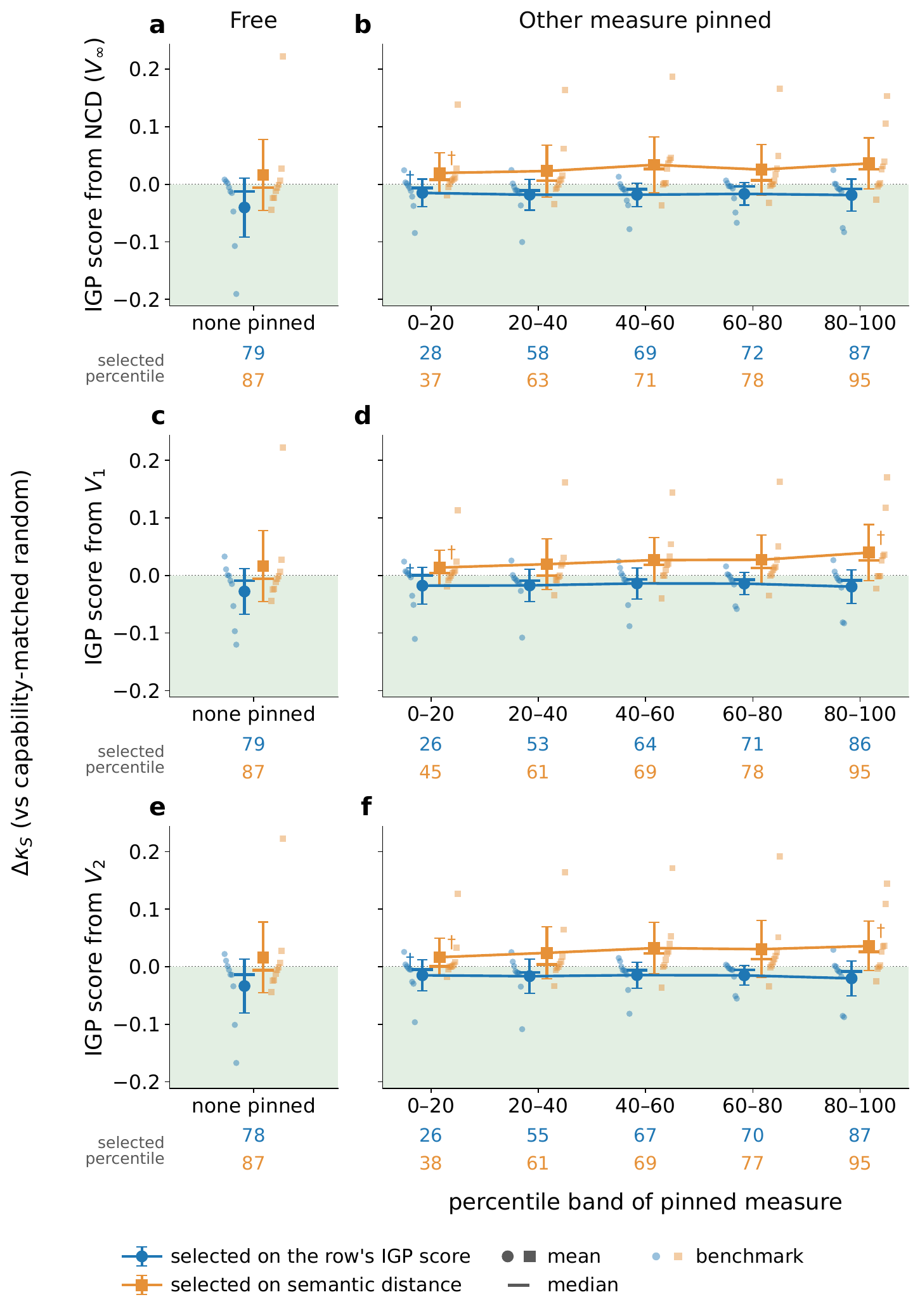}
  \caption{\textbf{Vendi selection free and pinned to a percentile band of the other \rev{measure}, for each \rev{version of the IGP score}.} As Figure~\ref{fig:ensemble}, whose panels are (a, b) here. \rev{Rows are the population IGP score computed from NCD ($V_\infty$) in (a, b), from $V_1$ in (c, d) and from $V_2$ in (e, f).} Large markers are the difference in $\kappa_S$ (Equation~\ref{eq:cfr}) between the top 200 five-model ensembles by Vendi score of one measure and capability-matched random ensembles\revr{, after the capability adjustment of Equation~\ref{eq:ensemble-delta}}, averaged over nine source benchmarks; small points are the individual benchmarks. Left column: free selection. Right column: the other measure's mean \rev{pairwise value} pinned to each percentile band. Bars are 95\% $t$-intervals and horizontal marks are medians; the green region below zero marks lower $\kappa_S$ than capability-matched random. Controls match mean accuracy and accuracy gap, and in the right column lie in the same band within 0.1 SD of the pinned \rev{value}. Coloured numerals are the median percentile of the selected ensembles on the selected measure. Daggers mark bands in which at least one selection had fewer than 50 controls.}
  \label{fig:supp-ensemble-vendi}
\end{figure}

\begin{figure}[p]
  \centering
  \includegraphics[width=0.85\linewidth]{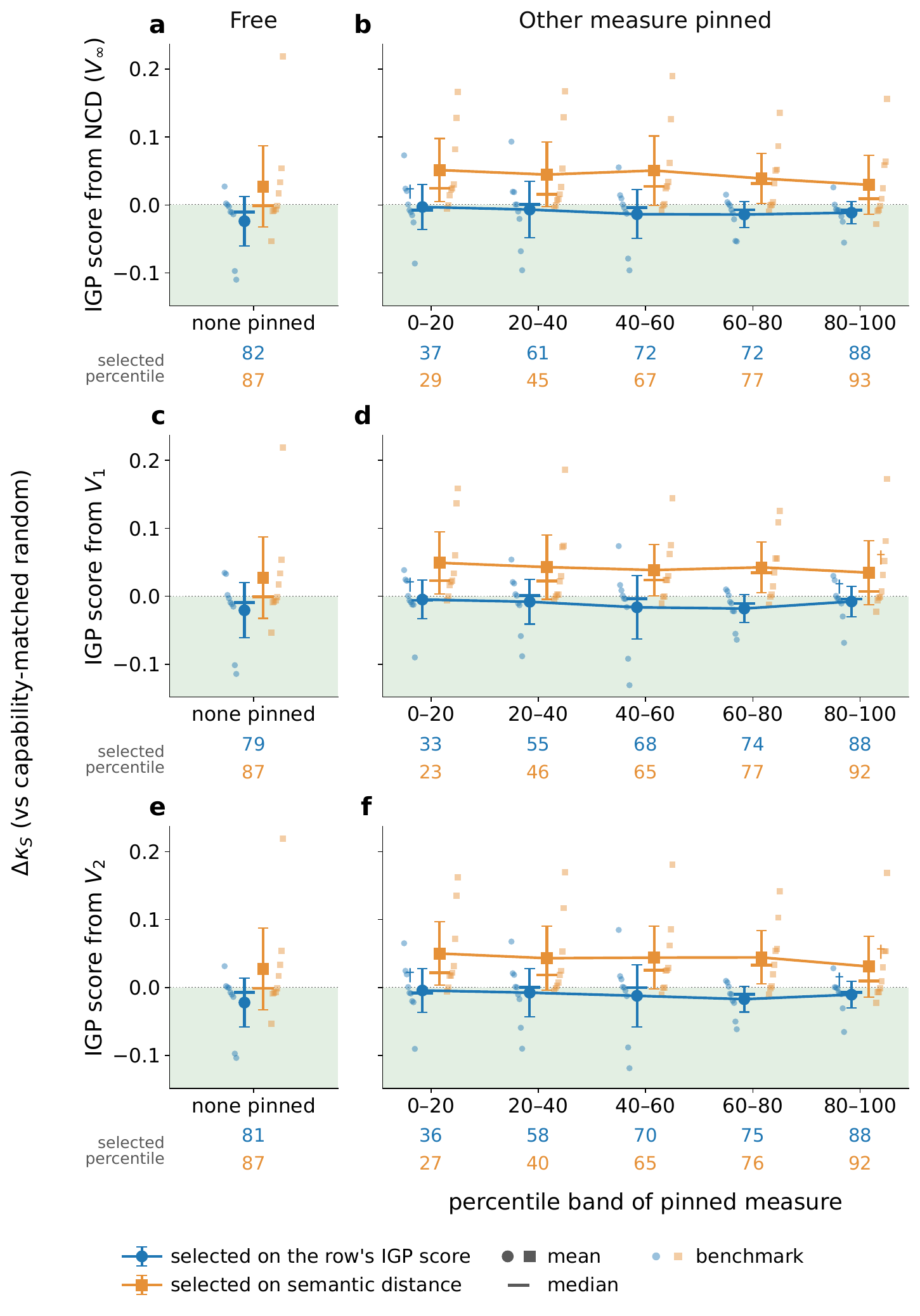}
  \caption{\textbf{Log-determinant selection free and pinned to a percentile band of the other \rev{measure}, for each \rev{version of the IGP score}.} As Figure~\ref{fig:ensemble}, with ensembles selected by the log-determinant of the \rev{ensemble} kernel. \rev{Rows are the population IGP score computed from NCD ($V_\infty$) in (a, b), from $V_1$ in (c, d) and from $V_2$ in (e, f).} Large markers are the difference in $\kappa_S$ (Equation~\ref{eq:cfr}) between the top 200 five-model ensembles by log-determinant of one measure and capability-matched random ensembles\revr{, after the capability adjustment of Equation~\ref{eq:ensemble-delta}}, averaged over nine source benchmarks; small points are the individual benchmarks. Left column: free selection. Right column: the other measure's mean \rev{pairwise value} pinned to each percentile band. Bars are 95\% $t$-intervals and horizontal marks are medians; the green region below zero marks lower $\kappa_S$ than capability-matched random. Controls match mean accuracy and accuracy gap, and in the right column lie in the same band within 0.1 SD of the pinned \rev{value}. Coloured numerals are the median percentile of the selected ensembles on the selected measure. Daggers mark bands in which at least one selection had fewer than 50 controls.}
  \label{fig:supp-ensemble-logdet}
\end{figure}

\begin{figure}[p]
  \centering
  \includegraphics[width=0.85\linewidth]{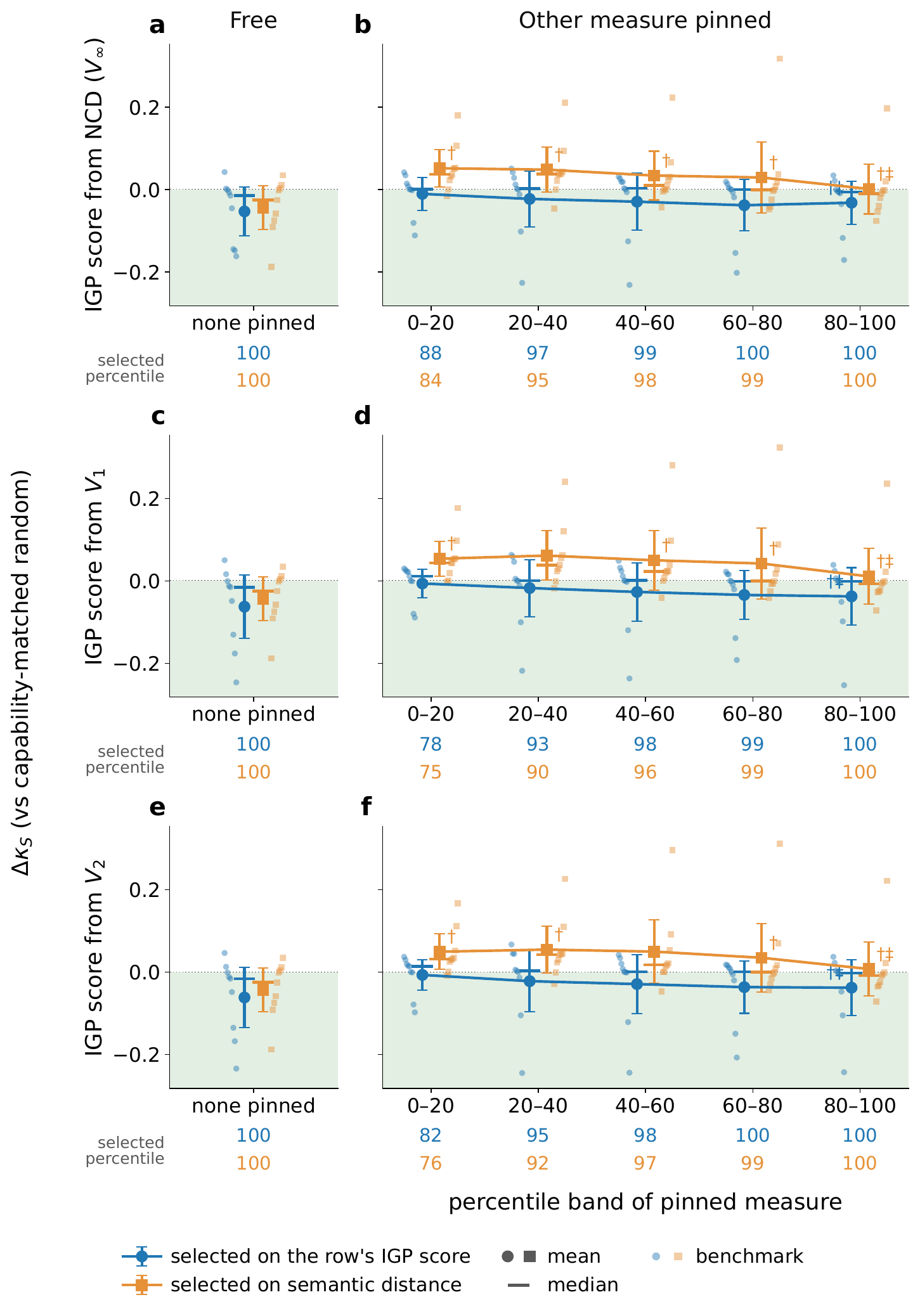}
  \caption{\textbf{Mean-\rev{value} selection free and pinned to a percentile band of the other \rev{measure}, for each \rev{version of the IGP score}.} As Figure~\ref{fig:ensemble}, with ensembles selected by mean pairwise \rev{value}. \rev{Rows are the population IGP score computed from NCD ($V_\infty$) in (a, b), from $V_1$ in (c, d) and from $V_2$ in (e, f).} Large markers are the difference in $\kappa_S$ (Equation~\ref{eq:cfr}) between the top 200 five-model ensembles by mean pairwise \rev{value} of one measure and capability-matched random ensembles\revr{, after the capability adjustment of Equation~\ref{eq:ensemble-delta}}, averaged over nine source benchmarks; small points are the individual benchmarks. Left column: free selection. Right column: the other measure's mean \rev{pairwise value} pinned to each percentile band. Bars are 95\% $t$-intervals and horizontal marks are medians; the green region below zero marks lower $\kappa_S$ than capability-matched random. Controls match mean accuracy and accuracy gap, and in the right column lie in the same band within 0.1 SD of the pinned \rev{value}. Coloured numerals are the median percentile of the selected ensembles on the selected measure. Daggers mark bands in which at least one selection had fewer than 50 controls; double daggers mark bands in which at least one selection's only control was itself.}
  \label{fig:supp-ensemble-mean}
\end{figure}

\subsection{\rev{Pooled across-band estimates}}
\label{app:ensemble-pooled}

\begin{revblock}
Figure~\ref{fig:ensemble}b gives five band estimates. To give one estimate of selection with semantic distance held fixed, we average the five band differences within each benchmark and then summarise across benchmarks (Table~\ref{tab:rev-pooled}). For CFR $\kappa$ the pooled estimate is negative under all three embeddings. The rate at which all five members are wrong is lower in the selected ensembles under every embedding\revr{, by 0.21 to 0.28 per 100 questions}. \revr{Relative to each benchmark's control level, both rates fall by 4\% to 5\%, and without AIME, on which the rate of a shared wrong answer is close to zero, that rate falls by 10\% to 12\%. Under free selection, all five members are wrong on 0.39 fewer questions per 100, $[-0.72,-0.05]$.} For the reverse selection, on semantic distance with the population IGP score held within a band, the pooled CFR $\kappa$ estimate is positive \revr{under every embedding}. Appendix~\ref{app:vote-bands} gives the pooled estimates for the vote outcomes.
\end{revblock}

\begin{table}[htbp]
  \revcolor
  \caption{\textbf{Pooled across-band estimates in the design of Figure~\ref{fig:ensemble}b.} Selection on the population IGP score with semantic distance held within a band, selected minus capability-matched ensembles\revr{ after the capability adjustment of Equation~\ref{eq:ensemble-delta}}. Mean of the five band differences within each benchmark, then the mean across nine benchmarks with its 95\% interval and the number of benchmarks with a negative estimate. The last row is the reverse selection, on semantic distance with the population IGP score held within a band. \rev{The rows in \% change divide each benchmark's difference by the mean of its capability-matched ensembles. AIME is unstable on that scale. All five give the same wrong answer on about 0.05 of every 100 AIME questions, so a change of $+0.02$ reads as \revr{$+46\%$}.}}
  \label{tab:rev-pooled}
  \centering
  \resizebox{\linewidth}{!}{%
  \begin{tabular}{llll}
    \toprule
    Outcome & MiniLM & TE3-small & Qwen3 \\
    \midrule
    CFR $\kappa$ & \revr{$-0.017$ $[-0.039,+0.004]$ 7/9} & \revr{$-0.008$ $[-0.026,+0.009]$ 5/9} & \revr{$-0.009$ $[-0.029,+0.011]$ 5/9} \\
    All five give the same wrong answer, per 100 & \revr{$-0.51$ $[-1.30,+0.29]$ 7/9} & \revr{$-0.46$ $[-1.16,+0.24]$ 7/9} & \revr{$-0.44$ $[-1.12,+0.25]$ 7/9} \\
    All five wrong, per 100 & \revr{$-0.28$ $[-0.58,+0.02]$ 7/9} & \revr{$-0.23$ $[-0.53,+0.08]$ 6/9} & \revr{$-0.21$ $[-0.51,+0.09]$ 6/9} \\
    \rev{All five give the same wrong answer, \% change} & \revr{$-5.4$ $[-22.6,+11.9]$ 7/9} & \revr{$-5.1$ $[-20.7,+10.6]$ 7/9} & \revr{$-4.3$ $[-19.3,+10.8]$ 7/9} \\
    \rev{\quad without AIME} & \revr{$-11.7$ $[-22.3,-1.1]$ 7/8} & \revr{$-10.7$ $[-20.7,-0.8]$ 7/8} & \revr{$-9.7$ $[-19.3,-0.2]$ 7/8} \\
    \rev{All five wrong, \% change} & \revr{$-5.0$ $[-10.8,+0.9]$ 7/9} & \revr{$-4.6$ $[-10.5,+1.3]$ 6/9} & \revr{$-4.2$ $[-9.5,+1.1]$ 6/9} \\
    \midrule
    CFR $\kappa$, reverse selection & \revr{$+0.028$ $[-0.014,+0.069]$ 3/9} & \revr{$+0.032$ $[+0.0003,+0.064]$ 2/9} & \revr{$+0.017$ $[-0.003,+0.037]$ 2/9} \\
    \bottomrule
  \end{tabular}}
\end{table}

\subsection{\rev{Conditioning on at least one wrong member}}
\label{app:ensemble-pair-rule}

\begin{revblock}
CFR conditions on every member being wrong because that is the worst case for an ensemble. If the members then give different wrong answers, their disagreement shows the failure. If they all give the same one, the failure goes unnoticed. The pairwise outcome asks a different question, whether a second model shares the first one's error, which needs only one of the two to be wrong. To check that the ensemble result does not depend on this choice, we also score ensembles under the pairwise rule, with at least one member wrong in place of all members wrong. On question $q$ let $R_q$ count the epoch tuples in which every member is right, so that $W_q=T_q-R_q$ counts the tuples with at least one member wrong. The pair-rule outcome keeps the numerator and the chance count of $\kappa_S$ (Equation~\ref{eq:cfr-appendix}) and widens only the denominator:
\begin{equation}
O^{\mathrm{pair}}_S=\frac{\sum_{q\in\mathcal{Q}_S}C_q}{\sum_{q\in\mathcal{Q}_S}W_q},
\qquad
X^{\mathrm{pair}}_S=\frac{\sum_{q\in\mathcal{Q}_S}A_q\,b_q}{\sum_{q\in\mathcal{Q}_S}W_q},
\qquad
\kappa^{\mathrm{pair}}_S=\frac{O^{\mathrm{pair}}_S-X^{\mathrm{pair}}_S}{1-X^{\mathrm{pair}}_S}.
\label{eq:pair-rule}
\end{equation}
All three sums run over the questions $\mathcal{Q}_S$ on which $\kappa_S$ is estimable. $O^{\mathrm{pair}}_S$ and $X^{\mathrm{pair}}_S$ are $\CFR_S$ and $\CFR^{\mathrm{exp}}_S$ multiplied by $f_S=\sum_q A_q/\sum_q W_q$, the share of tuples with a wrong member in which every member is wrong. Hence $\kappa^{\mathrm{pair}}_S=f_S(\CFR_S-\CFR^{\mathrm{exp}}_S)/(1-f_S\,\CFR^{\mathrm{exp}}_S)$, which lies closer to zero than $\kappa_S$ for every ensemble. The chance count still comes from the tuples in which every member is wrong, so the pair rule does not credit an ensemble for failing together less often. That effect is the rate at which all five are wrong in Table~\ref{tab:rev-pooled}. Because the two rules put $\kappa$ on different scales, their values cannot be compared in size. Under both rules the 200 ensembles with the highest Vendi score on the population IGP score give the same wrong answer less often than capability-matched ensembles, at \revr{$-0.014$ $[-0.029,+0.002]$} under the pair rule and \revr{$-0.040$ $[-0.092,+0.011]$} under the ensemble rule. With semantic distance held within a band, the sign agrees in \revr{13 of the} 15 band cells (Table~\ref{tab:rev-pair-rule})\revr{. The two exceptions are ensemble-rule means of $+0.002$ in the 20--40 band under TE3-small and Qwen3.}
\end{revblock}

\begin{table}[!htbp]
  \revcolor
  \caption{\textbf{Ensembles under the pair rule and under the ensemble rule.} Top 200 ensembles by Vendi score on the population IGP score minus capability-matched ensembles\revr{ after the capability adjustment of Equation~\ref{eq:ensemble-delta}}, with no band and with semantic distance held within each percentile band. Mean over nine benchmarks with its 95\% interval. The selection with no band does not use the embedding.}
  \label{tab:rev-pair-rule}
  \centering
  \resizebox{\linewidth}{!}{%
  \begin{tabular}{llll}
    \toprule
    Band of semantic distance & MiniLM & TE3-small & Qwen3 \\
    \midrule
    \multicolumn{4}{l}{\emph{Pair rule, at least one member wrong}} \\
    No band & \revr{$-0.014$ $[-0.029,+0.002]$} & \revr{$-0.014$ $[-0.029,+0.002]$} & \revr{$-0.014$ $[-0.029,+0.002]$} \\
    0--20 & \revr{$-0.006$ $[-0.019,+0.006]$} & \revr{$-0.004$ $[-0.019,+0.010]$} & \revr{$-0.005$ $[-0.019,+0.009]$} \\
    20--40 & \revr{$-0.010$ $[-0.025,+0.004]$} & \revr{$-0.006$ $[-0.019,+0.008]$} & \revr{$-0.004$ $[-0.016,+0.008]$} \\
    40--60 & \revr{$-0.009$ $[-0.020,+0.002]$} & \revr{$-0.006$ $[-0.015,+0.004]$} & \revr{$-0.005$ $[-0.016,+0.005]$} \\
    60--80 & \revr{$-0.008$ $[-0.017,+0.0001]$} & \revr{$-0.006$ $[-0.013,+0.002]$} & \revr{$-0.006$ $[-0.016,+0.003]$} \\
    80--100 & \revr{$-0.010$ $[-0.023,+0.002]$} & \revr{$-0.008$ $[-0.017,-0.0003]$} & \revr{$-0.008$ $[-0.017,+0.001]$} \\
    \midrule
    \multicolumn{4}{l}{\emph{Ensemble rule, all members wrong (CFR $\kappa$)}} \\
    No band & \revr{$-0.040$ $[-0.092,+0.011]$} & \revr{$-0.040$ $[-0.092,+0.011]$} & \revr{$-0.040$ $[-0.092,+0.011]$} \\
    0--20 & \revr{$-0.015$ $[-0.039,+0.009]$} & \revr{$-0.007$ $[-0.038,+0.024]$} & \revr{$-0.006$ $[-0.034,+0.022]$} \\
    20--40 & \revr{$-0.018$ $[-0.045,+0.008]$} & \revr{$+0.002$ $[-0.030,+0.033]$} & \revr{$+0.002$ $[-0.023,+0.027]$} \\
    40--60 & \revr{$-0.018$ $[-0.039,+0.002]$} & \revr{$-0.008$ $[-0.025,+0.008]$} & \revr{$-0.007$ $[-0.030,+0.015]$} \\
    60--80 & \revr{$-0.017$ $[-0.036,+0.003]$} & \revr{$-0.007$ $[-0.023,+0.008]$} & \revr{$-0.015$ $[-0.041,+0.011]$} \\
    80--100 & \revr{$-0.019$ $[-0.047,+0.009]$} & \revr{$-0.021$ $[-0.058,+0.016]$} & \revr{$-0.018$ $[-0.054,+0.018]$} \\
    \bottomrule
  \end{tabular}}
\end{table}

\end{addblock}

\FloatBarrier

\section{Statistical Methods}
\label{app:statistics}

\subsection{Estimand and partial rank correlation}
\label{app:estimand}

For paired vectors $x$ and $y$, Pearson's correlation is the covariance standardised by their sample standard deviations,
\begin{equation}
r(x,y)=\frac{\sum_i(x_i-\bar{x})(y_i-\bar{y})}
{\sqrt{\sum_i(x_i-\bar{x})^2\sum_i(y_i-\bar{y})^2}}.
\label{eq:pearson}
\end{equation}
It measures linear association on the observed scale. Spearman's $\rho_s$ applies the same calculation to the componentwise midranks, $\rho_s(x,y)=r\{R(x),R(y)\}$, and therefore measures monotone association while being invariant to strictly increasing transformations \citep{spearman_proof_1904}. We use Pearson correlations for explicitly linear-scale diagnostics and Spearman correlations when comparing pair or model orderings.

The primary estimand is the partial Spearman association between a pair's \rev{population IGP score} and chance-corrected CWA, conditional on semantic distance, capability level, and capability gap. Let $\widetilde{x}=R(x)$ and $\widetilde{y}=R(y)$, and let $\widetilde{Z}$ contain an intercept and the midranks of all controls. With $P_Z=\widetilde{Z}(\widetilde{Z}^{\mathsf T}\widetilde{Z})^{-1}\widetilde{Z}^{\mathsf T}$, we compute
\begin{equation}
\rho_s(x,y\mid Z)
=r\left\{(I-P_Z)\widetilde{x},(I-P_Z)\widetilde{y}\right\}.
\label{eq:partial-spearman}
\end{equation}
\rev{Thus, we correlate the parts of the ranked predictor and of the ranked outcome that the same ranked controls do not explain.} Simultaneous control matters. Holding capability gap alone gives \rev{$-0.053$}, whereas holding level and gap gives $-0.148$. Rank partialling removes dependence linear in ranks and can absorb monotone nonlinear confounding, but can leave \rev{part of any confounding that is additive in raw values}.

Capability level is $(a_i+a_j)/2$ and gap is $|a_i-a_j|$. The gap enters as a pre-existing pair attribute. Because capability is estimated from the same benchmark correctness matrices, both terms function as analytic controls for \rev{the performance of a pair}. \rev{With the capability controls entered in linear, quadratic and rank form, the partial rank correlation of Equation~\ref{eq:partial-spearman} is $-0.077$ $[-0.188,+0.034]$, $-0.140$ $[-0.246,-0.035]$ and $-0.148$ $[-0.245,-0.050]$, with 8, 9 and 9 of the 10 benchmarks negative. The linear form uses the raw values of level and gap, the quadratic form adds their squares and their cross term, and the rank form uses their midranks.} \rev{The quadratic and rank forms give nearly the same estimate, which} supports the rank specification used in the primary analysis. \rev{The linear form gives the weakest estimate because the relation between capability and each ranked variable is curved. After the linear adjustment, the squares and the product of level and gap still explain about 30\% of the remaining variation in the ranked $\kappa$ and part of the remaining variation in the ranked population IGP score. On seven of the ten benchmarks these two explained parts are positively correlated, which makes the linear estimate less negative.}

\rev{The association is negative under all three forms of the capability control, and its interval includes zero under the linear form. The question requires the controls. Capability and semantic distance relate both to the measure and to shared wrong answers, so an estimate without them would mix these confounds with the source of diversity that we want to isolate. With no control at all, the Spearman correlation between the population IGP score and $\kappa$ is $-0.269$ $[-0.318,-0.220]$, negative on all ten benchmarks. The same holds for the other three measures: the population IGP distance gives $-0.388$ $[-0.426,-0.349]$, and against member-adjusted CWA the pair-specific IGP score and distance give $-0.569$ $[-0.688,-0.450]$ and $-0.650$ $[-0.778,-0.521]$, each negative on all ten benchmarks. The pair-specific measures reduce the dependence on the form of the capability controls. They remove each model's additive effect from the measure, the outcome and the controls, so capability level, which is a sum of two single-model terms, drops out in every form. The capability gap and the accuracy product remain as controls (Appendix~\ref{app:pair-specific}).}

\subsection{Dyadic dependence and model-node resampling}
\label{app:node-bootstrap}

\rev{Pairs that share a model are dependent \citep{aronow_cluster_2015}.} To make the dependence explicit, write a generic pair quantity as $x_{ij}=\mu+\rev{g_i+g_j}+e_{ij}$, where the independent node effects have variance $\operatorname{Var}(\rev{g_i})=\rev{\sigma_g^2}$, the independent dyad residuals have variance $\operatorname{Var}(e_{ij})=\sigma_e^2$, and $\bar{x}$ averages all unordered pairs. Then
\begin{equation}
\operatorname{Var}(\bar{x})=\frac{4\rev{\sigma_g^2}}{M}+\frac{\sigma_e^2}{\binom{M}{2}}.
\end{equation}
The model component decreases with the number of nodes $M$, not with the number of dyads $\binom{M}{2}$, because each \rev{$g_i$} is shared by the $M-1$ pairs incident to model $i$. This component is material in the observed data. An incidence model attributes 58.6\% of \rev{the population IGP score}'s rank variance and 61--85\% of outcome rank variance to model-level structure.

The bootstrap preserves this incidence structure directly. For each replicate, we draw $M$ model slots with replacement, construct every unordered pair of distinct slots, map those induced dyads to the observed pair rows, and recompute the complete rank-partial statistic, including all controls. Selecting a model more than once repeats all of its incident dyads; self-pairs and non-estimable dyads are omitted. The interval is given by the 2.5th and 97.5th percentiles of the resulting statistic \rev{over 2,000 replicates}. A delete-one-model jackknife closely agrees with these intervals on every benchmark (mean width ratio 0.95, range 0.89--1.02). In 300 simulations in which the true association was zero, calibrated to the measured model-level variance shares, the node-bootstrap interval excluded zero in 3.0\% of simulations at a nominal 5\% level. The intervals remain conditional on prompts, generations, benchmark questions, evaluation epochs, and the composition of the evaluated model population.

\subsection{Paired comparison of conditional associations}
\label{app:paired-difference}

The two cross-controlled partial correlations are estimated on the same model pairs and outcome within each benchmark. Let $Z_b$ contain capability level and gap for benchmark $b$, and define
\begin{equation}
\begin{aligned}
\rho^{\IGP}_b
&=\rho_s(\spop,\kappa_b\mid d^{\mathrm{sem}},Z_b),\\
\rho^{\mathrm{sem}}_b
&=\rho_s(d^{\mathrm{sem}},\kappa_b\mid\spop,Z_b),
\qquad
\Delta_b=\rho^{\IGP}_b-\rho^{\mathrm{sem}}_b.
\end{aligned}
\label{eq:paired-partial-difference}
\end{equation}
Both coefficients are partial correlations on ranked variables and therefore lie in $[-1,1]$. Their difference $\Delta_b$ is expressed in correlation units but is not itself a correlation coefficient; its theoretical range is $[-2,2]$. \rev{Negative values mean that the population IGP score has the more negative association with correlated failure.}

\begin{figure}[!t]
  \centering
  \includegraphics[width=\linewidth]{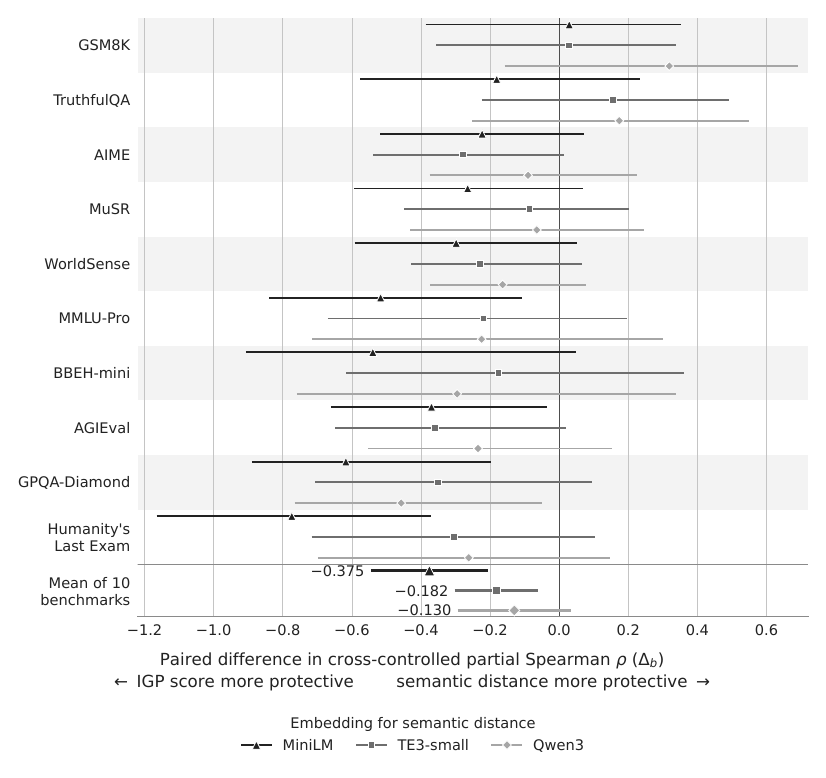}
  \caption{\textbf{Paired comparison of the cross-controlled partial correlations\rev{, under three embeddings}.} Each benchmark row shows $\Delta_b=\rho^{\IGP}_b-\rho^{\mathrm{sem}}_b$, a difference in correlation units with theoretical range $[-2,2]$. \rev{The marker gives the embedding used for semantic distance (Appendix~\ref{app:embeddings}).} \rev{Markers} are point estimates and horizontal bars are 95\% model-node-bootstrap intervals obtained by differencing the two correlations within each resample. \rev{The last row is the unweighted mean over the ten benchmarks with its 95\% $t$-interval, and the number beside it is that mean. Rows are in the order of Figure~\ref{fig:dissociation}.} Negative values \rev{mean that the population IGP score} has the more negative conditional association with correlated failure.}
  \label{fig:supp-paired-difference}
\end{figure}

The two coefficients are dependent because they share the same models, dyads, outcome, and control variables. We therefore compute both coefficients within each model-node bootstrap replicate and difference them within that replicate. The percentile interval for each benchmark consequently retains the sampling covariance between the two estimates. Comparing the overlap of their marginal intervals would not test $\Delta_b=0$. Across benchmarks, we report the unweighted mean $\bar\Delta=10^{-1}\sum_b\Delta_b$ and its $t$-interval over the ten benchmark estimates. Figure~\ref{fig:supp-paired-difference} shows $\bar\Delta=-0.375$ with 95\% interval $[-0.545,-0.206]$ \rev{under MiniLM, $-0.182$ $[-0.302,-0.062]$ under TE3-small and $-0.130$ $[-0.294,+0.034]$ under Qwen3, the embeddings of Appendix~\ref{app:embeddings}. Under the stronger embeddings both estimates move towards zero (Table~\ref{tab:rev-semantic}), so the difference is smaller, and under Qwen3 its interval includes zero}.

\subsection{Repeated-measures ranks and multiplicity}
\label{app:rank-and-multiplicity}

The within-model analysis compares $K=38$ models repeatedly across $B=100$ prompt blocks. Within each prompt, the models are ranked on the response-variability measure. Let $R_j$ be the rank sum for model $j$ across prompts and $\bar{R}=B(K+1)/2$. In the absence of ties, Kendall's coefficient of concordance is
\begin{equation}
W=\frac{12\sum_{j=1}^{K}(R_j-\bar{R})^2}{B^2(K^3-K)},
\qquad 0\leq W\leq1,
\label{eq:kendall-w}
\end{equation}
with the standard tie correction used when ranks coincide \citep{kendall_problem_1939}. Here, $W=0$ indicates no stable model ordering across prompts and $W=1$ indicates complete agreement among the prompt-specific rankings. It is an effect-size measure, not a test of any particular model pair.

The Friedman test uses the same blocked ranks to test the omnibus hypothesis that the models have no systematic differences in rank location across prompts \citep{friedman_use_1937}. Its tie-corrected statistic $Q$ satisfies $Q=B(K-1)W$ and is compared with a $\chi^2_{K-1}$ reference distribution. Rejection establishes that at least one model differs in its repeated rank pattern; it neither identifies which models differ nor supplies pairwise comparisons. The $W$ values in Appendix~\ref{app:within-model} describe the magnitude of the stable ordering, while the associated Friedman tests assess whether that ordering is distinguishable from the equal-rank hypothesis.

The compressor analysis uses a different paired procedure. For each of the $U=72$ matched length--replicate units, let $d_u$ be PPMd's shared-context effect minus that of one alternative compressor. The one-sided sign-flip test compares the observed $\bar d$ with 20,000 values $\bar d_b^*=U^{-1}\sum_u s_{bu}d_u$, where the signs $s_{bu}\in\{-1,+1\}$ are sampled independently with equal probability. Its Monte Carlo value is
\begin{equation}
p=\frac{1+\#\{b:\bar d_b^{*}\geq\bar d\}}{20{,}001},
\label{eq:sign-flip}
\end{equation}
so a reported value cannot be zero. The \rev{15} PPMd-versus-compressor tests \rev{(byte-level CTW at three depths, binary-decomposition CTW at nine settings, bzip2, LZMA and gzip)} form one comparison family. If their ordered unadjusted values are $p_{(1)}\leq\cdots\leq p_{(m)}$, with \rev{$m=15$}, the Holm-adjusted values are
\begin{equation}
\widetilde p_{(i)}=
\min\left\{1,\max_{1\leq j\leq i}(m-j+1)p_{(j)}\right\}.
\label{eq:holm}
\end{equation}
This step-down adjustment controls the family-wise probability of at least one false rejection while retaining more power than applying the same Bonferroni threshold to every test \citep{holm_simple_1979}. The value \rev{$7.5\times10^{-4}$} reported in Appendix~\ref{app:compressor-choice} is the adjusted value for \rev{each contrast whose unadjusted value is the smallest attainable, $1/20{,}001$}.

\rev{The vote outcomes use the same adjustment. Appendix~\ref{app:vote-inference} defines their two families of tests.}

\section{Extended Related Work}
\label{app:extended-related-work}

\addtext{Resilience research holds that redundancy protects a system only when components performing the same function respond differently to perturbation \citep{elmqvist_response_2003,holling_resilience_1973,yachi_biodiversity_1999}. In model populations, organisational labels do not supply that independence. Models from different providers select the same wrong answers above pair-specific chance \citep{goel_great_2025,kim_correlated_2025}, multi-agent safety work catalogues the population-level failure modes this creates \citep{franklin_ai_2026,hammond_multi-agent_2025,tomasev_distributional_2026}, and the Hugging Face incident supplies a deployed instance \citep{greenblatt_hugging_2026}. The Artificial Hivemind measures cross-model similarity through embeddings and supplies our semantic baseline, but shared meaning is not \rev{a shared generative process} \citep{jiang_artificial_2025}. \rev{Other measures also compare outputs: PhyloLM computes a genetic distance over token completions \citep{yax_phylolm_2024}, while DFPE and LLM-TOPLA select ensembles by response fingerprints and by error diversity on the target task \citep{cohen_dfpe_2025,tekin_llm-topla_2024}. Our measure needs no labelled data or tuning on the target task, as it is computed from raw responses to unrelated prompts. Appendix~\ref{app:phylolm} compares our measure with PhyloLM.} AIT offers a feature-free alternative in which NCD approximates \rev{the information distance} with compressed lengths \citep{cilibrasi_clustering_2005,vitanyi_normalized_2009}. Whether that relation predicts correlated failure beyond semantic distance and capability is the question we test.}

\subsection{Diversity and resilience in collective systems}

Diversity can contribute to collective performance through several distinct mechanisms, including the buffering or insurance effects produced when components respond differently to fluctuating conditions \citep{holling_resilience_1973,yachi_biodiversity_1999}. Functional diversity describes variation in the functions performed by system components. Response diversity describes variation in how components contributing to the same function respond to perturbation \citep{elmqvist_response_2003,ross_how_2023}. The latter is particularly relevant to redundancy. Components are not interchangeable safeguards if they respond identically under the conditions that cause failure. Work on resilience and critical transitions also shows that the organisation of heterogeneity and coupling can determine whether local perturbations remain local or become system-wide \citep{scheffer_anticipating_2012}. Diversity is therefore multidimensional. Its effect depends on the system function and the disturbance under study.

Collective-adaptation research reaches a related conclusion from a problem-solving perspective. Heterogeneous information and strategies can expand the set of solutions explored by a group. Rapid information sharing can improve diffusion, but it can also accelerate convergence on a common and potentially suboptimal solution \citep{galesic_beyond_2023,tieman_landscape_2025}. The arrangement of diversity matters as much as its amount because coupling determines whether differences remain available to the collective when conditions change. These findings motivate population-level analysis of AI systems, but they do not specify which representation of model diversity is appropriate. Our use of resilience theory is methodological rather than analogical. It determines the outcome-relative definition of diversity and the criterion used to validate its measure.

\subsection{Diversity and correlated failure in AI systems}

Multi-agent safety work has begun to identify failure modes that are not reducible to the behaviour of an isolated model. Homogeneous agents can respond synchronously to common information, propagate shared errors through a network, or provide ineffective mutual oversight \citep{hammond_multi-agent_2025,tomasev_distributional_2026,franklin_ai_2026}. These risks extend concerns about algorithmic monoculture and are especially important for defence-in-depth designs whose reliability assumes that separate components fail under different conditions \citep{kleinberg_algorithmic_2021}.

Empirical evidence suggests that organisational and architectural labels are weak proxies for this independence. Language models from different providers select identical wrong answers at rates above pair-specific chance expectations \citep{goel_great_2025,kim_correlated_2025}. Provider count, parameter count, and model-family count may therefore overstate the effective diversity of a multi-model system. A useful audit must estimate behavioural relationships among the models themselves and validate those relationships against a relevant failure outcome.

The Artificial Hivemind measures one such relationship using embedding similarity among open-ended responses \citep{jiang_artificial_2025}. Its finding of high semantic similarity across the model population motivates our study and supplies the semantic baseline. Our objective differs in the property being estimated. Embedding similarity asks whether response meanings are similar. Inferred generative-process similarity asks whether observable sequences exhibit shared generative organisation. We evaluate whether this distinction explains variation in correlated failure that semantic similarity does not.

\begin{revblock}
Three further methods compare models through their outputs. PhyloLM treats a prompt as a gene and the first four characters of a continuation as an allele, and takes a genetic distance between two models' allele distributions \citep{yax_phylolm_2024}. DFPE and LLM-TOPLA select ensembles by response fingerprints and by error diversity on the target task \citep{cohen_dfpe_2025,tekin_llm-topla_2024}. Both select on embedding diversity and use target-task outputs and labels. Our measure differs in two ways. First, it uses responses to open-ended prompts only. No benchmark question or label enters it, and we test it on disjoint benchmarks. Second, it targets the part of the compression distance that response length and byte composition do not predict, where the other methods compare token frequencies or embedded meaning. Selection on embedding diversity corresponds to selection on semantic distance in Figure~\ref{fig:ensemble}, which does not reduce convergent failure once the population IGP score is held within a band. Appendix~\ref{app:phylolm} compares the measure with PhyloLM on our fleet.
\end{revblock}

\subsection{AIT and generative structure}

AIT characterises an object by the length of the shortest program that generates it. Kolmogorov complexity is uncomputable, but it motivates universal similarity measures based on shared algorithmic information. Normalised Information Distance compares the information needed to describe either object given the other. NCD approximates this relation by replacing program length with compressed length \citep{cilibrasi_clustering_2005,vitanyi_normalized_2009}. NCD has the practical advantage of operating directly on discrete sequences and requiring neither task-specific features nor a learned representation.

AIT approaches have also been used to distinguish generative regularity from apparent statistical randomness \citep{zenil_low-algorithmic-complexity_2017,zenil_algorithmic_2023}. This perspective clarifies why generative and statistical descriptions need not coincide. An object may appear complex under one selected representation while retaining simple generative organisation. We make a narrower empirical claim than recovering a model's generating program. A compressor does not identify a latent algorithm, and finite-sample NCD is compressor-dependent. We use NCD as a comparative statistic of shared sequential structure and test its validity through prediction of disjoint correlated-failure outcomes.

\section{\rev{IGP Scores and Distances}}
\label{app:igp-measures}

\begin{revblock}
This appendix defines the four IGP measures and reports the main analysis for each (Table~\ref{tab:rev-four-measures}). The population IGP score $\spop$ is the averaged residual of Equation~\ref{eq:resid}. The pair-specific IGP score $\spair$ removes each model's own effect before the residualisation. The IGP distance turns a signed score into a distance through the ensemble kernel of Appendix~\ref{app:ensemble-kernel}, which gives the population IGP distance $\dpop$ and the pair-specific IGP distance $\dpair$. The two scores are signed and are not distances. The two distances are metrics. The population IGP score is the most conservative of the four, because it mixes properties of single models with properties of particular pairs, which dilutes it. The main figures use it.
\end{revblock}

\begin{table}[htbp]
  \revcolor
  \caption{\textbf{The four IGP measures.}}
  \label{tab:rev-four-measures}
  \centering
  \begin{tabular}{p{0.16\linewidth}p{0.38\linewidth}p{0.38\linewidth}}
    \toprule
     & Score (signed) & Distance (metric) \\
    \midrule
    Population & Population IGP score $\spop$: the residual of NCD on its permutation control, averaged over prompts (Equation~\ref{eq:resid}). & Population IGP distance $\dpop$: distance between two models' feature vectors in the ensemble kernel built on the population IGP score. \\
    Pair-specific & Pair-specific IGP score $\spair$: the residualisation applied after each model's additive effect is removed (Equation~\ref{eq:pair-specific}). & Pair-specific IGP distance $\dpair$: the same distance, with the kernel built on the pair-specific IGP score. \\
    \bottomrule
  \end{tabular}
\end{table}

\subsection{\rev{Model effects and the pair-specific IGP score}}
\label{app:pair-specific}

\begin{revblock}
The regression of Equation~\ref{eq:resid} is fitted across all 703 pairs. The population IGP score therefore mixes properties of single models, such as their response length, style and capability, with properties of particular pairs, and model-level structure accounts for 58.6\% of its rank variance (Appendix~\ref{app:node-bootstrap}). Its association with correlated failure then depends on how the capability controls separate the two.

For any pair quantity $x_{ij}$ we fit, by least squares over pairs,
\begin{equation}
x_{ij}=\mu+g_i+g_j+e_{ij},
\label{eq:model-effect}
\end{equation}
where the model effect $g_i$ is model $i$'s typical value with every other model and $e_{ij}$ is the pair-specific part. For NCD, $g_i$ is the model's typical distance to all others, which reflects its response length and style. For CWA it mostly tracks the model's accuracy. We call the removal of the fitted model effects the member adjustment. For a pair quantity, and for a quantity $x_S$ of an ensemble $S$ with effects fitted by least squares over the pool of ensembles, it gives
\begin{equation}
\widetilde{x}_{ij}=x_{ij}-\widehat{\mu}-\widehat{g}_i-\widehat{g}_j,
\qquad
\widetilde{x}_{S}=x_{S}-\widehat{\mu}-\sum_{m\in S}\widehat{g}_m.
\label{eq:member-adjustment}
\end{equation}
A positive $\widetilde{x}_S$ means that the ensemble exceeds what its members predict. The pair-specific IGP score applies the adjustment to the raw and to the permuted NCD within each prompt, regresses one remainder on the other through the origin, and averages the residual over the prompts on which the pair is present:
\begin{equation}
\widehat{\varepsilon}^{\,\mathrm{pair}}_{ijq}=e^{\mathrm{NCD}}_{ijq}-\widehat{\beta}_q\,e^{\mathrm{perm}}_{ijq},
\qquad
\widehat{\beta}_q=\frac{\sum_{i<j}e^{\mathrm{NCD}}_{ijq}e^{\mathrm{perm}}_{ijq}}{\sum_{i<j}\big(e^{\mathrm{perm}}_{ijq}\big)^2},
\qquad
\spair_{ij}=\frac{1}{Q}\sum_{q=1}^{Q}\widehat{\varepsilon}^{\,\mathrm{pair}}_{ijq}.
\label{eq:pair-specific}
\end{equation}
We test it against member-adjusted CWA, the pair-specific part $e_{ij}$ of CWA $\kappa$ fitted within each benchmark. We apply the same adjustment to the IGP measure, to semantic distance and to the capability gap. After the adjustment the capability level is zero and drops out. We add the product of the two models' accuracies, $a_ia_j$, as a control, which keeps the part of capability that depends on the combination, such as two weak models meeting. The analysis asks whether two models that differ more than their own general levels predict also share fewer wrong answers than those levels predict. No trait of a single model can produce such an association.
\end{revblock}

\subsection{\rev{The IGP distance}}
\label{app:igp-distance}

\begin{figure}[htbp]
  \centering
  \includegraphics[width=0.95\linewidth]{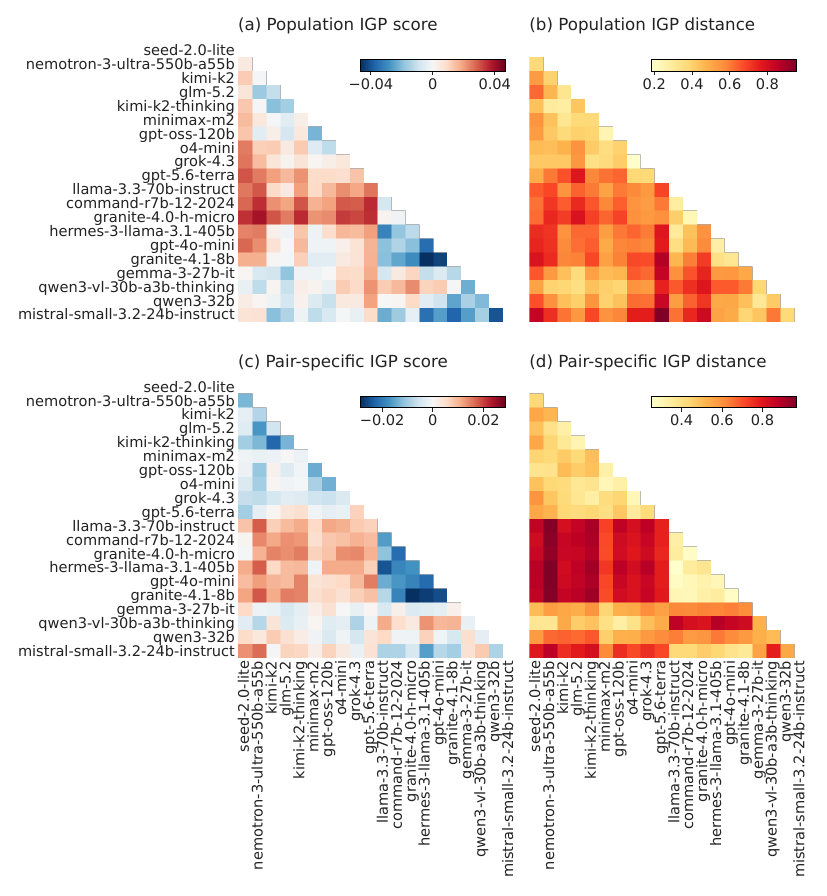}
  \caption{\revcolor{}\textbf{The four IGP measures as matrices.} The 20 models of Figure~\ref{fig:views}, in the same order. (a), (c) The signed scores on a diverging scale centred at zero; positive values are pairs farther apart than response length and byte composition predict. (b), (d) The distances on a sequential scale. Each panel has its own colour bar. Reported associations use all 703 pairs.}
  \label{fig:rev-four-measures}
\end{figure}

\begin{revblock}
The signed score is not a distance. The ensemble analysis maps it to a kernel (Equation~\ref{eq:ensemble-kernel}), in which each model $i$ has a unit feature vector $u_i$ built from its scores against all 38 models, with positive and negative scores in separate channels. The IGP distance is the distance between two models' feature vectors,
\begin{equation}
\operatorname{dist}(i,j)=\lVert u_i-u_j\rVert_2=\sqrt{2-2K_{ij}}.
\label{eq:igp-distance}
\end{equation}
It is non-negative, zero between a model and itself, and satisfies the triangle inequality. Two models are close when they relate to all other models in the same way, so the distance between two models depends on the whole fleet. With it, the pairwise analysis compares one distance with another, since semantic distance is a distance, and tests the quantity that the ensemble analysis selects on. The population IGP distance $\dpop$ builds the kernel on the population IGP score $\spop$, and the pair-specific IGP distance $\dpair$ builds it on the pair-specific IGP score $\spair$. Figure~\ref{fig:rev-four-measures} shows the four measures for the 20 models of Figure~\ref{fig:views}.
\end{revblock}

\subsection{\rev{Results for the four measures}}
\label{app:igp-measures-results}

\begin{revblock}
Table~\ref{tab:main-measures} gives the cross-benchmark estimates for the four measures under all three embeddings, and Table~\ref{tab:rev-measures-bench} the benchmark-level estimates under MiniLM. Table~\ref{tab:rev-compressors-measures} repeats the cross-benchmark estimates for the other compressors. All four are negative under every embedding and every interval excludes zero. The estimates grow from the population IGP score to the pair-specific IGP distance, and the pair-specific estimates do not shrink under the stronger embeddings. The population IGP score is the most conservative of the four, because differences between single models dilute it. In the pair-specific columns of Table~\ref{tab:rev-measures-bench}, WorldSense is the only benchmark with a positive estimate.
\end{revblock}

\begin{table}[htbp]
  \revcolor
  \caption{\textbf{Benchmark-level estimates for the four IGP measures, PPMd and MiniLM.} Partial Spearman correlation with CWA $\kappa$, controlling capability and semantic distance, with 95\% model-node bootstrap intervals. The population columns use CWA as computed with capability level and gap. The pair-specific columns use member-adjusted CWA with the accuracy-product control. The last row is the mean over benchmarks with its 95\% $t$-interval.}
  \label{tab:rev-measures-bench}
  \centering
  \resizebox{\linewidth}{!}{%
  \begin{tabular}{lllll}
    \toprule
    Benchmark & Population score & Population distance & Pair-specific score & Pair-specific distance \\
    \midrule
    TruthfulQA & $-0.043$ $[-0.31,+0.18]$ & $-0.280$ $[-0.49,-0.03]$ & $-0.278$ $[-0.45,-0.09]$ & $-0.362$ $[-0.56,-0.13]$ \\
    MMLU-Pro & $-0.238$ $[-0.47,-0.02]$ & $-0.193$ $[-0.36,+0.00]$ & $-0.353$ $[-0.50,-0.17]$ & $-0.485$ $[-0.63,-0.29]$ \\
    WorldSense & $-0.171$ $[-0.37,+0.04]$ & $-0.152$ $[-0.35,+0.05]$ & $+0.003$ $[-0.23,+0.15]$ & $+0.039$ $[-0.29,+0.24]$ \\
    BBEH-mini & $-0.279$ $[-0.48,+0.05]$ & $-0.260$ $[-0.36,-0.11]$ & $-0.307$ $[-0.47,-0.07]$ & $-0.387$ $[-0.60,-0.01]$ \\
    GSM8K & $-0.031$ $[-0.27,+0.15]$ & $-0.205$ $[-0.38,-0.01]$ & $-0.353$ $[-0.49,-0.17]$ & $-0.375$ $[-0.53,-0.17]$ \\
    AIME & $-0.145$ $[-0.33,+0.03]$ & $-0.099$ $[-0.33,+0.11]$ & $-0.023$ $[-0.18,+0.15]$ & $-0.043$ $[-0.23,+0.14]$ \\
    MuSR & $-0.168$ $[-0.37,+0.00]$ & $-0.292$ $[-0.45,-0.09]$ & $-0.221$ $[-0.38,-0.07]$ & $-0.211$ $[-0.44,+0.05]$ \\
    GPQA-Diamond & $-0.340$ $[-0.49,-0.13]$ & $-0.238$ $[-0.37,-0.08]$ & $-0.236$ $[-0.41,-0.04]$ & $-0.381$ $[-0.61,-0.12]$ \\
    Humanity's Last Exam & $-0.452$ $[-0.66,-0.22]$ & $-0.431$ $[-0.58,-0.24]$ & $-0.515$ $[-0.62,-0.36]$ & $-0.676$ $[-0.77,-0.51]$ \\
    AGIEval & $-0.287$ $[-0.43,-0.08]$ & $-0.119$ $[-0.27,+0.07]$ & $-0.292$ $[-0.47,-0.11]$ & $-0.293$ $[-0.54,-0.07]$ \\
    \midrule
    Mean & $-0.216$ $[-0.309,-0.122]$ & $-0.227$ $[-0.297,-0.158]$ & $-0.257$ $[-0.367,-0.147]$ & $-0.317$ $[-0.465,-0.169]$ \\
    \bottomrule
  \end{tabular}}
\end{table}

\FloatBarrier

\section{\rev{Semantic Embeddings}}
\label{app:embeddings}

\begin{figure}[!t]
  \centering
  \includegraphics[width=0.9\linewidth]{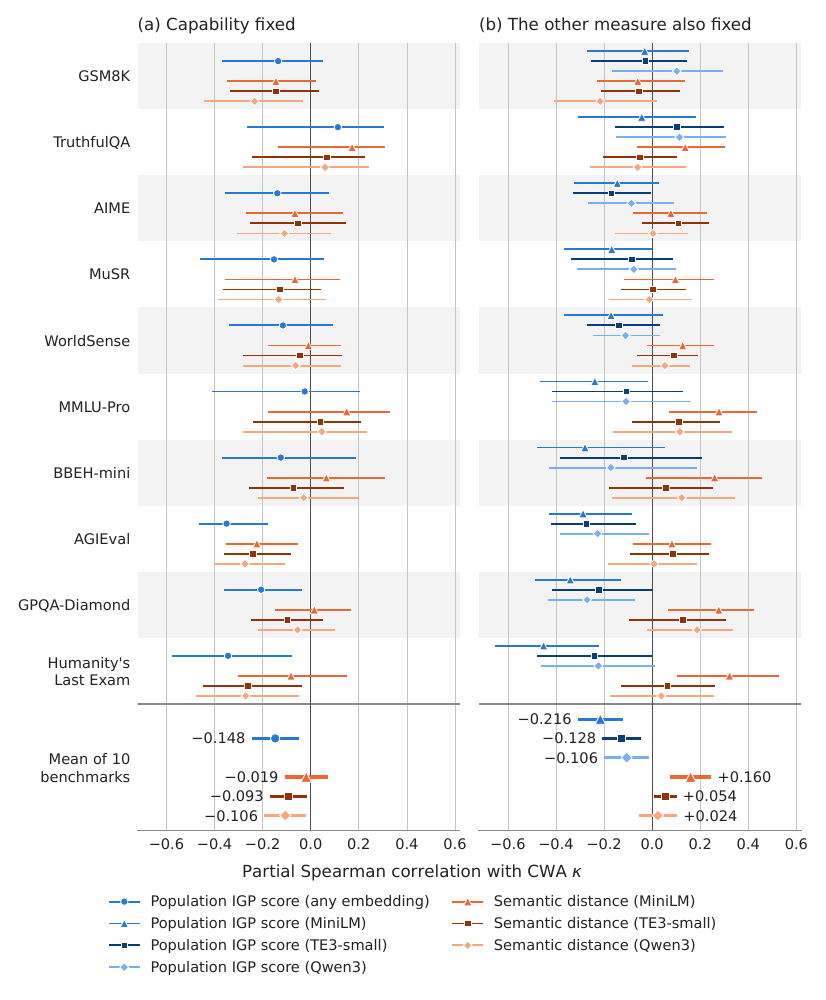}
  \caption{\revcolor{}\textbf{Population IGP score and semantic distance against CWA $\kappa$ under three embeddings, PPMd.} Each point is the partial Spearman correlation on one benchmark, in the layout of Figure~\ref{fig:dissociation}. Blue markers are the population IGP score and orange markers semantic distance; the marker shape gives the embedding. (a) Capability level and gap controlled. The population IGP score does not involve an embedding there, so it has one marker. (b) The other measure also controlled. Thin bars are 95\% model-node bootstrap intervals. The last row is the mean over the ten benchmarks with its 95\% $t$-interval.}
  \label{fig:rev-embedding-forest}
\end{figure}

\begin{revblock}
We use three embedding models to assess how far the results depend on the semantic baseline: MiniLM (\texttt{all-MiniLM-L6-v2}), the primary embedding \citep{reimers_sentencebert_2019}; text-embedding-3-small (TE3-small) \citep{openai_new_2024}, the model used in Artificial Hivemind, from which we take the open-ended prompts \citep{jiang_artificial_2025}; and Qwen3-Embedding (Qwen3), in its 8B version \citep{zhang_qwen3_2025}. Semantic distance is built the same way for each. The embeddings differ in how much of a response they read. MiniLM reads the first 256 tokens. TE3-small accepts 8,191 tokens and Qwen3 32,000, so they read almost every response in full. Only 27 and 11 of the 189,925 responses exceed these limits, and the embedding reads those up to its limit. The three semantic distances rank the 703 model pairs alike, with Spearman correlations of 0.85 to 0.92 between embeddings. The population IGP score agrees with semantic distance at Spearman 0.79 $[0.69,0.88]$ under MiniLM, 0.85 $[0.77,0.92]$ under TE3-small and 0.84 $[0.74,0.91]$ under Qwen3, with model-node bootstrap intervals.

The two stronger embeddings explain more of the shared-error variance, and the population IGP score estimate shrinks accordingly, from $-0.216$ under MiniLM to $-0.128$ and $-0.106$ (Table~\ref{tab:main-measures}, Figure~\ref{fig:rev-embedding-forest}). The comparison is deliberately asymmetric: it sets an untuned general-purpose compressor on raw bytes against a strong embedding model. Table~\ref{tab:rev-semantic} gives the converse, what semantic distance adds beyond each IGP measure. Beyond the population IGP score the semantic estimate is positive and falls towards zero under the stronger embeddings. Beyond either IGP distance it adds nothing. Five of the six intervals include zero and the sixth lies above zero. Beyond the pair-specific IGP score it keeps a small negative association, and all three intervals exclude zero. On its own, with capability controlled, semantic distance is not associated with CWA under MiniLM and is negatively associated under the two stronger embeddings.
\end{revblock}

\begin{table}[htbp]
  \revcolor
  \caption{\textbf{Semantic distance against CWA, beyond IGP and on its own.} Mean partial Spearman correlation over ten benchmarks with its 95\% interval. The pair-specific rows use member-adjusted CWA with the accuracy-product control.}
  \label{tab:rev-semantic}
  \centering
  \resizebox{\linewidth}{!}{%
  \begin{tabular}{llll}
    \toprule
    Semantic distance & MiniLM & TE3-small & Qwen3 \\
    \midrule
    Beyond population IGP score & $+0.160$ $[+0.074,+0.246]$ & $+0.054$ $[+0.007,+0.102]$ & $+0.024$ $[-0.056,+0.104]$ \\
    Beyond population IGP distance & $+0.083$ $[-0.011,+0.177]$ & $+0.033$ $[-0.036,+0.103]$ & $+0.024$ $[-0.064,+0.112]$ \\
    Beyond pair-specific IGP score & $-0.081$ $[-0.143,-0.019]$ & $-0.063$ $[-0.115,-0.011]$ & $-0.105$ $[-0.185,-0.025]$ \\
    Beyond pair-specific IGP distance & $+0.019$ $[-0.025,+0.064]$ & $+0.037$ $[+0.005,+0.068]$ & $+0.019$ $[-0.028,+0.065]$ \\
    Alone, CWA as computed & $-0.019$ $[-0.108,+0.070]$ & $-0.093$ $[-0.169,-0.017]$ & $-0.106$ $[-0.193,-0.019]$ \\
    Alone, member-adjusted CWA & $-0.205$ $[-0.326,-0.085]$ & $-0.166$ $[-0.270,-0.063]$ & $-0.224$ $[-0.359,-0.088]$ \\
    \bottomrule
  \end{tabular}}
\end{table}

\section{\rev{Comparison with PhyloLM}}
\label{app:phylolm}

\begin{revblock}
PhyloLM infers the lineage of language models from their outputs \citep{yax_phylolm_2024}. It treats a prompt as a gene and the first four characters of a model's continuation as an allele, and takes the Nei genetic distance between two models' allele distributions. We ran the authors' released analysis code, unmodified, on our fleet's responses to their 128 math genes, with 32 responses per gene and model. We sent each gene as a bare user message, as their code does for chat models, and generated the responses under our sampling protocol, temperature 1.0 and top-$p$ 0.9, so that every model is on one setting. PhyloLM's defaults are temperature 0.7, top-$p$ 1 and five output tokens. We keep our 32,768-token output budget, because reasoning models return no visible text within five tokens, and the allele is the first four characters either way. We generate 50 responses per gene and use the first 32, their number. The run covers 30 of our 38 models. The other eight are no longer served: Seed-1.6-Flash, DeepSeek-R1-Distill-Llama-70B, Granite-4.0-H-Micro, Granite-4.1-8B, MiniMax-M2, Hermes-3-Llama-3.1-405B, o4-mini and Laguna-XS-2.1.

\begin{figure}[!t]
  \centering
  \includegraphics[width=\linewidth]{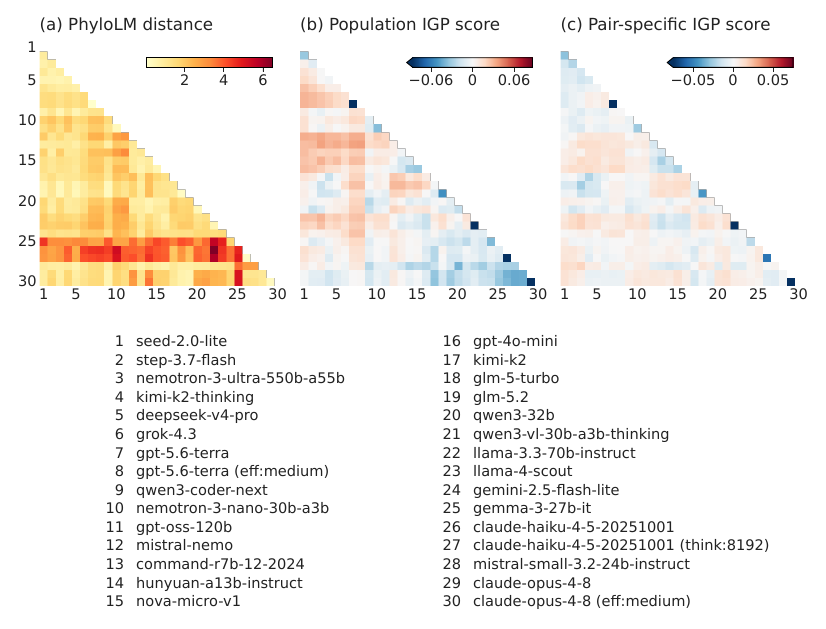}
  \caption{\revcolor{}\textbf{PhyloLM distance beside the two IGP scores, 30 models.} (a) PhyloLM distance on its 128 math genes. (b) Population IGP score. (c) Pair-specific IGP score. The models are in one order in all panels, the average-linkage order on MiniLM semantic distance, and the key gives their names. Panel (a) uses a sequential scale. Panels (b) and (c) use a diverging scale centred at zero whose limits come from pairs of different models. Two pairs that are two settings of one model lie below the scale and take its end colour.}
  \label{fig:rev-phylolm-matrices}
\end{figure}

Table~\ref{tab:rev-phylolm} gives the statistic of Table~\ref{tab:main-measures} with PhyloLM distance in place of semantic distance, on those 30 models, and Figure~\ref{fig:rev-phylolm-matrices} shows the three matrices. PhyloLM distance has no association with correlated wrong-answer agreement beyond IGP, for either population or pair-specific IGP, and IGP is unchanged when PhyloLM is held fixed. The one interval that includes zero is for the population IGP score, estimated here on 30 models. The result does not depend on the prompt set. With our 100 prompts as the genes, on the same 30 models, the pair-specific IGP score beyond PhyloLM is $-0.273$ $[-0.413,-0.134]$ and PhyloLM beyond the pair-specific IGP score is $+0.019$ $[-0.016,+0.054]$. The two PhyloLM matrices agree only moderately with each other, at Spearman 0.60.

The two methods rank model pairs differently. PhyloLM distance correlates with the population IGP score at Spearman 0.10 and with the pair-specific IGP score at 0.30. PhyloLM fingerprints lineage from the first characters of a continuation of a bare text snippet, which assumes raw completion access. Providers serve the models in our fleet through chat endpoints, which wrap every input in a chat template, so the first characters are largely the opening of a reply. The authors note this bias and analyse chat and completion models separately \citep[Sections 3.2.1 and 4]{yax_phylolm_2024}. We could only run PhyloLM in the chat setting, which limits the comparison. It is also the practical difference between the two methods. IGP needs only the outputs that a deployed model returns to ordinary chat prompts.
\end{revblock}

\begin{table}[htbp]
  \revcolor
  \caption{\textbf{IGP and PhyloLM distance against CWA $\kappa$, 30 models.} Mean partial Spearman correlation over ten benchmarks with its 95\% interval and the number of benchmarks with a negative estimate. ``Alone'' controls capability only. The member-adjusted rows include the accuracy-product control.}
  \label{tab:rev-phylolm}
  \centering
  \resizebox{\linewidth}{!}{%
  \begin{tabular}{llrll}
    \toprule
    Outcome & Quantity & $\bar\rho$ & 95\% interval & Negative \\
    \midrule
    CWA as computed & PhyloLM alone & $+0.017$ & $[-0.040,+0.073]$ & 5/10 \\
     & PhyloLM beyond population IGP score & $+0.023$ & $[-0.031,+0.077]$ & 5/10 \\
     & PhyloLM beyond population IGP distance & $+0.026$ & $[-0.026,+0.079]$ & 4/10 \\
     & Population IGP score beyond PhyloLM & $-0.112$ & $[-0.232,+0.008]$ & 7/10 \\
     & Population IGP distance beyond PhyloLM & $-0.216$ & $[-0.272,-0.161]$ & 10/10 \\
    Member-adjusted CWA & PhyloLM alone & $-0.086$ & $[-0.157,-0.016]$ & 8/10 \\
     & PhyloLM beyond pair-specific IGP score & $-0.023$ & $[-0.060,+0.015]$ & 7/10 \\
     & PhyloLM beyond pair-specific IGP distance & $+0.006$ & $[-0.023,+0.035]$ & 5/10 \\
     & Pair-specific IGP score beyond PhyloLM & $-0.270$ & $[-0.402,-0.138]$ & 8/10 \\
     & Pair-specific IGP distance beyond PhyloLM & $-0.310$ & $[-0.472,-0.148]$ & 9/10 \\
    \bottomrule
  \end{tabular}}
\end{table}

\FloatBarrier

\FloatBarrier

\clearpage
\section{\rev{Vote and Abstention Outcomes}}
\label{app:ensemble-votes}

\begin{revblock}
The convergent failure rate conditions on every member being wrong, so it does not show whether a selected ensemble answers more questions correctly. This appendix scores ensembles under explicit decision rules. Its design differs from that of Figure~\ref{fig:ensemble} in the pool, the matching, the estimate and the treatment of multiplicity (Table~\ref{tab:rev-vote-design}).
\end{revblock}

\begin{table}[htbp]
  \revcolor
  \caption{\textbf{The two ensemble designs.}}
  \label{tab:rev-vote-design}
  \centering
  \small
  \begin{tabular}{p{0.15\linewidth}p{0.37\linewidth}p{0.39\linewidth}}
    \toprule
     & Design of Figure~\ref{fig:ensemble} (Appendix~\ref{app:ensemble}) & Vote design (this appendix) \\
    \midrule
    Outcomes & CFR $\kappa$ and its companions & Correct answers, wrong answers and questions answered under four voting rules \\
    Pool & Per benchmark, the ensembles with at least 30 estimable questions & All 501,942 ensembles on every benchmark \\
    Benchmarks & Nine; GSM8K has too few eligible ensembles & Ten \\
    Selection & Top 200 by Vendi score within each benchmark's pool & The same top 200 on every benchmark \\
    Matching & Capability on all questions & Capability on half A of the questions, over 20 splits \\
    Scoring & All questions & Half B of the same split \\
    Estimate & Selected minus matched mean & The same after a regression adjustment, with the unadjusted difference beside it \\
    Multiplicity & No correction & Holm, in two families of tests \\
    \bottomrule
  \end{tabular}
\end{table}

\subsection{\rev{Rules and outcomes}}
\label{app:vote-rules}

\begin{revblock}
Under a $t$-of-5 rule an ensemble answers a question when at least $t$ of its five members give the same answer, for $t=3$, 4 and 5, and abstains otherwise. Majority vote is 3 of 5. Under plurality the answer with strictly the most votes wins, and a tie for first place abstains. Each member casts one vote. Every rate averages over all combinations of the members' parsed epochs, a member with no parsed answer casts no vote, and each question counts once.

For each rule we report three outcomes per 100 questions. Correct answers are the questions that the ensemble answers correctly. Wrong answers are the questions that it answers wrongly, so an abstention does not count as a wrong answer. Questions answered are the sum of the two. An ensemble answers a question correctly, answers it wrongly or does not answer it, so within a rule the three differences between two groups of ensembles sum to zero. We also report the share of answers that are correct, computed within each benchmark. The vote outcomes cover every question of a benchmark and need no estimability rule.
\end{revblock}

\subsection{\rev{Selection, split design and matching}}
\label{app:vote-design}

\begin{revblock}
Selection uses the open-ended responses alone. On every benchmark the selected ensembles are the 200 with the highest Vendi score on a measure among all 501,942 ensembles (Appendix~\ref{app:ensemble-kernel}), so we score the same 200 ensembles on all ten benchmarks. The analysis of Figure~\ref{fig:ensemble} selects within each benchmark's eligible pool instead (Appendix~\ref{app:ensemble-eligibility}). Its cut-off on the Vendi score of the population IGP score runs from 1.757 on AIME to 2.065, and it differs from the single cut-off of 2.065 used here on seven of its eleven benchmark keys. The two analyses therefore select different ensembles on those keys, and the vote results cover ten benchmarks where the CFR results cover nine.

Matching and scoring use disjoint questions. For each benchmark we use 20 stored random splits of its questions into half A and half B. In each split the matching coordinates are the members' mean accuracy and the mean absolute accuracy gap over the ten member pairs, both computed on half A and divided by their standard deviation over the pool. A capability-matched ensemble lies within $\pm\tau$ SD of the selected ensemble on both coordinates. The window starts at $\tau=0.5$, widens by a factor of 1.5 until it holds 50 ensembles, and stops at $\tau=2$. A selected ensemble is a member of its own matched pool, as in Appendix~\ref{app:ensemble-matching}. When semantic distance is held within a band, a matched ensemble also lies in the same band and within 0.1 SD of the selected ensemble's mean semantic distance. We score every outcome on half B of the same split, so no ensemble is scored on a question that helped to match it. The split keeps matching and scoring apart, but the benchmarks and the 38 models are those of the main analysis, so the vote results do not replicate the finding on new models or new tasks.
\end{revblock}

\subsection{\rev{The matched difference and its regression adjustment}}
\label{app:vote-adjustment}

\begin{revblock}
For one benchmark, the unadjusted difference is the mean over splits of the selected ensembles' mean outcome minus the mean of their matched values, where the matched value of a selected ensemble is the mean outcome over its matched pool. The matching leaves the selected ensembles slightly more capable on half A than their matched ensembles\revr{, for the reason given in Appendix~\ref{app:ensemble-matching}}. On GPQA-Diamond the imbalance is $+0.06$ to $+0.15$ SD on mean accuracy and $+0.08$ to $+0.16$ SD on the accuracy gap.

We therefore report a regression-adjusted difference as the primary estimate. \revr{It is the adjustment of Equation~\ref{eq:ensemble-delta} (Appendix~\ref{app:ensemble-inference}), fitted within each split, with the half-B outcome as $y$ and the two half-A matching coordinates as $z$. The selected ensembles are 0.05\% to 0.5\% of the rows of the fit.} Under majority vote the adjustment lowers the gain in correct answers from 1.95 to 0.84 per 100 and the reduction in wrong answers from 1.58 to 1.26, so Table~\ref{tab:rev-votes-full} gives the unadjusted difference beside every adjusted one.
\end{revblock}

\subsection{\rev{Inference and Holm families}}
\label{app:vote-inference}

\begin{revblock}
Inference is across benchmarks. The 12 benchmark keys collapse to ten benchmarks, with the three BBEH-mini keys weighted by their question counts. We report the mean difference over the ten benchmarks, its 95\% $t$-interval and the $p$-value of the $t$-test.

The vote outcomes form two different families of tests, each with its own Holm correction (Equation~\ref{eq:holm}). The majority-vote family holds correct and wrong answers at 3 of 5, and tests whether selection improves a voting ensemble. The abstention family holds the same two outcomes at 4 of 5 and at 5 of 5, and tests what a stricter agreement rule changes. Plurality, a sensitivity analysis, and questions answered are in no family and keep their unadjusted $p$-values.
\end{revblock}

\subsection{\rev{Results for the 200 selected ensembles}}
\label{app:vote-results}

\begin{revblock}
Table~\ref{tab:rev-votes-full} gives the outcomes behind Figure~\ref{fig:rev-vote-rules} for the ensembles selected on the population IGP score, with no semantic distance held fixed. Under plurality and majority vote they give more correct and fewer wrong answers than capability-matched ensembles. Under majority vote the adjusted differences are $+0.84$ $[+0.10,+1.57]$ correct and $-1.26$ $[-2.17,-0.35]$ wrong answers per 100, and the unadjusted differences are $+1.95$ $[+0.70,+3.21]$ and $-1.58$ $[-2.52,-0.64]$. The Holm-adjusted $p$-values are 0.030 and 0.024 in the majority-vote family. Under the 4-of-5 rule the selected ensembles answer fewer questions, so they give fewer correct and fewer wrong answers, and when they answer they are right as often as the matched ensembles. Under 5 of 5 no interval excludes zero.

Ensembles selected on semantic distance gain less. Their majority-vote wrong answers fall by 0.48, 0.45 and 0.26 per 100 under MiniLM, TE3-small and Qwen3, against 1.26 for the population IGP score, and their correct answers rise by 0.65, 0.07 and 0.50.
\end{revblock}

\begin{table}[htbp]
  \revcolor
  \caption{\textbf{Top 200 ensembles by Vendi score on the population IGP score against capability-matched ensembles, per 100 questions.} No semantic distance is held fixed. Matched is the adjusted level of the matched ensembles, and Difference the adjusted difference with its 95\% interval over ten benchmarks. Unadjusted is the difference before the regression adjustment. Holm $p$ is for the adjusted difference, within its family. Correct when answering is the share of answers that are correct, in per cent. CFR $\kappa$ is not per 100 questions.}
  \label{tab:rev-votes-full}
  \centering
  \resizebox{\linewidth}{!}{%
  \begin{tabular}{lrrllr}
    \toprule
    Outcome & Selected & Matched & Difference & Unadjusted & Holm $p$ \\
    \midrule
    \multicolumn{6}{l}{\emph{Plurality}} \\
    Correct answers & 78.1 & 76.8 & $+1.23$ $[+0.61,+1.85]$ & $+2.34$ $[+1.01,+3.67]$ & --- \\
    Wrong answers & 14.6 & 15.7 & $-1.10$ $[-1.61,-0.59]$ & $-1.61$ $[-2.20,-1.01]$ & --- \\
    Questions answered & 92.7 & 92.5 & $+0.13$ $[-0.50,+0.76]$ & $+0.73$ $[-0.43,+1.90]$ & --- \\
    \midrule
    \multicolumn{6}{l}{\emph{Majority vote (3 of 5)}} \\
    Correct answers & 72.4 & 71.6 & $+0.84$ $[+0.10,+1.57]$ & $+1.95$ $[+0.70,+3.21]$ & 0.030 \\
    Wrong answers & 11.4 & 12.7 & $-1.26$ $[-2.17,-0.35]$ & $-1.58$ $[-2.52,-0.64]$ & 0.024 \\
    Questions answered & 83.9 & 84.3 & $-0.42$ $[-1.74,+0.89]$ & $+0.37$ $[-1.39,+2.13]$ & --- \\
    Correct when answering (\%) & 83.6 & 82.6 & $+1.00$ $[+0.38,+1.62]$ & --- & --- \\
    \midrule
    \multicolumn{6}{l}{\emph{4 of 5}} \\
    Correct answers & 50.7 & 51.8 & $-1.04$ $[-1.94,-0.14]$ & $-0.65$ $[-1.39,+0.09]$ & 0.11 \\
    Wrong answers & 5.0 & 5.7 & $-0.77$ $[-1.69,+0.14]$ & $-0.90$ $[-1.81,+0.01]$ & 0.26 \\
    Questions answered & 55.7 & 57.5 & $-1.82$ $[-3.26,-0.37]$ & $-1.55$ $[-2.93,-0.16]$ & --- \\
    Correct when answering (\%) & 84.3 & 84.5 & $-0.26$ $[-1.48,+0.96]$ & --- & --- \\
    \midrule
    \multicolumn{6}{l}{\emph{5 of 5}} \\
    Correct answers & 28.0 & 28.2 & $-0.17$ $[-0.93,+0.58]$ & $-0.63$ $[-1.36,+0.11]$ & 0.61 \\
    Wrong answers & 1.4 & 1.9 & $-0.44$ $[-1.07,+0.18]$ & $-0.51$ $[-1.13,+0.12]$ & 0.29 \\
    Questions answered & 29.4 & 30.0 & $-0.62$ $[-1.69,+0.46]$ & $-1.13$ $[-2.11,-0.15]$ & --- \\
    \midrule
    \multicolumn{6}{l}{\emph{Design of Figure~\ref{fig:ensemble}a, nine benchmarks}} \\
    CFR $\kappa$ & --- & --- & $-0.047$ $[-0.095,+0.000]$ & --- & --- \\
    All five give the same wrong answer & --- & --- & $-0.64$ $[-1.56,+0.29]$ & --- & --- \\
    All five wrong & --- & --- & $-0.63$ $[-1.05,-0.21]$ & --- & --- \\
    \bottomrule
  \end{tabular}}
\end{table}

\FloatBarrier

\subsection{\rev{Wrong answers at the same coverage}}
\label{app:vote-coverage}

\begin{revblock}
An ensemble that answers less also gives fewer wrong answers. To separate the two, we compare wrong answers at the same coverage, that is at the same number of questions answered. For each benchmark we take the matched ensembles' adjusted levels of questions answered and of wrong answers at 5 of 5, 4 of 5 and 3 of 5, and join the three points by straight lines. We read this line at the selected ensembles' coverage under the rule of interest and subtract the value from the selected ensembles' wrong answers. We do not extrapolate, so a benchmark counts only when the selected ensembles' coverage lies within the matched ensembles' range.

Under majority vote the range ends at the matched ensembles' own majority-vote point, so a benchmark drops out when the selected ensembles answer more questions than the matched ones. This removes GPQA-Diamond, GSM8K and TruthfulQA and leaves seven benchmarks, where BBEH-mini counts through one of its three answer-type keys. On those seven the selected ensembles give 14.4 wrong answers per 100 against 15.4 for matched ensembles at the same coverage, a difference of $-0.99$ $[-1.55,-0.42]$. Under the 4-of-5 rule all ten benchmarks count and the difference is $-0.11$ $[-0.40,+0.17]$, at 5.0 against 5.1. Under 5 of 5 only three benchmarks count, and the difference is $-0.12$ $[-0.42,+0.17]$. The reduction in wrong answers therefore holds at the same coverage under majority vote. Under the 4-of-5 rule it reflects answering less.
\end{revblock}

\subsection{\rev{Results by benchmark}}
\label{app:vote-benchmarks}

\begin{figure}[!t]
  \centering
  \includegraphics[width=0.85\linewidth]{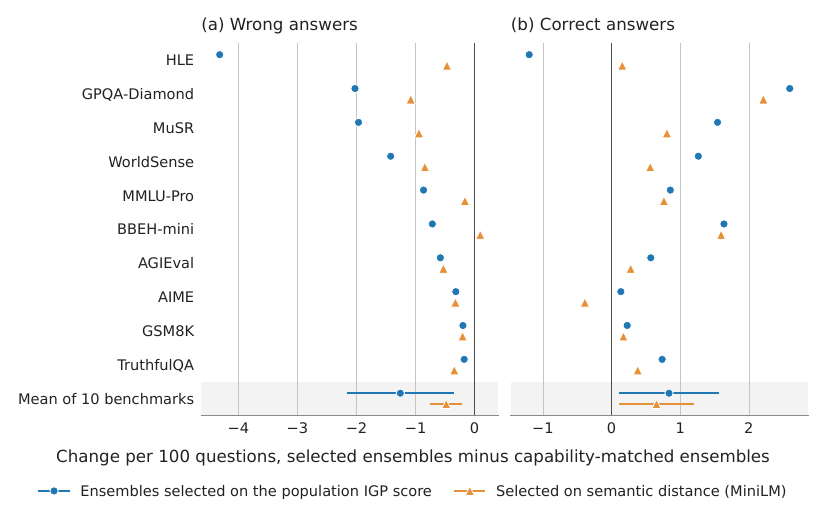}
  \caption{\revcolor{}\textbf{Majority vote by benchmark.} Change per 100 questions in (a) wrong answers and (b) correct answers, for the 200 ensembles with the highest Vendi score minus capability-matched ensembles, after the regression adjustment. Blue circles are ensembles selected on the population IGP score, and orange triangles are ensembles selected on semantic distance under MiniLM. We order the benchmarks by the change in wrong answers for the population IGP score, and HLE is Humanity's Last Exam. The last row is the mean over the ten benchmarks with its 95\% $t$-interval.}
  \label{fig:rev-vote-benchmarks}
\end{figure}

\begin{revblock}
The majority-vote result holds across benchmarks (Figure~\ref{fig:rev-vote-benchmarks}, Table~\ref{tab:rev-vote-benchmarks}). The selected ensembles give fewer wrong answers on all ten benchmarks and more correct answers on nine. Humanity's Last Exam is the outlier. It has the largest reduction in wrong answers, 4.3 per 100, and it is the only benchmark on which the selected ensembles give fewer correct answers, by 1.2 per 100. There they leave 5.5 more questions per 100 unanswered.
\end{revblock}

\begin{table}[!t]
  \revcolor
  \caption{\textbf{Majority vote by benchmark.} Top 200 ensembles by Vendi score on the population IGP score against capability-matched ensembles, per 100 questions. Levels are selected / matched, and differences are regression-adjusted. The last row is the mean over benchmarks with its 95\% $t$-interval.}
  \label{tab:rev-vote-benchmarks}
  \centering
  \resizebox{\linewidth}{!}{%
  \begin{tabular}{llllll}
    \toprule
    Benchmark & Correct, levels & Correct, difference & Wrong, levels & Wrong, difference & Answered, difference \\
    \midrule
    TruthfulQA & 85.5 / 84.8 & $+0.74$ & 7.8 / 8.0 & $-0.18$ & $+0.56$ \\
    MMLU-Pro & 83.9 / 83.1 & $+0.86$ & 9.0 / 9.9 & $-0.86$ & $-0.01$ \\
    WorldSense & 96.0 / 94.8 & $+1.26$ & 3.2 / 4.6 & $-1.42$ & $-0.16$ \\
    BBEH-mini & 39.2 / 37.6 & $+1.64$ & 10.0 / 10.7 & $-0.72$ & $+0.92$ \\
    GSM8K & 96.7 / 96.5 & $+0.23$ & 2.2 / 2.4 & $-0.20$ & $+0.03$ \\
    AIME & 71.1 / 71.0 & $+0.14$ & 0.2 / 0.5 & $-0.32$ & $-0.18$ \\
    MuSR & 70.7 / 69.1 & $+1.54$ & 25.2 / 27.2 & $-1.97$ & $-0.42$ \\
    GPQA-Diamond & 79.2 / 76.6 & $+2.60$ & 7.3 / 9.3 & $-2.03$ & $+0.57$ \\
    Humanity's Last Exam & 11.2 / 12.4 & $-1.20$ & 42.7 / 47.0 & $-4.32$ & $-5.52$ \\
    AGIEval & 90.7 / 90.1 & $+0.57$ & 6.8 / 7.3 & $-0.58$ & $-0.01$ \\
    \midrule
    Mean & 72.4 / 71.6 & $+0.84$ $[+0.10,+1.57]$ & 11.4 / 12.7 & $-1.26$ $[-2.17,-0.35]$ & $-0.42$ $[-1.74,+0.89]$ \\
    \bottomrule
  \end{tabular}}
\end{table}

\newpage
\subsection{\rev{Semantic distance held within a band}}
\label{app:vote-bands}

\begin{figure}[!t]
  \centering
  \includegraphics[width=0.9\linewidth]{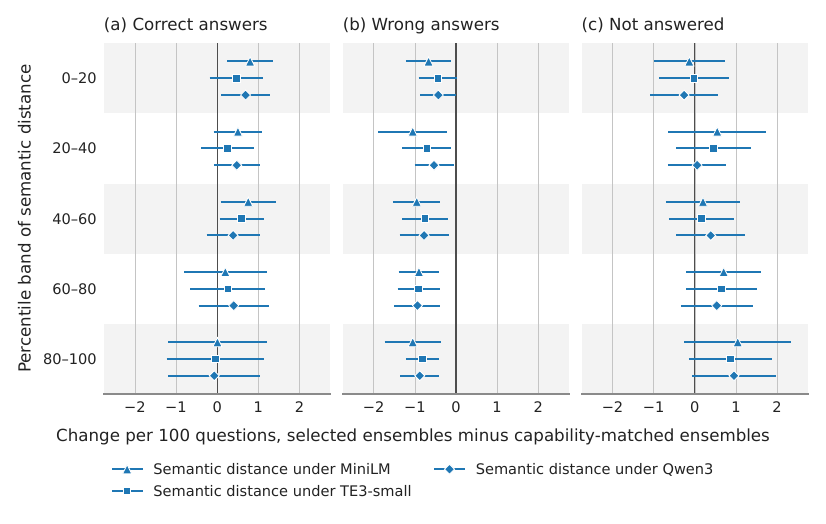}
  \caption{\revcolor{}\textbf{Majority vote with semantic distance held within a band.} Each point is the mean over ten benchmarks of the change per 100 questions in (a) correct answers, (b) wrong answers and (c) questions not answered. The change is the 200 ensembles with the highest Vendi score on the population IGP score within a percentile band of semantic distance, minus capability-matched ensembles from the same band. The marker gives the embedding that defines semantic distance. Bars are unadjusted 95\% $t$-intervals over benchmarks.}
  \label{fig:rev-vote-bands}
\end{figure}

\begin{revblock}
The band design of Figure~\ref{fig:ensemble}b carries over to the vote outcomes. Within each of five percentile bands of mean pairwise semantic distance, we select the 200 ensembles with the highest Vendi score on the population IGP score and compare them with capability-matched ensembles from the same band (Appendix~\ref{app:vote-design}). Figure~\ref{fig:rev-vote-bands} and Table~\ref{tab:rev-vote-bands} give the majority-vote differences.

Under MiniLM, wrong answers fall by 0.7 to 1.1 per 100 in every band, and all five intervals lie below zero. Across the three embeddings they fall in all 15 band cells, and 13 of the 15 unadjusted intervals exclude zero. After the Holm adjustment, 11 cells stay below 0.05 in the majority-vote family. Correct answers change by $+0.0$ to $+0.8$ per 100 under MiniLM, and two of the five intervals lie above zero.

Pooled across the five bands as in Appendix~\ref{app:ensemble-pooled}, wrong answers fall by 0.92 per 100 under MiniLM, on all ten benchmarks, and by 0.73 and 0.71 under TE3-small and Qwen3, and every interval lies below zero. The pooled change in correct answers is positive under every embedding and its interval includes zero. For the reverse selection, on semantic distance with the population IGP score held within a band, the pooled change in wrong answers is $-0.29$ $[-0.67,+0.09]$ under MiniLM.
\end{revblock}

\begin{table}[!t]
  \revcolor
  \caption{\textbf{Majority vote with semantic distance held within a band.} Top 200 ensembles by Vendi score on the population IGP score within each percentile band of semantic distance, minus capability-matched ensembles from the same band, per 100 questions, with the 95\% interval over ten benchmarks. $^{\ast}$ Holm $p<0.05$ in the majority-vote family. The pooled rows average the five bands within each benchmark and give the number of benchmarks with fewer wrong answers or more correct answers.}
  \label{tab:rev-vote-bands}
  \centering
  \resizebox{\linewidth}{!}{%
  \begin{tabular}{llll}
    \toprule
    Band of semantic distance & MiniLM & TE3-small & Qwen3 \\
    \midrule
    \multicolumn{4}{l}{\emph{Wrong answers}} \\
    0--20 & $-0.67$ $[-1.22,-0.12]$$^{\ast}$ & $-0.44$ $[-0.91,+0.02]$ & $-0.43$ $[-0.87,+0.01]$ \\
    20--40 & $-1.05$ $[-1.89,-0.21]$$^{\ast}$ & $-0.71$ $[-1.31,-0.11]$ & $-0.53$ $[-1.01,-0.06]$ \\
    40--60 & $-0.95$ $[-1.53,-0.38]$$^{\ast}$ & $-0.76$ $[-1.31,-0.20]$$^{\ast}$ & $-0.78$ $[-1.37,-0.18]$$^{\ast}$ \\
    60--80 & $-0.90$ $[-1.39,-0.41]$$^{\ast}$ & $-0.91$ $[-1.42,-0.39]$$^{\ast}$ & $-0.94$ $[-1.50,-0.38]$$^{\ast}$ \\
    80--100 & $-1.05$ $[-1.73,-0.37]$$^{\ast}$ & $-0.82$ $[-1.22,-0.41]$$^{\ast}$ & $-0.88$ $[-1.35,-0.41]$$^{\ast}$ \\
    Pooled & $-0.92$ $[-1.53,-0.32]$ 10/10 & $-0.73$ $[-1.20,-0.25]$ 9/10 & $-0.71$ $[-1.19,-0.24]$ 9/10 \\
    \midrule
    \multicolumn{4}{l}{\emph{Correct answers}} \\
    0--20 & $+0.80$ $[+0.25,+1.35]$$^{\ast}$ & $+0.46$ $[-0.18,+1.11]$ & $+0.69$ $[+0.10,+1.28]$ \\
    20--40 & $+0.51$ $[-0.08,+1.09]$ & $+0.26$ $[-0.39,+0.90]$ & $+0.48$ $[-0.08,+1.03]$ \\
    40--60 & $+0.76$ $[+0.09,+1.42]$$^{\ast}$ & $+0.59$ $[+0.06,+1.13]$$^{\ast}$ & $+0.39$ $[-0.26,+1.03]$ \\
    60--80 & $+0.20$ $[-0.80,+1.20]$ & $+0.26$ $[-0.65,+1.17]$ & $+0.40$ $[-0.45,+1.25]$ \\
    80--100 & $+0.01$ $[-1.19,+1.20]$ & $-0.05$ $[-1.23,+1.13]$ & $-0.07$ $[-1.19,+1.05]$ \\
    Pooled & $+0.45$ $[-0.22,+1.12]$ 8/10 & $+0.30$ $[-0.42,+1.03]$ 8/10 & $+0.38$ $[-0.25,+1.01]$ 7/10 \\
    \bottomrule
  \end{tabular}}
\end{table}

\subsection{\rev{Pair-specific IGP on member-adjusted outcomes}}
\label{app:vote-pair-specific}

\begin{revblock}
We test the pair-specific IGP score only against member-adjusted outcomes. The member adjustment of an ensemble outcome subtracts the least-squares sum of its five members' effects, fitted over the pool (Equation~\ref{eq:member-adjustment}), so a positive value means that the ensemble does better than its members predict. The split design, the matching and the regression adjustment are those above.

Under majority vote, the 200 ensembles with the highest Vendi score on the pair-specific IGP score exceed what their members predict by 2.67 correct answers per 100, and their capability-matched ensembles by 1.99. The difference is $+0.68$ $[+0.20,+1.16]$. For wrong answers the two values are $-0.79$ and $-0.57$, a difference of $-0.22$ $[-0.41,-0.02]$. The Holm-adjusted $p$-values are 0.022 and 0.034 in the majority-vote family. Under the 4-of-5 rule these ensembles answer fewer questions than capability-matched ensembles, by 1.10 $[0.53,1.66]$ per 100, as the ensembles selected on the population IGP score do. In the design of Figure~\ref{fig:ensemble}a, on nine benchmarks and member-adjusted outcomes, their CFR $\kappa$ differs from that of capability-matched ensembles by $-0.014$ $[-0.026,-0.002]$. With semantic distance held within a band, the pooled majority-vote differences under MiniLM are $-0.29$ $[-0.48,-0.11]$ wrong and $+0.53$ $[+0.16,+0.91]$ correct answers per 100.
\end{revblock}

\subsection{\rev{The ensemble against its own members}}
\label{app:vote-members}

\begin{figure}[htbp]
  \centering
  \includegraphics[width=0.85\linewidth]{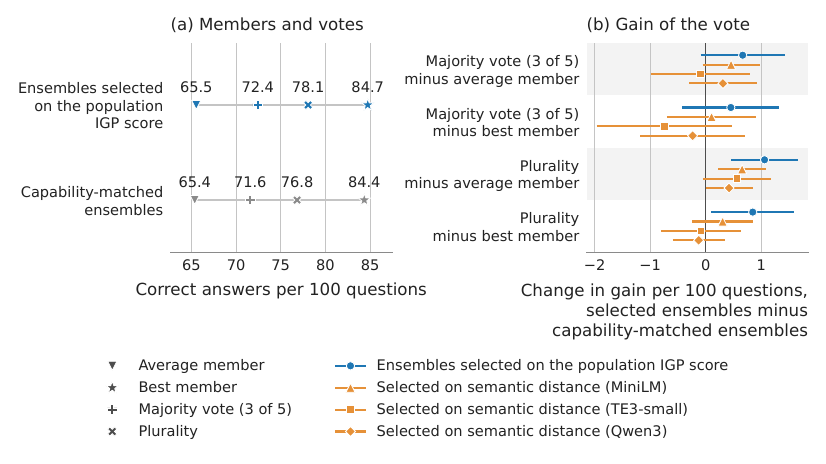}
  \caption{\revcolor{}\textbf{The ensemble against its own members.} (a) Correct answers per 100 questions for the ensembles selected on the population IGP score and for their capability-matched ensembles, as means over ten benchmarks. The average member is the mean of the five members' own accuracies. The best member is the member with the highest accuracy on the half of the questions not used for scoring. (b) Change in the gain of the vote over a member, selected minus capability-matched ensembles, with 95\% $t$-intervals over benchmarks. Blue circles are ensembles selected on the population IGP score, and orange markers are ensembles selected on semantic distance under MiniLM, TE3-small and Qwen3.}
  \label{fig:rev-ensemble-members}
\end{figure}

\begin{revblock}
Figure~\ref{fig:rev-ensemble-members} compares each vote with the ensemble's own members. The average member is the mean of the five members' own accuracies, where a member's accuracy on a question is the share of its parsed epochs that are correct. The best member is the member with the highest accuracy on half A, so it is the best single model that one could pick without seeing the scored questions.

For the ensembles selected on the population IGP score, the average member answers 65.5 questions per 100 correctly, majority vote 72.4, plurality 78.1 and the best member 84.7. For the matched ensembles the four levels are 65.4, 71.6, 76.8 and 84.4. Both votes improve on the average member and neither reaches the best member, for selected and matched ensembles alike. A majority vote counts an unanswered question as not correct, whereas plurality leaves only ties unanswered, so plurality is the like-for-like comparison with a single member. The gain of plurality over the average member is 1.06 $[0.46,1.66]$ per 100 larger in the selected ensembles than in the matched ones, and its gain over the best member is 0.85 $[0.10,1.59]$ larger. For majority vote the two differences are $+0.67$ $[-0.09,+1.42]$ and $+0.45$ $[-0.42,+1.33]$. These comparisons are descriptive and are in no Holm family.
\end{revblock}

\FloatBarrier

\subsection{\rev{Outcomes along each Vendi score}}
\label{app:vote-vendi}

\begin{figure}[htbp]
  \centering
  \includegraphics[width=\linewidth]{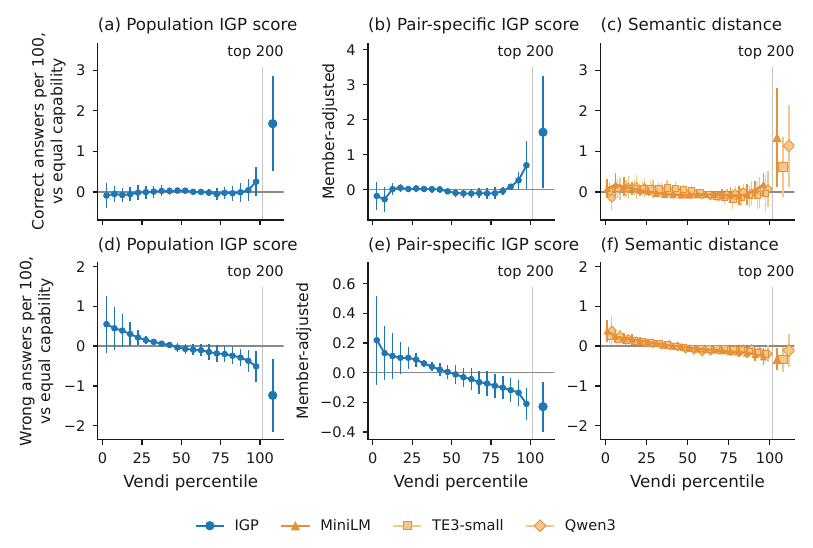}
  \caption{\revcolor{}\textbf{Majority-vote outcomes along each Vendi score, against ensembles of the same capability.} All 501,942 ensembles in 20 equal bins by the percentile of a Vendi score, with the 200 highest-scoring ensembles shown separately at the right. Points are means over ten benchmarks of correct answers (top) and wrong answers (bottom) per 100 questions, with 95\% $t$-intervals. In (a), (c), (d) and (f) the outcome is the residual of a regression on the members' mean accuracy, its square and their mean pairwise accuracy gap, so zero is the average of ensembles of the same capability, and within a row these panels share a vertical range. In (b) and (e) the outcome is also member-adjusted, so zero is what the members predict, and the two panels have their own vertical range. In (c) and (f) MiniLM is the primary embedding and the other two embeddings are drawn lighter.}
  \label{fig:rev-vendi-accuracy}
\end{figure}

\begin{figure}[htbp]
  \centering
  \includegraphics[width=0.9\linewidth]{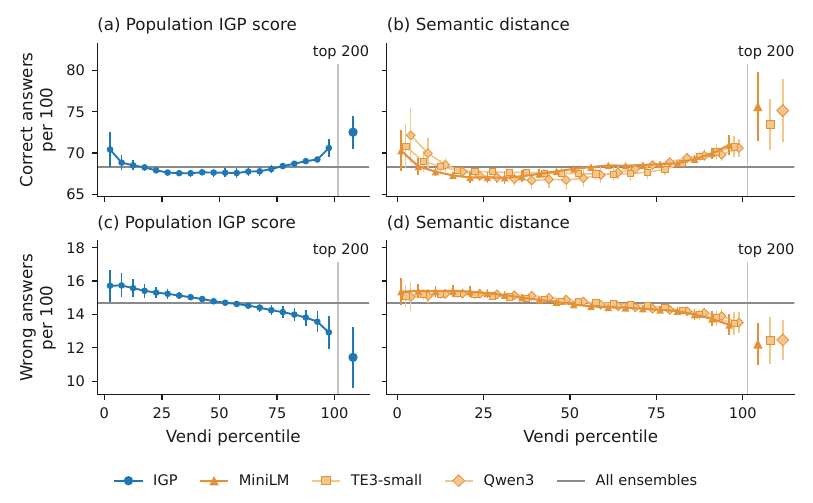}
  \caption{\revcolor{}\textbf{Majority-vote outcomes along each Vendi score, with no adjustment.} The bins and the 200 highest-scoring ensembles are those of Figure~\ref{fig:rev-vendi-accuracy}. Points are means over ten benchmarks of the questions per 100 that the ensembles answer correctly (top) or wrongly (bottom). Each bar is the 95\% $t$-interval over benchmarks of the difference between the bin and the benchmark's mean over all ensembles, and the grey line is that mean. Nothing is held fixed, so the bins also differ in how capable their members are. The pair-specific IGP score has no panel because we test it only on member-adjusted outcomes.}
  \label{fig:rev-vendi-accuracy-raw}
\end{figure}

\begin{revblock}
The association is not confined to the 200 selected ensembles. Figure~\ref{fig:rev-vendi-accuracy} splits all 501,942 ensembles into 20 equal bins by the percentile of a Vendi score and shows the 200 highest-scoring ensembles separately. With capability removed by regression, majority-vote wrong answers fall steadily as the Vendi score rises, for the population and the pair-specific IGP score, and less steeply for semantic distance. Figure~\ref{fig:rev-vendi-accuracy-raw} shows the same bins with no adjustment, in the outcome's own units. There the bins also differ in how capable their members are.

Across all ensembles of a benchmark, with the ranks of the members' mean accuracy and of their mean accuracy gap controlled, the partial Spearman correlation between the Vendi score on the population IGP score and majority-vote wrong answers is $-0.209$ $[-0.264,-0.155]$ as a mean over ten benchmarks. It is $-0.120$ to $-0.143$ with the semantic Vendi score also controlled. The semantic Vendi score gives $-0.153$ to $-0.175$ on its own and $-0.034$ to $+0.025$ beyond the IGP score. For correct answers the correlation with the IGP Vendi score is $+0.096$ $[-0.008,+0.200]$, so across the whole pool the association is mainly with wrong answers. For the pair-specific IGP score on member-adjusted outcomes the correlation with wrong answers is $-0.227$ $[-0.295,-0.159]$.
\end{revblock}

\FloatBarrier

\end{document}